\documentclass[pdflatex,sn-mathphys-num]{sn-jnl}

\usepackage{graphicx} % Required for inserting images
\usepackage{amssymb}
\usepackage{pifont}
\usepackage{url}
\usepackage{hyperref}
\usepackage{amsmath}
\usepackage{amsfonts}
\usepackage{bm}
\usepackage{subcaption}
\usepackage{bbm}
\usepackage{multirow}
\usepackage{enumitem}
\usepackage{etoolbox}

\usepackage[dvipsnames]{xcolor}

\begin{document}

\title[DeepRehabPile]{A Standardized Benchmark for Skeleton-Based Rehabilitation Assessment Using Deep Learning}

\author*[1]{\fnm{Ali} \sur{Ismail-Fawaz}}\email{ali-el-hadi.ismail-fawaz@uha.fr}\equalcont{These authors contributed equally to this work.}
\author[1]{\fnm{Maxime} \sur{Devanne}}\email{maxime.devanne@uha.fr}\equalcont{These authors contributed equally to this work.}
\author[2]{\fnm{Stefano} \sur{Berretti}}\email{stefano.berretti@unifi.it}
\author[1]{\fnm{Jonathan} \sur{Weber}}\email{jonathan.weber@uha.fr}
\author[1,3]{\fnm{Germain} \sur{Forestier}}\email{germain.forestier@uha.fr}\email{germain.forestier@monash.edu}

\affil*[1]{\orgdiv{IRIMAS}, \orgname{Universite de Haute-Alsace}, \city{Mulhouse}, \country{France}}
\affil[2]{\orgdiv{MICC}, \orgname{University of Florence}, \city{Florence}, \country{Iraly}}
\affil[3]{\orgdiv{DSAI}, \orgname{Monash University}, \city{Melbourne}, \country{Australia}}

% \equalcont{Ali Ismail-Fawaz and Maxime Devanne contributed equally to this work.}

\abstract{
Automated assessment of human motion plays a vital role in rehabilitation, enabling objective evaluation of patient performance and progress.
Unlike general human activity recognition, rehabilitation motion assessment focuses on analyzing the quality of movement within the same action class, requiring the detection of subtle deviations from ideal motion.
Recent advances in deep learning and video-based skeleton extraction have opened new possibilities for accessible, scalable motion assessment using affordable devices such as smartphones or webcams.
However, the field lacks standardized benchmarks, consistent evaluation protocols, and reproducible methodologies, limiting progress and comparability across studies.
In this work, we address these gaps by (i) aggregating existing rehabilitation datasets into a unified archive called \textbf{Rehab-Pile}, (ii) proposing a general benchmarking framework for evaluating deep learning methods in this domain, and (iii) conducting extensive benchmarking of multiple architectures across classification and regression tasks.
All datasets and implementations are released to the community to support transparency and reproducibility.
This paper aims to establish a solid foundation for future research in automated rehabilitation assessment and foster the development of reliable, accessible, and personalized rehabilitation solutions.
The datasets, source-code and results of this article are all publicly available: \url{https://msd-irimas.github.io/pages/DeepRehabPile/}
}


\maketitle

\begin{figure}[h]
    \centering
    \includegraphics[width=0.8\linewidth]{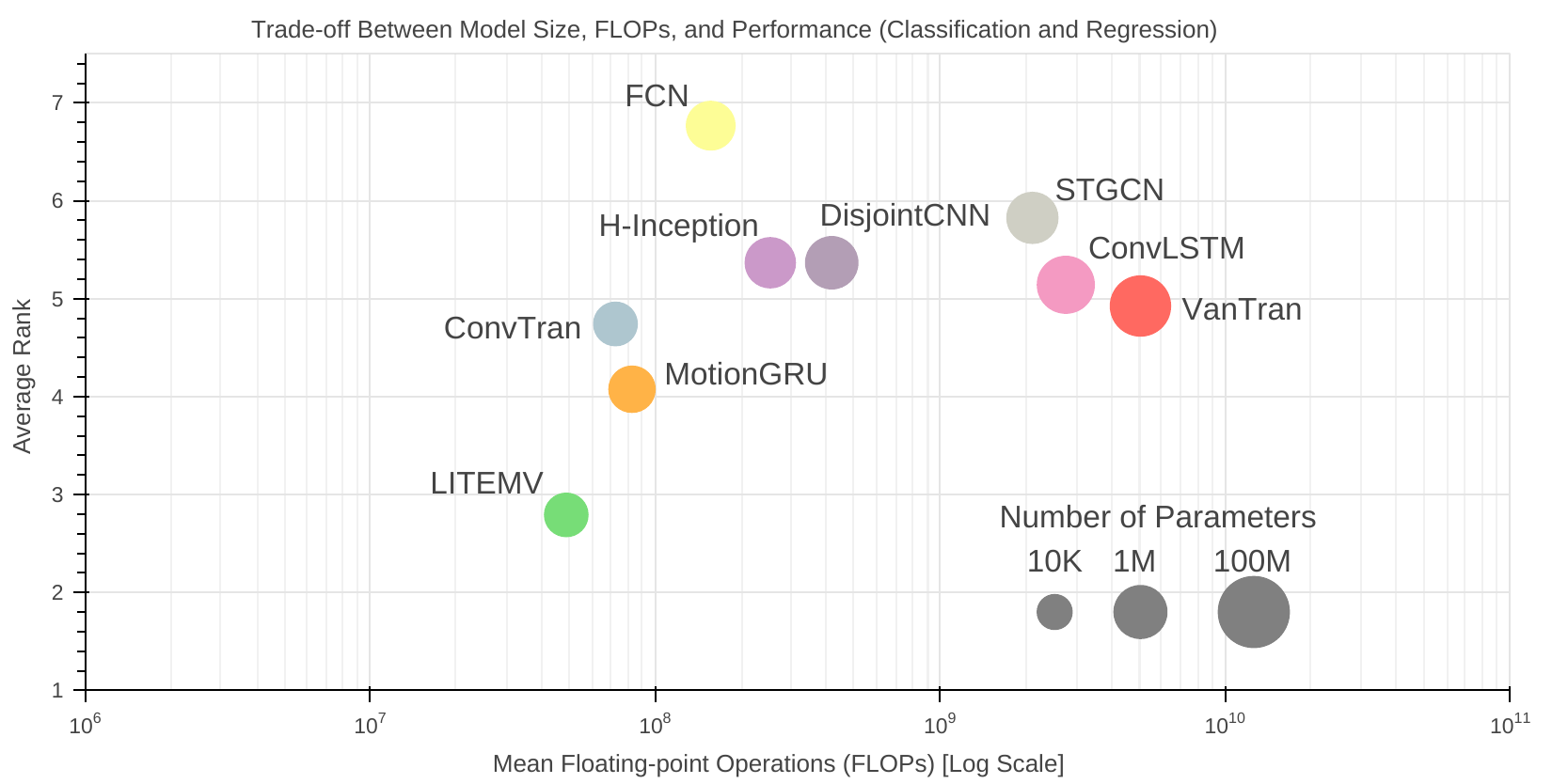}
    \caption{
    % \jwc{start average rank to 1 and FLOPS to $10^6$}\hfc{fixed}
    Comparing nine different deep learning models in terms of performance and efficiency (FLOPs and number of trainable parameters) on all $60$ datasets of classification and extrinsic regression. The performance is measured on the average rank over all datasets using both the accuracy and negated Mean Absolute Error (MAE).}
    \label{fig:trade-off-perf-eff}
\end{figure}

\section{Introduction}

The assessment and analysis of human motion are critical in many fields, particularly in healthcare and rehabilitation~\cite{devanne2017multi,2024_FineRehab,ismail2025deep,zhu2024walking}. 
With advances in computer vision~\cite{zaher2025fusing}, motion sensors~\cite{garcia2022database}, and artificial intelligence~\cite{sardari2023artificial,sumner2023artificial,kim2017fitmine}, there is a growing opportunity to enhance rehabilitation therapy through objective and automated motion evaluation.
This is especially important for monitoring patient progress, tailoring treatments, and ensuring the correct execution of prescribed exercises.  

Rehabilitation motion assessment~\cite{liao2020review,ismail2024weighted,teodorescu2001fuzzy}, however, differs significantly from other human motion analysis tasks, such as action recognition~\cite{amass-archive,gru-svae}. 
While action recognition focuses on classifying motion types (e.g., walking vs. running), rehabilitation assessment requires evaluating the quality and performance of movements within the same class of activity. 
For example, the task is to determine whether a squat is performed correctly, assess its biomechanical correctness, and quantify the deviation from an ideal motion. 
This distinction requires methods capable of capturing subtle performance variations rather than simply identifying motion categories.  

In recent years, deep learning has emerged as a powerful tool for addressing the challenges of rehabilitation motion assessment~\cite{ismail2025deep,mennella2023deep,litemv}. 
This work particularly focuses on skeleton data extracted from video streams, whether RGB or RGB-D. Skeleton data represent human motion as a set of key joint coordinates, offering a compact and interpretable representation of body movement over time.
The primary motivation for relying on skeleton data from video streams is to promote broad accessibility. 
Unlike motion capture (MoCap) systems~\cite{canton2010marker}, which are highly accurate and robust but costly, intrusive, and often impractical for large-scale or at-home use, video-based approaches~\cite{asteriadis2013estimating,Cao2019ITPAMI} can be implemented with affordable, readily available devices such as smartphones, webcams, or depth cameras. 
This accessibility aligns with the goal of making automated rehabilitation assessment widely available to the general public. 
Deep learning methods, with their ability to model complex spatio-temporal patterns in skeleton data, provide a promising avenue for evaluating motion quality in this context.  

In every field of machine learning research, the development of standardized benchmarks is crucial~\cite{dl4tsc,bake-off-tsc,ismail2025data-aug,meyer2025pooling}. 
Benchmarks enable consistent evaluation, promote reproducibility, and support the incremental advancement of methods by providing a common ground for comparison. 
Across domains, researchers have recognized this need and responded by creating and adopting widely accepted benchmark datasets. 
For example, ImageNet~\cite{imagenet-archive} has become the standard for image classification, AMASS~\cite{amass-archive} for human activity recognition, and the UCR~\cite{ucr-archive} UEA~\cite{uea-archive} and MONSTER~\cite{monster-archive} archives for time series classification~\cite{ismail2023enhancing,ismail2024finding,badi2024cocalite,lee2024leveraging} and clustering~\cite{ismail2023shapedba,holder2024review} as well as the Monash~\cite{tan2020monash} archive for time series extrinsic regression~\cite{bagnall2024hands}. 
In natural language processing, benchmarks like GLUE~\cite{glue-archive}, SuperGLUE~\cite{superglue-archive}, and SQuAD~\cite{squad-archive} have served a similar role. 
Despite these advancements in other areas, human rehabilitation assessment still lacks such a unifying benchmark framework. 
This absence hampers progress by making it difficult to evaluate models consistently, reproduce results, or compare approaches objectively. 
Addressing this gap is essential for accelerating innovation and building reliable, generalizable tools in rehabilitation research.

% ...

First, many proposed deep learning approaches have been evaluated on a limited number of datasets, most notably KIMORE~\cite{capecci2019kimore} and UI-PRMD~\cite{vakanski2018data}. 
While these datasets have been very important in advancing the field, the reliance on a small set of benchmarks raises concerns about the generalizability of the methods. 
It remains unclear how well these approaches perform across diverse populations, exercise types, or real-world settings.  

Second, even when the same datasets are used for evaluation, inconsistencies in experimental protocols make it difficult to ensure fair comparisons. 
Differences in data splits, pre-processing steps, or evaluation metrics are often not clearly reported, leading to ambiguity in reported performance and impeding the ability to draw reliable conclusions.  

Finally, reproducibility remains a significant challenge in this domain. 
The code and implementation details of many proposed methods are not fully released, which complicates independent validation of results and slows the adoption of effective techniques in practical applications. 
Without open and transparent practices, it is difficult to establish a solid foundation for future research.  

To address these limitations, this work makes the following key contributions:
\begin{itemize}
    \item \textbf{Aggregation of datasets}: We aggregate existing rehabilitation motion repositories into a single, dedicated archive containing $60$ datasets. 
    This extended and diverse dataset is designed to support more robust analysis of existing and future methods.
    \item \textbf{Unified framework}: We propose a general and unified framework to analyze and compare deep learning models for rehabilitation motion assessment, ensuring consistent evaluation across methods.
    \item \textbf{Benchmarking existing approaches}: We benchmark a variety of existing approaches, exploring multiple neural network architectures and focusing on two primary tasks in rehabilitation assessment: classification ($39$ datasets) and extrinsic regression ($21$ datasets). 
    \item \textbf{Open access for reproducibility}: To promote transparency, reproducibility, and further research, we publicly release the aggregated dataset as well as all re-implemented codes. 
    This ensures that our work can be reused, extended, and verified by the research community.
\end{itemize}

By addressing these challenges, this paper aims to advance the state-of-the-art in rehabilitation motion assessment, posing a reliable foundation for future research and applications in personalized and scalable rehabilitation solutions.

The remainder of this paper is organized as follows: Section~\ref{rehabil-pile} introduces the novel archive aggregated from existing datasets; In Section~\ref{dl4rehab}, the selected Deep Learning architectures are individually presented in detail; In Section~\ref{experimental-setup}, we describe our experimental protocol and our resulting open-source framework; Section~\ref{experimental-results} presents experimental results and discusses the findings; Finally, Section~\ref{conclusion} concludes the paper with key insights and directions for future work.

\section{Definitions}\label{sec:definitions}

\begin{itemize}
    \item \textbf{A univariate time series} $\textbf{x}=\{x_1,x_2,\ldots,x_L\}$ is a sequence of $L$ data points with a specific ordering, recorded at different time stamps.
    
    \item \textbf{A multivariate time series} $\textbf{X}=\{\textbf{x}_1,\textbf{x}_2,\ldots,\textbf{x}_M\}$ consists of $M$ univariate time series, each of the same length $L$. Each pair of variables exhibits a correlation, sometimes with a certain ordering among them.
    
    \item \textbf{A time series classification dataset} $\mathcal{D}_{clf}=\{\textbf{X}_i,\textbf{y}_i\}_{i=1}^{N}$ is a collection of $N$ multivariate time series, each containing $M$ variables of length $L$. Each time series is associated with a label $\textbf{y}_i$, which represents a discrete probability distribution over $C$ elements, where $C$ is the total number of possible classes. Each element in $\textbf{y}_i$ indicates the probability that $\textbf{X}_i$ belongs to one of the possible classes $\{c_1, c_2, \ldots,c_C\}$.
    
    \item \textbf{A time series extrinsic regression dataset} $\mathcal{D}_{reg}=\{\textbf{X}_i,y_i\}$ is a collection of $N$ multivariate time series, each containing $M$ variables of length $L$. Each time series is associated with a continuous label $y_i$.

    \item \textbf{A skeleton-based human motion sequence} $\textbf{S}$ is a sequence of $L$ recorded frames, where each frame is a $D$-dimensional skeleton of $J$ human joints. Such sequence could be represented as a multivariate time series of $J.D$ variables of length $L$ each.
\end{itemize}

% Each dataset $E$ can thus be represented as:
% \begin{equation}
%  E = \{S_i, y_i\}_{i=1}^N,
% \end{equation}

% \noindent where $N$ corresponds to the number of skeleton sequences $S_i$ in the dataset, each associated with a corresponding label $y_i$. In the case of a classification dataset $E_{clf}$, $y_i$ is a discrete label representing the class of the sequence. In the case of an extrinsic regression dataset $E_{reg}$, $y_i$ is a continuous label representing a score. The sequences $S_i$ are of dimensions $(T,J,D)$ where $T$ represents the length of the sequence, $J$ the number of skeleton joints, and $D$ the number of coordinates characterizing each joint.

\section{The Rehab Pile}
\label{rehabil-pile}

% \mdc{TODO: link with figures for each dataset}
% \mdc{Tables (2) summarizing all } 

% \hfc{need to find a way to separate between "dataset" and "exercise dataset", might be confusing for a reader. Will think of something and prompt ChatGPT}

We propose the creation of an archive dedicated to the analysis of physical exercises in a medical context, from skeleton sequences. Our archive, Rehab-Pile, is thus derived from 9 publicly available repositories. For each repositories, the associated motion analysis task is either binary or multiclass classification, extrinsic regression, or both. Further, since the analysis is carried out per exercise, we consider each exercise as a separate dataset. Thus, our Rehab-Pile archive contains a total of $60$ datasets: $39$ for a classification task and $21$ for an extrinsic regression task.

In what follows, we refer to the original datasets that give rise to our own in the \textit{Rehab-Pile} archive as ``repositories.'' Each \textit{repository} (i.e., original dataset) contains multiple exercises. We treat each exercise as an independent task, since assessing multiple exercises with a single model is not feasible.
Throughout the remainder of this work, each exercise is referred to as a ``dataset.'' In other words, each ``repository'' gives rise to multiple ``datasets,'' where each dataset corresponds to a specific exercise within the original repository.

% \subsection{Datasets}
% \mdc{I propose to have a general notation by considering the number of channels $C$ (should we have something about joints and coordinate per joint ?), and the number of frames $T$. To have something in the form $(C,T)$ and for each dataset description we describe the value of $C$ and $T$.}
% \hfc{Yes i agree, maybe we can emphasize that $C=J.D$ and $J$ is number of joints and $D$ is the number of dimensions.}

\subsection{Regression}

Regression datasets are introduced below and summarized in Table~\ref{tab:reg-datasets}.

\subsubsection{EHE\_REG}

The \textit{Elderly Home Exercise (EHE)}~\cite{bruce2021skeleton} repository includes six exercises performed daily by people in a nursing home, under the guidance of caregivers.
The movements of $25$ people were captured using a Kinect camera.
Each sequence is associated with a label corresponding to the severity level of Alzheimer's disease.
In this work, we consider the problem of estimating disease severity as an extrinsic regression task.
% We follow~\cite{bruce2021skeleton} and employ a 5-fold between-subjects validation protocol, using exactly the same folds as in the original work.
Our Rehab-Pile archive therefore contains six datasets from the original EHE repository.
The distribution of the label score values for the six EHE datasets is presented in Figure~\ref{fig:ehe-labels} of~\ref{apx:label-distribution}.

\subsubsection{KIMORE\_REG}

The original repository, \textit{KInematic assessment of MOvement for remote monitoring of physical REhabilitation (KIMORE)}, presented in~\cite{capecci2019kimore} comprises $71$ patients, including $40$ healthy patients and $31$ unhealthy subjects (patients) with a pathology performing five different exercises.
% Each patient performs one of the five proposed exercises several times.
For each exercise, the patient performed it multiple times during a single recording session.
An expert evaluates their performance and assigns them a continuous score between $0$ and $100$.
The movement sequences are captured using a Kinect camera.
% and their length is resized to the average duration for each exercise separately
Thus, we constructed a total of five extrinsic regression datasets from the KIMORE repository for our Rehab-Pile archive.
The distribution of the label score values for the five KIMORE datasets is presented in Figure~\ref{fig:kimore-reg-labels} of~\ref{apx:label-distribution}.

\subsubsection{UI-PRMD\_REG}\label{sec:uiprmd-reg}

The \textit{University of Idaho-Physical Rehabilitation Movement Data (UI-PRMD)} repository, introduced in~\cite{vakanski2018data}, comprises recordings of $10$ healthy subjects performing physical exercises, correctly and incorrectly, captured using a Kinect camera as well.
We consider the segmented version of the repository, where each skeleton sequence corresponds to a realization of one of the $10$ exercises.
It is important to note that not all exercises are performed by all the $10$ subjects.
Each sequence is associated with a continuous label, allowing us to consider an extrinsic regression task.
Thus, we constructed a total of $10$ datasets from the UI-PRMD repository dedicated to extrinsic regression for our Rehab-Pile archive.
The distribution of the label score values for the $10$ UI-PRMD datasets is presented in Figure~\ref{fig:uiprmd-reg-labels} of~\ref{apx:label-distribution}.

\begin{table}[!h]
    \tiny
    \centering
    \begin{tabular}{cccccc}
        \hline
        \textbf{Exercise} & \textbf{$\#$ samples} & \textbf{$\#$ frames} & \textbf{$\#$ joints} & \textbf{$\#$ dimensions} & \textbf{Label-Range} \\
        \hline
        \multicolumn{6}{c}{\textbf{EHE\_REG}} \\
        \hline
        \textbf{BWL}      & 196 & 42 & 25 & 8 & [0.0 , 10.0] \\
        \textbf{WB}      & 71 & 86 & 25 & 8 & [0.0 , 10.0] \\
        \textbf{BWR}      & 186 & 34 & 25 & 8 & [0.0 , 10.0] \\
        \textbf{WF}      & 74 & 80 & 25 & 8 & [0.0 , 10.0] \\
        \textbf{HUD}      & 144 & 75 & 25 & 8 & [0.0 , 10.0] \\
        \textbf{WH}      & 198 & 49 & 25 & 8 & [0.0 , 10.0] \\
        \hline
        \multicolumn{6}{c}{\textbf{KIMORE\_REG}} \\
        \hline
        \textbf{Sq}      & 71  & 557  & 18 & 3 & [0.0 , 100.0] \\
        \textbf{PR}       & 71  & 847  & 18 & 3 & [0.0 , 100.0] \\
        \textbf{LT} & 71  & 798  & 18 & 3 & [0.0 , 100.0] \\
        \textbf{LA}      & 71  & 725  & 18 & 3 & [0.0 , 100.0] \\
        \textbf{TR}     & 71  & 813  & 18 & 3 & [0.0 , 100.0] \\
        \hline
        \multicolumn{6}{c}{\textbf{UI-PRMD\_REG}} \\
        \hline
        \textbf{SL}     & 140  & 85  & 22 & 3 & [0.0 , 1.0] \\
        \textbf{SSE}    & 126  & 67  & 22 & 3 & [0.0 , 1.0] \\
        \textbf{DS}     & 180  & 81  & 22 & 3 & [0.0 , 1.0] \\
        \textbf{SASLR}  & 146  & 63  & 22 & 3 & [0.0 , 1.0] \\
        \textbf{SSA}    & 126  & 74  & 22 & 3 & [0.0 , 1.0] \\
        \textbf{SSS}    & 108  & 66  & 22 & 3 & [0.0 , 1.0] \\
        \textbf{HS}     & 110  & 69  & 22 & 3 & [0.0 , 1.0] \\
        \textbf{SSIER}  & 120  & 74  & 22 & 3 & [0.0 , 1.0] \\
        \textbf{IL}     & 102  & 77  & 22 & 3 & [0.0 , 1.0] \\
        \textbf{STS}    & 168  & 88  & 22 & 3 & [0.0 , 1.0] \\
        \hline
    \end{tabular}
    \caption{Details for the 21 extrinsic regression datasets.}
    \label{tab:reg-datasets}
\end{table}

\subsection{Classification}

The datasets we introduce for classification are reported below and summarized in Table~\ref{tab:clf-datasets}.

\subsubsection{IRDS\_CLF\_BN}

The \textit{IntelliRehabDS (IRDS)}~\cite{miron2021intellirehabds} repository contains $9$ exercises, acquired using a Kinect sensor and performed by $14$ healthy subjects and $15$ unhealthy subjects.
Some subjects perform the exercises in a standing position, while others perform them sitting on a chair, in a wheelchair, or using a support.
Each sequence is associated with a binary label indicating whether the exercise is correctly performed or not. 
Our Rehab-Pile archive thus contains $9$ datasets targeting a binary classification task.
The distribution of the class labels for the $9$ IRDS datasets is presented in Figure~\ref{fig:irds-labels} of~\ref{apx:label-distribution}.

\subsubsection{KERAAL\_CLF\_BN and KERAAL\_CLF\_MC}

The KERAAL repository~\cite{Nguyen2024IJCNN} was recorded as part of a double-blind long-term rehabilitation clinical study~\cite{Blanchard2022BRI} involving $31$ patients with low back pain, aged $18$ to $70$ years.
The repository includes recordings of healthy subjects and $12$ rehabilitation patients (unhealthy subjects) performing each of three predefined exercises.
The repository includes videos and skeletal data collected from a Microsoft Kinect 2 and a Vicon.
The patient recordings were annotated by a doctor.
In particular, we follow~\cite{Nguyen2024IJCNN} and consider two classification problems, binary and multi-class, related to challenges~1 and ~2, respectively.
For the evaluation, we keep the same logic by including all healthy subjects (groups 2A and 3) in the training set. For patient sequences, we follow a {leave-one-subject-out} protocol.
Thus, our Rehab-Pile archive includes $6$ datasets from the KERAAL repository, three for binary and three for multi-class classification.
The distribution of the class labels for the three KERAAL datasets is presented in Figure~\ref{fig:keraal-labels} of~\ref{apx:label-distribution} for both binary and multi-class cases.

\subsubsection{KIMORE\_CLF\_BN}

In addition to the clinical score prediction task (extrinsic regression) associated with this repository, we also consider a binary classification problem as proposed by Ismail-Fawaz et al. 2025~\cite{litemv} to evaluate the success level of each exercise of the five KIMORE\_REG datasets.
All sequences associated with a score less than or equal to $50$ are labeled by $0$ (incorrect) and all sequences associated with a score greater than $50$ are labeled by $1$ (correct).
Thus, considering each exercise separately, this results in five binary classification datasets in our Rehab-Pile archive.
The distribution of the class labels for the five KIMORE datasets is presented in Figure~\ref{fig:kimore-clf-labels} of~\ref{apx:label-distribution}.

%\mdc{todo: adapt for separate classification and regression, possibly by first having regression datasets}\hfc{I agree, given the original dataset is regression, we would put it in regression no? and mention the classification version there ?}

\subsubsection{KINECAL\_CLF\_BN}

Aiming to assess balance disorders, the KINECAL repository~\cite{maudsley2023kinecal} contains movement sequences of $90$ participants performing $11$ different exercises, recorded using a Kinect sensor.
For each recording, clinical labels are provided, indicating whether the participant is considered clinically at risk.
We use this annotation and thus consider a binary classification problem.
However, among the $90$ participants, only $6$ are considered clinically at risk.
To allow for a fair assessment, we only retain exercises performed by at least five participants labeled as clinically at risk.
Therefore, only four exercises are retained in our archive.
The distribution of the class labels for the four KINECAL datasets is presented in Figure~\ref{fig:kinecal-labels} of~\ref{apx:label-distribution}.

\subsubsection{SPHERE\_CLF\_BN}

% \hfc{TODO: talk about the classification task ? instead of the folds}
The oldest repository in our archive, \textit{Sensor Platform for Healthcare in a Residential Environment (SPHERE) Walking Up Stair (WUS)}~\cite{paiement2014online} contains sequences recorded by a Kinect sensor of $6$ participants walking up stairs.
% The original objective of this dataset is anomaly detection.
% Thus, the original training set contains only normal sequences.
% To remain consistent with the other datasets in our archive, we reformulate the problem as binary classification.
% We keep the original training set, but reorganize the test set to match a \textit{leave-one-subject-out} protocol.
% For each fold, the sequences of 5 subjects are concatenated to the original training set.
% The sequences of the remaining subject constitute the test set.
The original goal of this repository was anomaly detection, specifically, identifying whether the exercise was performed correctly and locating mistakes within a sequence.
We reformulate this task as a binary classification problem, where the objective is to determine whether a given sequence contains any anomaly.
Thus, only a single dataset here is considered in our Rehab-Pile archive.
The distribution of the class labels for the SPHERE WUS dataset is presented in Figure~\ref{fig:sphere-labels} of~\ref{apx:label-distribution}.

\subsubsection{UCDHE\_CLF\_BN and UCDHE\_CLF\_MC}

Proposed in~\cite{singh2023examination,mp-2021,mp-2022}, the repository \textit{University College Dublin Human Exercises (UCDHE)} includes two exercises performed by $52$ and $54$ healthy participants respectively.
Movement data are captured from videos using the OpenPose method~\cite{Cao2019ITPAMI} providing the two-dimensional positions of $8$ joints.
Each sequence is annotated with a discrete label indicating whether the exercise is correctly performed or the type of error, if any.
We consider two variants of this dataset targeting the original multi-class classification task, but also a binary classification task, where we merge the error types into a single class label (incorrect).
Thus, $4$ datasets are considered in our Rehab-Pile archive.
The distribution of the class labels for the two UCDHE datasets is presented in Figure~\ref{fig:ucdhe-labels} of~\ref{apx:label-distribution} for both binary and multi-class cases.

\subsubsection{UI-PRMD\_CLF\_BN}

The original work~\cite{vakanski2018data} also provides for each sequence one binary label indicating if the exercises are correctly performed or not. 
It then allows us to consider the task of binary classification, resulting in additional $10$ datasets dedicated to binary classification, the same number of datasets in the UI-PRMD regression repository (see Section~\ref{sec:uiprmd-reg}).
The distribution of the class labels for the UI-PRMD dataset is presented in Figure~\ref{fig:uiprmd-clf-labels} of~\ref{apx:label-distribution}.

\begin{table}
    \tiny
    \centering
    \begin{tabular}{cccccc}
        \hline
        \textbf{Exercise} & \textbf{$\#$ samples} & \textbf{$\#$ frames} & \textbf{$\#$ joints} & \textbf{$\#$ dimensions} & \textbf{$\#$ classes} \\
        \hline
        \multicolumn{6}{c}{\textbf{IRDS\_CLF\_BN}} \\
        \hline
        \textbf{SAL}      & 269 & 82 & 25 & 3 & 2 \\
        \textbf{SFL}      & 369 & 111 & 25 & 3 & 2 \\
        \textbf{SFE}      & 252 & 95 & 25 & 3 & 2 \\
        \textbf{EFL}      & 249 & 60 & 25 & 3 & 2 \\
        \textbf{STL}      & 264 & 66 & 25 & 3 & 2 \\
        \textbf{SAR}      & 251 & 76 & 25 & 3 & 2 \\
        \textbf{SFR}      & 310 & 106 & 25 & 3 & 2 \\
        \textbf{EFR}      & 271 & 75 & 25 & 3 & 2 \\
        \textbf{STR}      & 281 & 66 & 25 & 3 & 2 \\
        \hline
        \multicolumn{6}{c}{\rule{0pt}{1em} \textbf{KERAAL\_CLF\_MC}} \\
        \hline
        \textbf{ELK}       & 257  & 295  & 11 & 7 & 4 \\
        \textbf{RTK}     & 270  & 264  & 11 & 7 & 4 \\
        \textbf{CTK}     & 285  & 276  & 11 & 7 & 4 \\
        \hline
        \multicolumn{6}{c}{\textbf{KERAAL\_CLF\_BN}} \\
        \hline
        \textbf{ELK}   & 257  & 295  & 11 & 7 & 2 \\
        \textbf{RTK}   & 270  & 264  & 11 & 7 & 2 \\
        \textbf{CTK}   & 285  & 276  & 11 & 7 & 2 \\
        \hline
        \multicolumn{6}{c}{\textbf{KIMORE\_CLF\_BN}} \\
        \hline
        \textbf{Sq}      & 71  & 557  & 18 & 3 & 2 \\
        \textbf{PR}       & 71  & 847  & 18 & 3 & 2 \\
        \textbf{LT} & 71  & 798  & 18 & 3 & 2 \\
        \textbf{LA}      & 71  & 725  & 18 & 3 & 2 \\
        \textbf{TR}     & 71  & 813  & 18 & 3 & 2 \\
        \hline
        \multicolumn{6}{c}{\textbf{KINECAL\_CLF\_BN}} \\
        \hline
        \textbf{GGFV}      & 62  & 296 & 25 & 3 & 2 \\
        \textbf{3WFV}     & 60  & 257 & 25 & 3 & 2 \\
        \textbf{QSEO}     & 85  & 593  & 25 & 3 & 2 \\
        \textbf{QSEC}     & 87  & 592  & 25 & 3 & 2 \\
        \hline
        \multicolumn{6}{c}{\textbf{SPHERE\_CLF\_BN}} \\
        \hline
        \textbf{WUS}      & 48 & 214  & 15 & 3 & 2 \\
        \hline
        \multicolumn{6}{c}{\textbf{UCDHE\_CLF\_MC}} \\
        \hline
        \textbf{Rowing}     & 2432  & 161 & 8 & 2 & 5 \\
        \textbf{MP}     & 1857  & 161  & 8 & 2 & 4 \\
        \hline
        \multicolumn{6}{c}{\textbf{UCDHE\_CLF\_BN}} \\
        \hline
        \textbf{Rowing}     & 2432  & 161 & 8 & 2 & 2 \\
        \textbf{MP}     & 1857  & 161  & 8 & 2 & 2 \\
        \hline
        \multicolumn{6}{c}{\textbf{UI-PRMD\_CLF\_BN}} \\
        \hline
        \textbf{SL}     & 140  & 85  & 22 & 3 & 2 \\
        \textbf{SSE}    & 126  & 67  & 22 & 3 & 2 \\
        \textbf{DS}     & 180  & 81  & 22 & 3 & 2 \\
        \textbf{SASLR}  & 146  & 63  & 22 & 3 & 2 \\
        \textbf{SSA}    & 126  & 74  & 22 & 3 & 2 \\
        \textbf{SSS}    & 108  & 66  & 22 & 3 & 2 \\
        \textbf{HS}     & 110  & 69  & 22 & 3 & 2 \\
        \textbf{SSIER}  & 120  & 74  & 22 & 3 & 2 \\
        \textbf{IL}     & 102  & 77  & 22 & 3 & 2 \\
        \textbf{STS}    & 168  & 88  & 22 & 3 & 2 \\
        \hline
    \end{tabular}
    \caption{Details for the 39 classification datasets.}
    \label{tab:clf-datasets}
\end{table}

All of the above mentioned datasets detailed in Tables~\ref{tab:reg-datasets} and~\ref{tab:clf-datasets} are publicly available with a detailed step by step instructions on how they can be downloaded.
More information are presented in Section~\ref{sec:impl-det}.

\section{Deep Learning for Rehabilitation Assessment}
\label{dl4rehab}

In this section, we detail the deep learning architectures used in this study.
Some of these architectures were originally proposed for Time Series Classification (TSC) in both univariate and multivariate cases, while others were specifically designed for human motion datasets, such as those used in activity recognition and rehabilitation assessment.

The architectures range from convolution-based and recurrent-based models to self-attention-based and graph-convolution-based approaches.
Below, we briefly describe how each backbone processes temporal data and introduce key definitions necessary for their mathematical formulation.

\subsection{Background}

\subsubsection{Convolution Operation on Temporal Data}

\begin{figure}[!ht]
    \centering
    \includegraphics[width=0.8\linewidth]{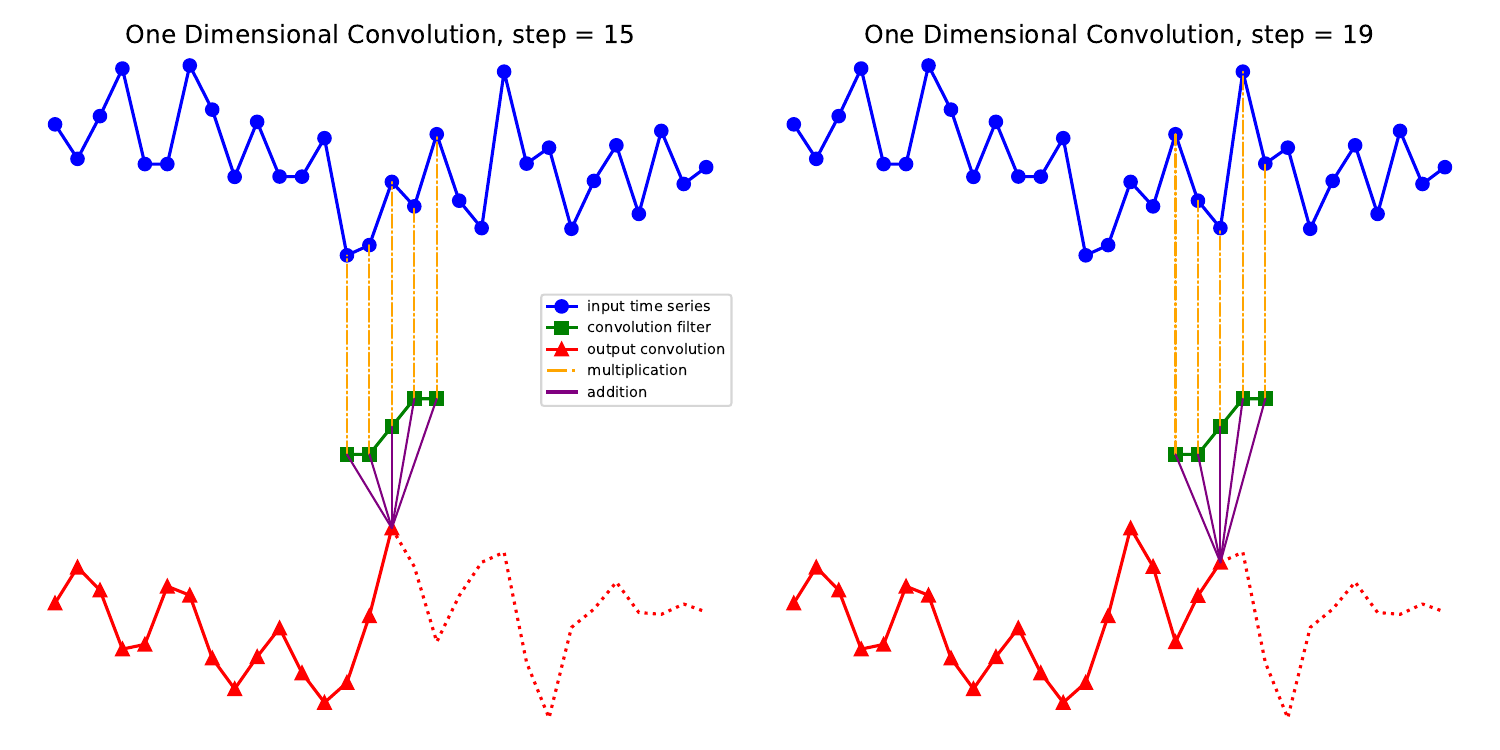}
    \caption{One-dimensional convolution operation on an input sequence.}
    \label{fig:conv-example}
\end{figure}

\begin{figure}[!ht]
    \centering
    \includegraphics[width=0.8\linewidth]{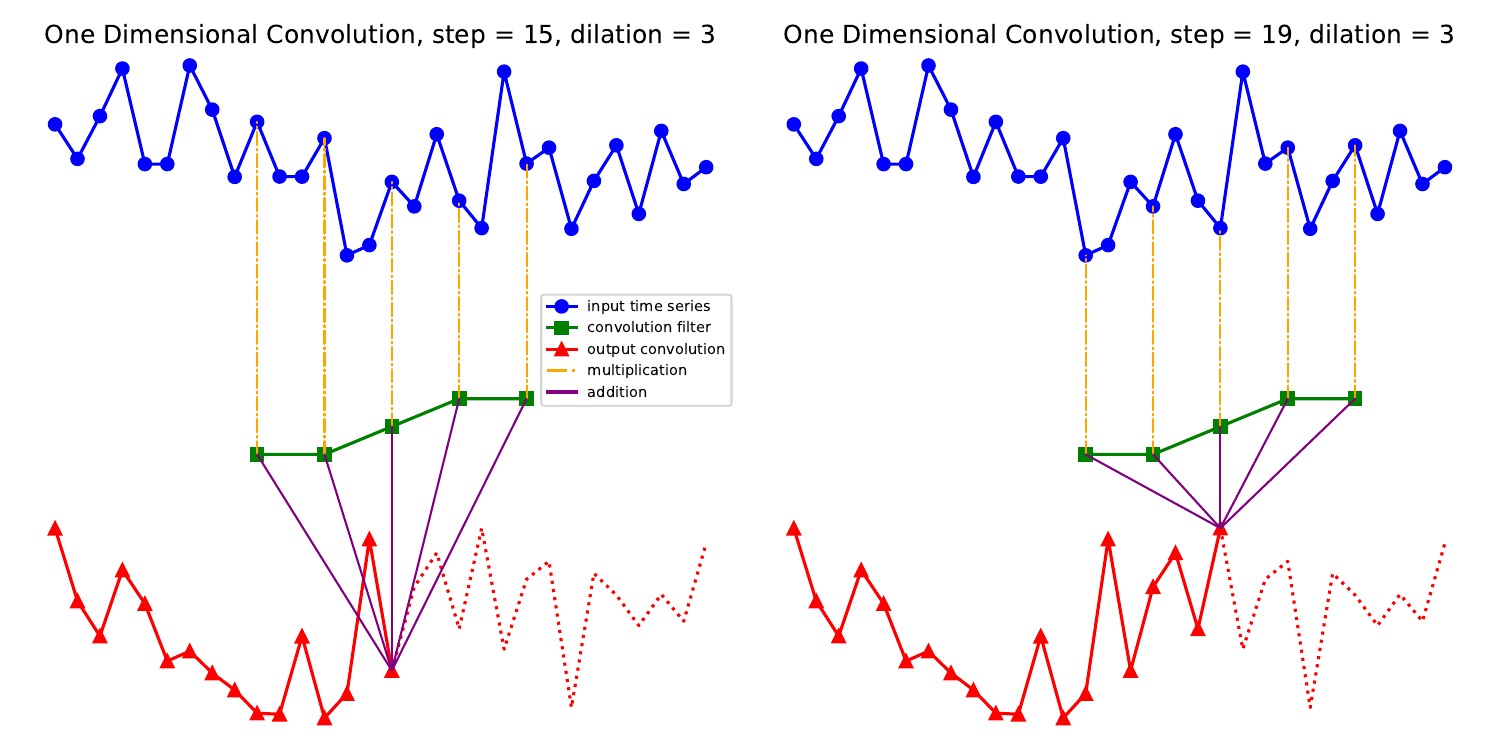}
    \caption{One-dimensional convolution operation with dilation rate set to $3$ (skipping $2$ frames per slide) on an input sequence.}
    \label{fig:dilated-conv-example}
\end{figure}

Originally proposed for image classification~\cite{lecun1989handwritten} in two dimensions, some work started adapting the convolution operation to work on one dimensional data such as time series data~\cite{mcdcnn}.
Given a one dimensional convolutional filter $\textbf{w}=\{w_1,w_2,\ldots,w_k\}$ of size $k$ and a dilation rate $d$, applying a convolution operation between an input univariate time series of length $L$ with $\textbf{w}$ is formulated as:
\begin{equation}
    o_i = \sum_{j=1}^k x_{i+(j-1).d}.w_j, \hspace{0.5cm} \forall~i~\in~[1,L-(k+1).d] ,
\end{equation}

\noindent where $\textbf{o}$ is the output convolution series of length $L-(k+1).d$.
The dilation rate controls the view of the filter on the input time series.
Two examples of a one-dimensional convolution operation, with and without dilation, on a univariate synthetic time series are presented in Figures~\ref{fig:conv-example} and~\ref{fig:dilated-conv-example}.
The goal of the convolution operation is to detect patterns that help differentiate input data across different categories.
This is particularly useful for classification tasks.

\subsubsection{Recurrent Neural Networks}

Recurrent Neural Networks (RNNs)~\cite{rnns} are a type of neural network designed to handle sequential data by maintaining a form of memory.
This memory lets the model retain information from earlier in the sequence and use it to inform future processing.
RNNs are commonly used in auto-regressive tasks such as text completion or forecasting, but they also perform well in discriminative tasks like classification or regression.
At each step in a sequence, the model looks at the current input, combines it with its memory of past inputs, generates an embedded representation, and updates the memory for the next step.
Among various RNN types, this work focuses on Long Short-Term Memory (LSTM) networks~\cite{lstms} and Gated Recurrent Unit (GRU) networks~\cite{grus}.

The LSTM network utilizes different gates to control how the information flows from a frame to another.
At each frame, it updates a hidden state $h_t$ and a separate cell $c_t$, both acting as short-long memories. 
For an input UTS $\textbf{x}$ of length $L$, the output sequence $\textbf{o}$ of the LSTM layer is computed at frame $t$ as follows:
\begin{equation}
\begin{aligned}
f_t &= \sigma(W_f x_t + U_f h_{t-1} + b_f) && \text{(forget gate)} \\
i_t &= \sigma(W_i x_t + U_i h_{t-1} + b_i) && \text{(input gate)} \\
\tilde{c}_t &= \tanh(W_c x_t + U_c h_{t-1} + b_c) && \text{(candidate cell state)} \\
c_t &= f_t \odot c_{t-1} + i_t \odot \tilde{c}_t && \text{(cell state update)} \\
o_t &= \sigma(W_o x_t + U_o h_{t-1} + b_o) && \text{(output gate)} \\
h_t &= o_t \odot \tanh(c_t) && \text{(hidden state)},
\end{aligned}
\end{equation}

\noindent where $h_t$ is the hidden state; $W_f$, $W_i$, $W_o$ are the forget, input, and output gates' transformation matrices; $W_c$ is the state cell candidate transformation matrix; $b_f$, $b_i$, $b_o$, $b_c$ are the bias terms; $\sigma$ is the sigmoid function and $\odot$ denotes the element-wise matrix multiplication operation.

The GRU network is an improvement and simplified version of LSTM as it combines forget and input gates into an update gate.
For an input UTS $\textbf{x}$, the output sequence $\textbf{o}$ of the GRU layer is computed at frame $t$ as follows:
\begin{equation}
\begin{aligned}
z_t &= \sigma(W_z [x_t, h_{t-1}] + b_z) && \text{(update gate)} \\
r_t &= \sigma(W_r [x_t, h_{t-1}] + b_r) && \text{(reset gate)} \\
\tilde{h}_t &= \tanh(W_h [x_t, r_t \odot h_{t-1}] + b_h) && \text{(candidate hidden state)} \\
h_t &= (1 - z_t) \odot h_{t-1} + z_t \odot \tilde{h}_t && \text{(new hidden state)},
\end{aligned}
\end{equation}

\noindent where $h_t$ is the hidden state; $W_z$, $W_r$ are the update and reset gates' transformation matrices; $W_h$ is the hidden state transformation matrix; $\tilde{h}_t$ is the candidate activation; $b_z$, $b_r$, $b_h$ are the bias terms; $\sigma$ is the sigmoid function; and $\odot$ denotes the element-wise matrix multiplication operation.

\subsubsection{Self-Attention}

Self-Attention has gained significant interest in recent years, particularly due to its role in Transformer architectures for Natural Language Processing (NLP)~\cite{attention-is-all-you-need}. This mechanism enables the model to capture global dependencies by relating different positions in the input sequence.

To compute the self-attention representation of an MTS $\textbf{X}$ of length $L$ and dimensions $M$, we first project at each frame $t$, the vector $\textbf{x}_t$ into three distinct representations:
\begin{equation}
    Q_t = W_Q . \textbf{x}_t, \hspace{1cm} K_t = W_K . \textbf{x}_t, \hspace{1cm} V_t = W_V . \textbf{x}_t,
\end{equation}
where $Q_t, K_t, V_t \in \mathbb{R}^{d_{model}}$ are the Query, Key, and Value vectors for time step $t$, and $W_Q, W_K, W_V \in \mathbb{R}^{M \times d_{model}}$ are learnable projection matrices.

Stacking these representations across all frames yields matrices $Q, K, V \in \mathbb{R}^{L \times d_{model}}$. The attention matrix is then computed as:
\begin{equation}
    \text{Att} = \text{softmax} \left( \frac{Q . K^\top}{\sqrt{d_{model}}} \right) ,
\end{equation}

\noindent where the softmax function is applied row-wise to compute the attention weights between all frames.

Finally, the output sequence $\textbf{o}$ of length $L$ and dimensions $M$ is computed as:
\begin{equation}
    \textbf{o} = \text{Att} . V . W_O,
\end{equation}

\noindent where $W_O \in \mathbb{R}^{d_{model} \times M}$ is an optional output projection matrix that adjusts the dimensionality of the final representation.

\subsubsection{Graph Convolution Neural Networks}
Graph Neural Networks (GNNs)~\cite{gnns} refer to any  neural network that is specifically designed to handle graph-structured data.
A graph $\mathcal{G}$ is defined as $\mathcal{G} = \{ \mathcal{V}, \mathcal{E}, \mathcal{A} \}$, where $\mathcal{V}$ is a set of $N$ nodes, $\mathcal{E}$ is the set of edges, and $\mathcal{A} \in \mathbb{R}^{N \times N}$ is the adjacency matrix encoding node connectivity.

A well-known type of GNN is the Graph Convolutional Network (GCN)~\cite{gcns}.  
Assuming an input sequence of skeleton-based human motion $\mathbf{S} \in \mathbb{R}^{L \times J \times D}$, representing $L$ frames, $J$ joints, and $D$-dimensional features per joint, the output $\mathbf{o}=\{o_1,o_2,\ldots,o_L\}$ of a GCN layer at frame $t$ is calculated as:
\begin{equation}
    \mathbf{o}_t = \text{ReLU}\left( \hat{\mathcal{A}} \cdot \mathbf{F}_t \cdot W \right) .
\end{equation}

\noindent Here, $\mathbf{F}_t \in \mathbb{R}^{N \times d}$ is the feature matrix at time $t$, extracted using a temporal convolution over $\mathbf{S}$; $W \in \mathbb{R}^{d \times d'}$ is a learnable weight matrix; and $\hat{\mathcal{A}}$ is the normalized adjacency matrix defined as:
\begin{equation}
    \hat{\mathcal{A}} = D^{-\frac{1}{2}} \tilde{\mathcal{A}} D^{-\frac{1}{2}} ,
\end{equation}

\noindent where $\tilde{\mathcal{A}} = \mathcal{A} + I_N$ adds self-loops, and $D$ is the degree matrix with entries:
\begin{equation}
    D_{ii} = \sum_{j=1}^{N} \tilde{\mathcal{A}}_{ij} .
\end{equation}

The goal of GCNs is to enable each node to update its representation by aggregating information from its neighbors in the graph. The ReLU activation introduces non-linearity, allowing the model to learn complex patterns from graph-structured data.

\subsection{Deep Learning Architectures}

\begin{figure}[!ht]
    \centering
    \includegraphics[width=0.6\linewidth]{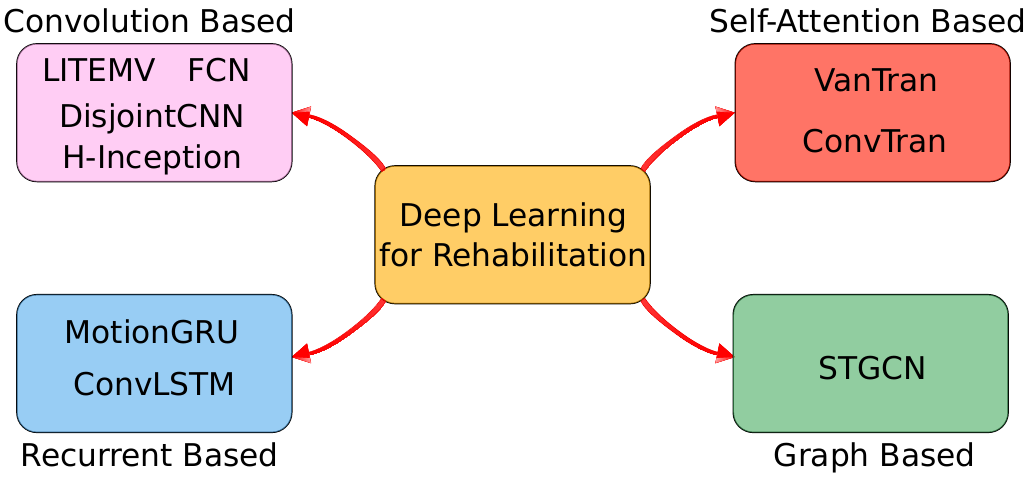}
    \caption{The deep learning architectures used in this study can be divided into four categories: (1) convolution based, (2) self-attention based, (3) recurrent based, and (4) graph based.}
    \label{fig:models-summary}
\end{figure}

In this section we present in details each of the architectures used in our study.
The architectures are divided into four groups, as detailed in Figure~\ref{fig:models-summary}: (1) Convolution based, (2) Recurrent based, (3) Self-Attention based, and (4) Graph based.

\subsubsection{Fully Convolutional Network (FCN)}\label{sec:fcn}

The Fully Convolutional Network (FCN), proposed in~\cite{fcn}, consists of three convolutional blocks.

Each block comprises a one-dimensional convolutional layer, followed by batch normalization and a ReLU activation function. The batch normalization layer, which is utilized by most of the models compared in this study, helps to scale the feature space and facilitates training, leading to faster convergence of the deep learning model.

The output of the final convolutional block undergoes a global average pooling operation, which replaces each feature with its average value over the time axis, while preserving the spatial axis. This output vector is then passed to the downstream task layer, which varies depending on whether the task is classification or extrinsic regression.

For classification tasks, the output layer applies a matrix transformation followed by a softmax activation to predict a probability distribution over the possible classes.

For extrinsic regression tasks, the output layer performs a matrix transformation to a single value in space, predicting a continuous label.

\begin{figure}[!ht]
    \centering    \includegraphics[width=0.8\linewidth]{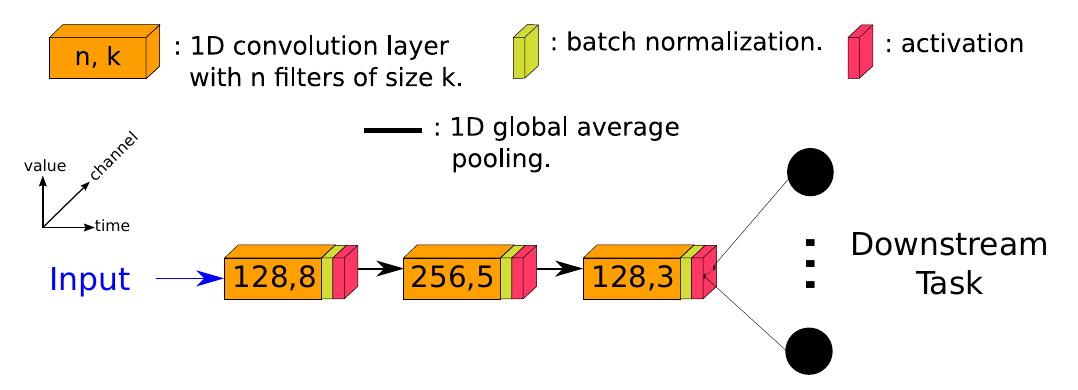}
    \caption{The Fully Convolutional Network (FCN)~\cite{fcn} deep learning architecture, originally proposed for univariate and multivariate Time Series Classification.}
    \label{fig:fcn}
\end{figure}

A detailed view of the FCN model is represented in Figure~\ref{fig:fcn}.

\subsubsection{Hybrid Inception (H-Inception)}\label{sec:hinception}

Initially proposed for image classification, InceptionV4~\cite{inception, inceptionv4} is a convolution-based deep learning model designed to tackle this task efficiently.
It differentiates itself from other deep learning architectures through a multiplexing convolution approach, which applies multiple convolutional layers with different characteristics to the same input, capturing patterns at various scales.

In 2020, Inception was adapted for time series classification (TSC)~\cite{inceptiontime} by replacing two-dimensional convolutions with one-dimensional ones, making it suitable for processing time series data.
The Inception architecture for TSC consists of six modules, each containing three parallel convolutional layers (multiplexing convolution) along with a max pooling operation.
The outputs of these four layers are concatenated and passed through a batch normalization layer followed by an activation function.
Similar to FCN, the final module's output is summarized using a global average pooling operation before being processed for the downstream task.

\begin{figure}[!ht]
    \centering
    \includegraphics[width=\linewidth]{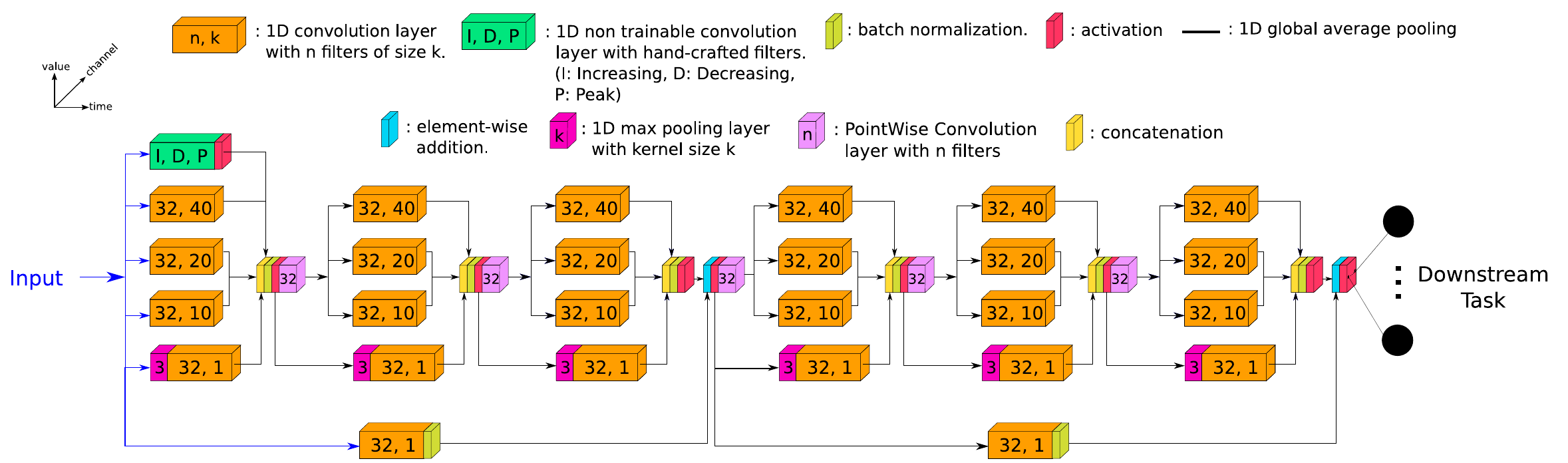}
    \caption{The Hybrid Inception (H-Inception)~\cite{inceptiontime,hinception} deep learning architecture, originally proposed for univariate and multivariate Time Series Classification.}
    \label{fig:hinception}
\end{figure}

In 2022, Ismail-Fawaz et al.~\cite{hinception} introduced a set of hand-crafted, non-trainable convolutional filters designed to detect increasing trends, decreasing trends, and peaks in time series data.
To incorporate these features, the Inception architecture was adapted by integrating these hand-crafted filters in the first layer, ensuring they remain unaffected during training.
This modified architecture, known as Hybrid Inception (H-Inception), is detailed in Figure~\ref{fig:hinception}.

\subsubsection{Light Inception with boosTing tEchniques MultiVariate (LITEMV)}\label{sec:litemv}

\begin{figure}[!ht]
    \centering
    \includegraphics[width=1\linewidth]{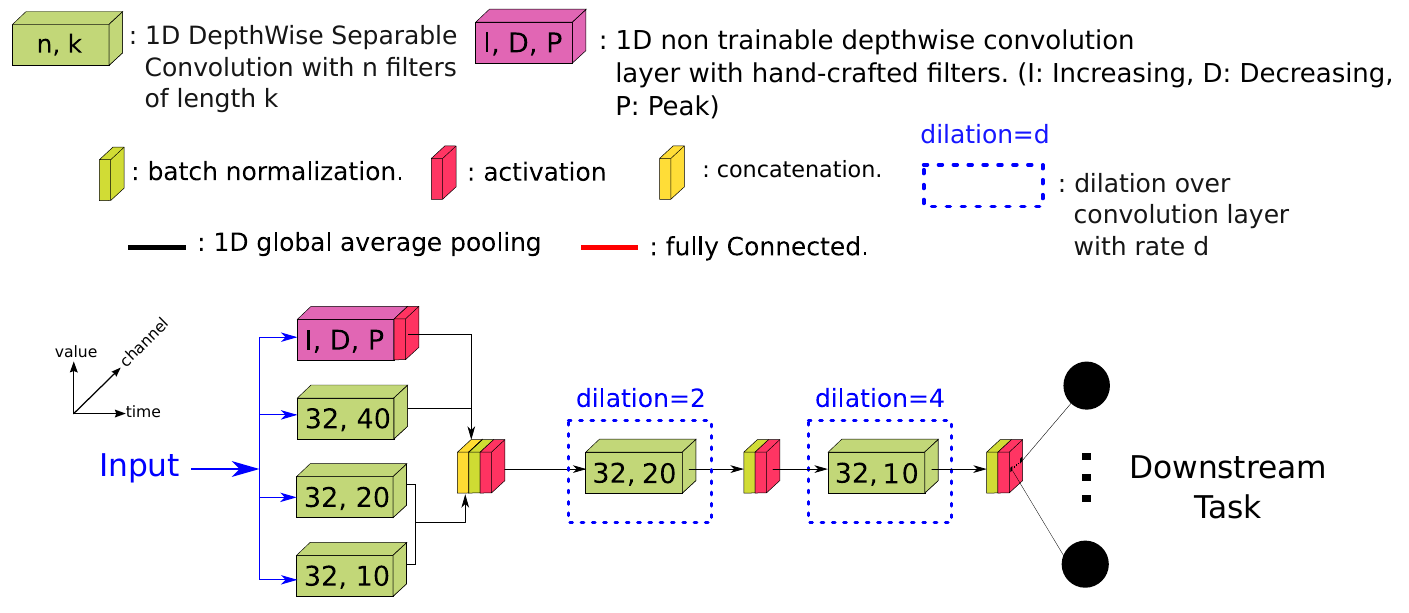}
    \caption{The Light Inception with boosTing tEchniques MultiVariate (LITEMV)~\cite{litemv} deep learning architecture, originally proposed for multivariate Time Series Classification.}
    \label{fig:litemv}
\end{figure}

Given the high number of parameters in models such as H-Inception, Ismail-Fawaz et al.~\cite{lite} proposed a significant reduction in the number of parameters in deep learning models for TSC, while maintaining the performance of state-of-the-art models. 
This was achieved through the introduction of the Light Inception with boosTing tEchniques (LITE) architecture, which leverages a simple three block structure similar to FCN, while incorporating boosting techniques to enhance performance.
These boosting techniques include multiplexing convolution, hand-crafted convolutional filters, and dilated convolutions.
The key to reducing the number of parameters lies in the use of DepthWise Separable Convolutions (DWSCs), which significantly decrease computational complexity without sacrificing accuracy.
The LITE architecture was later adapted to better handle multivariate time series by Ismail-Fawaz et al.~\cite{litemv}, leading to the LITEMV model.
The LITEMV architecture, detailed in Figure~\ref{fig:litemv}, consists of three convolutional blocks.
The first block contains three parallel DWSC layers (multiplexing), along with hand-crafted, non-trainable DepthWise convolutional filters.
The second and third blocks each contains a single DWSC layer, with dilation rates increasing exponentially with depth.
Each convolution block follows the convolution operation by a batch normalization layer and an activation function.
The final block's output undergoes a global average pooling operation before being fed into the downstream task layer. 
This architecture is shown in Figure~\ref{fig:litemv}. 

\subsubsection{Disjoint Convolutional Neural Network (DisjointCNN)}\label{sec:disjointcnn}

\begin{figure}[!ht]
    \centering
    \includegraphics[width=1\linewidth]{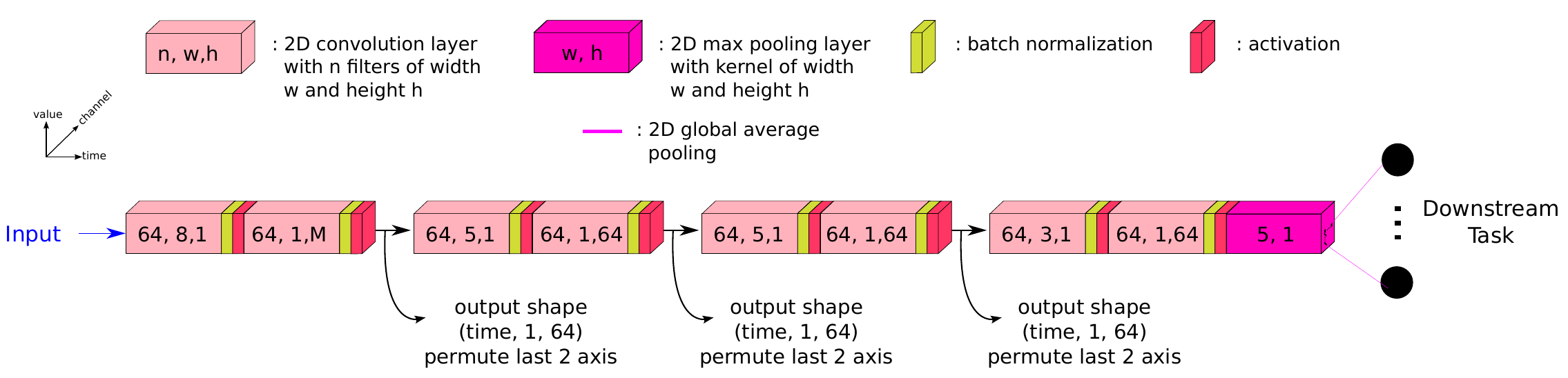}
    \caption{The Disjoint Convolutional Neural Network (DisjointCNN)~\cite{disjointcnn} deep learning architecture, originally proposed for multivariate Time Series Classification.}
    \label{fig:disjointcnn}
\end{figure}

Convolution operations on multivariate time series on a one-dimensional scale still lack the ability to effectively incorporate spatial information.
This issue is addressed in LITEMV using one-dimensional DWSC layers.
However, Foumani et al.~\cite{disjointcnn} proposed a novel approach for performing convolutions on multivariate time series, termed 1+1D convolution.
Similar to DWSC layers, this method consists of two convolution phases.
The first phase applies a two-dimensional convolution layer with the kernel's height set to 1, ensuring that temporal features are extracted without affecting the spatial dimensions.
The second phase employs another two-dimensional convolution layer, this time with the kernel's width set to 1 and the height matching the input dimensions of the series, allowing the model to learn spatial relationships.
In the first phase, 1+1D convolution detects patterns along the temporal axis without interfering with the spatial axis.
The second phase then captures spatial dependencies by learning relationships between spatial dimensions.
Each phase in the 1+1D convolution is followed by batch normalization and an activation function.
The Disjoint Convolutional Neural Network (DisjointCNN), introduced in~\cite{disjointcnn}, incorporates four 1+1D convolution blocks.
The output of the final block undergoes a two-dimensional local max pooling operation with the kernel height set to 1, followed by two-dimensional global average pooling before being passed to the downstream task layer.
A detailed view of this architecture is presented in Figure~\ref{fig:disjointcnn}.

\subsubsection{Convolution Long Short-Term Memory (ConvLSTM)}\label{sec:conv-lstm}

Originally proposed in~\cite{convlstm} for the action recognition task using wearable sensor activity, the Convolutional Long Short-Term Memory (ConvLSTM) network combines two well-known architectures for sequential data processing. The first component is convolution, which detects local patterns and extracts feature maps, while the second is LSTM recurrent layers, which capture global temporal dependencies.

\begin{figure}[!ht]
    \centering
    \includegraphics[width=1\linewidth]{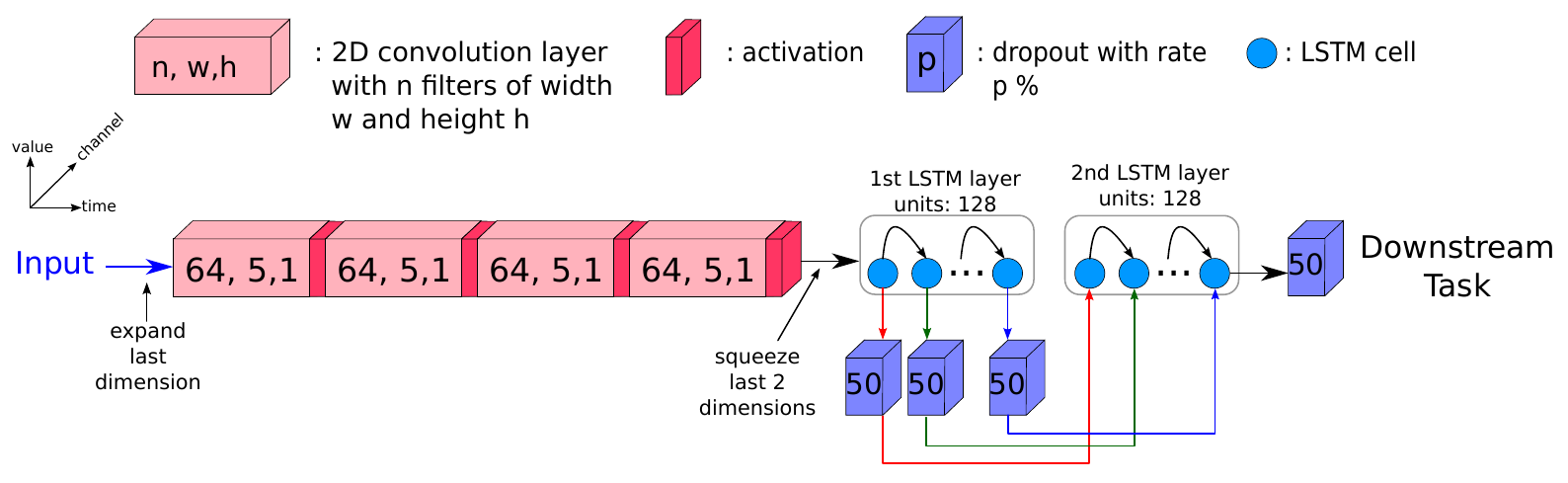}
    \caption{The Convolutional Long Short-Term Memory (ConvLSTM)~\cite{convlstm} deep learning architecture, originally proposed for wearable sensors in activity recognition.}
    \label{fig:conv-lstm}
\end{figure}

The implementation of ConvLSTM in our work is adapted from\footnote{\url{https://github.com/takumiw/Deep-Learning-for-Human-Activity-Recognition/}}. This architecture first applies four convolutional blocks, each consisting of a two-dimensional convolution operation with a kernel width of 1, followed by an activation function. 
The output of the fourth convolutional block is then passed through two LSTM layers and dropout operations before being fed into the final downstream task. More details about the ConvLSTM architecture are presented in Figure~\ref{fig:conv-lstm}.

\subsubsection{Motion Gated Recurrent Unit (MotionGRU)}\label{sec:gru}

The Motion Gated Recurrent Unit model (MotionGRU) was first proposed in~\cite{gru-action-2motion} and used as a feature extractor during the evaluation phase for metric calculation. It has since been applied in various human motion generation research studies~\cite{gru-svae,ismail2025establishing}.

The model utilizes GRU layers on raw input sequences to capture global information and summarize sequence patterns into a single vector. The architecture consists of two GRU layers followed by a fully connected layer with an activation function to reduce dimensionality before being fed into the downstream task layer.

\begin{figure}[!ht]
    \centering
    \includegraphics[width=0.8\linewidth]{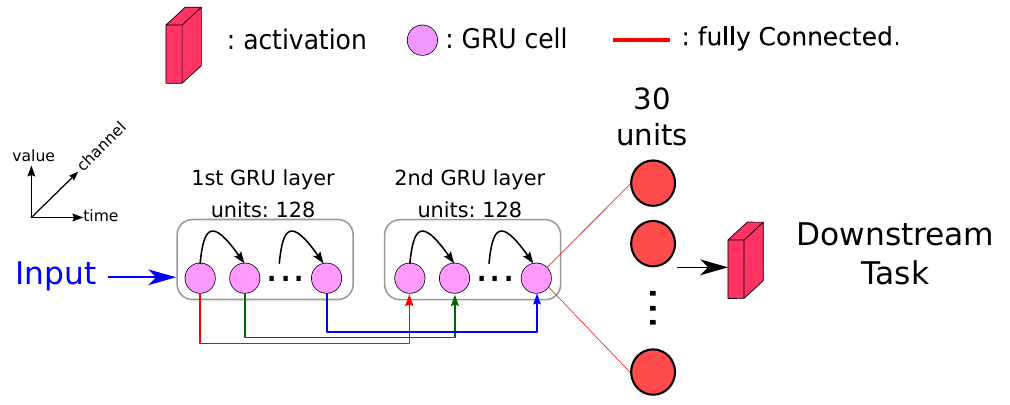}
    \caption{The Motion Gated Recurrent Unit (MotionGRU)~\cite{gru-action-2motion} deep learning architecture, originally proposed for skeleton based sequences in activity recognition.}
    \label{fig:motion-gru}
\end{figure}

More details about the architecture are presented in Figure~\ref{fig:motion-gru}.

\subsubsection{Vanilla Transformer (VanTran)}\label{sec:vantran}

The self-attention mechanism has attracted significant interest from researchers since its introduction in Transformers for text translation~\cite{attention-is-all-you-need}. Similarly, in the domain of human motion sequences, self-attention has emerged as a key component for various downstream tasks.
Specifically, for human motion generation, the Transformer architecture has recently gained considerable attention, particularly with the ACTOR model~\cite{actor}, an action-conditioned 3D generative model for skeleton-based human motion data. 
The ACTOR model~\cite{actor} is built upon a Variational Autoencoder (VAE) architecture.
Its encoder leverages the standard Transformer encoder from~\cite{attention-is-all-you-need}. 
In our work, we retain the encoder architecture and adapt it for classification or regression tasks in downstream applications.

\begin{figure}[!ht]
    \centering
    \includegraphics[width=1\linewidth]{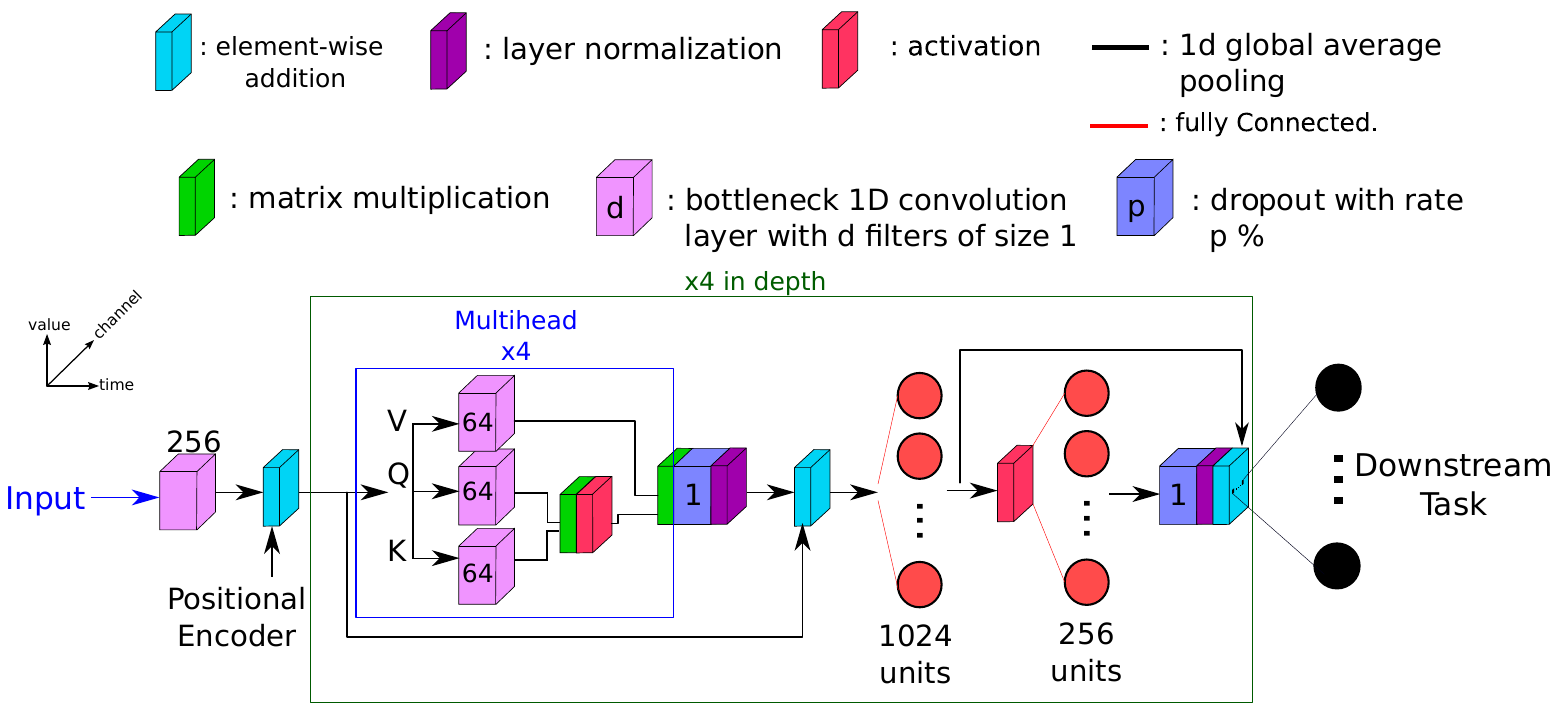}
    \caption{The Vanilla Transformer (VanTran)~\cite{actor} deep learning architecture, originally proposed for generating skeleton based sequences as a full encoder-decoder scheme; we retain the encoder part.}
    \label{fig:van-tran}
\end{figure}

A detailed view of the VanTran architecture is shown in Figure~\ref{fig:van-tran}.
First, the input multivariate sequence is embedded using a bottleneck layer, implemented as a one-dimensional convolution with a unit kernel size.
Next, a standard self-attention encoding mechanism is applied, following the original Transformer model~\cite{attention-is-all-you-need}.
In line with the ACTOR model~\cite{actor}, we use the sinusoidal Absolute Positional Encoder (APE) and a total of four attention blocks.
Each attention block consists of a self-attention mechanism, followed by a residual connection and a Feed-Forward Neural Network (FFNN).
Finally, a Global Average Pooling operation is applied before passing the output to the downstream task.

\subsubsection{Convolution Transformer (ConvTran)}\label{sec:convtran}

\begin{figure}[!ht]
    \centering
    \includegraphics[width=1\linewidth]{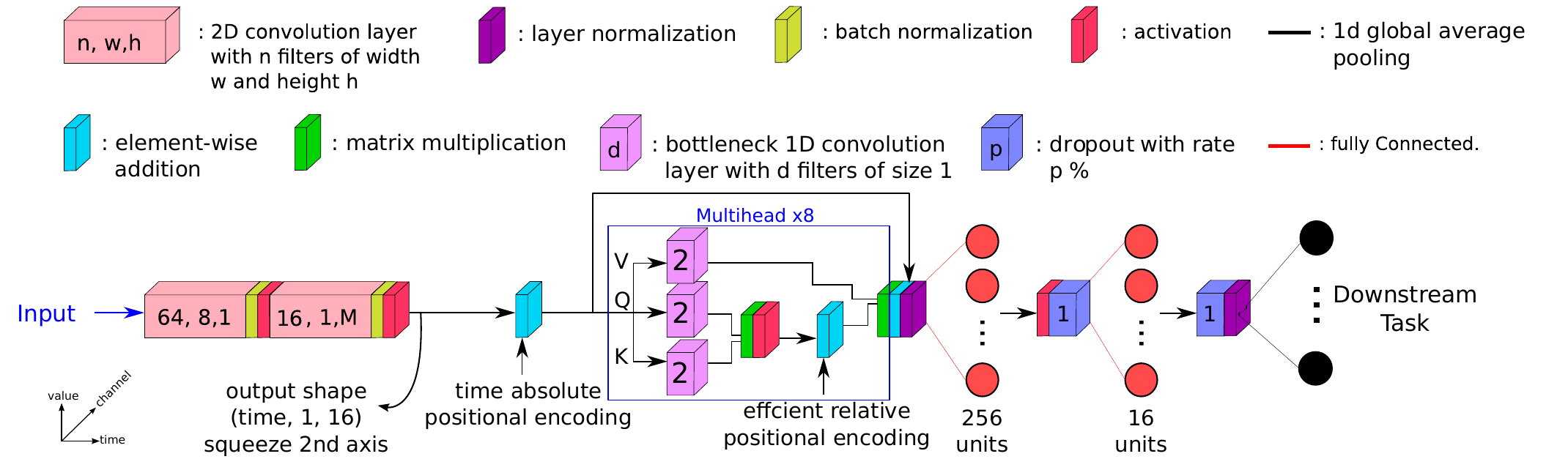}
    \caption{The Convolutional Transformer (ConvTran)~\cite{convtran} deep learning architecture, originally proposed for multivariate Time Series Classification.}
    \label{fig:conv-tran}
\end{figure}

Just as self-attention-based architectures have attracted significant attention in human motion generation, Transformer-like (self-attention-based) architectures have also become a growing focus in Multivariate Time Series Classification.
Recently, Foumani et al.~\cite{convtran} proposed a Convolution-Transformer architecture (ConvTran), illustrated in Figure~\ref{fig:conv-tran}.

The ConvTran architecture leverages the 1+1D convolutional embedding technique introduced in~\cite{disjointcnn} (see DisjointCNN in  Section~\ref{sec:disjointcnn}) followed by a sinusoidal time Absolute Positional Encoder (tAPE).
The difference between tAPE and the traditional sinusoidal APE is its frequency adaptation to the length and dimension of the Multivariate Time Series.
The sequence is then processed through a self-attention mechanism.
However, unlike VanTran (Section~\ref{sec:vantran}), ConvTran utilizes efficient Relative Positional Encoding (eRPE) in order to enhance the self-attention mechanism by encoding positional relationships directly into the attention computation. This architecture includes only a single self-attention block, with no residual connections, followed by a Feed-Forward Neural Network (FFNN) before the output is passed to the downstream task.

\subsubsection{Spatio-Temporal Graph Convolution Network (STGCN)}\label{sec:stgcn}

Graph Convolution Networks (GCNs)~\cite{gcns} have been widely used for temporal data with spatial information that can be represented on a graph, such as traffic data, and especially skeleton-based human motion data.

Originally, a Spatio-Temporal Graph Convolution Network (STGCN) was proposed for traffic forecasting~\cite{stgcn-traffic}.  
The STGCN architecture proposed in~\cite{stgcn-traffic} consists of multiple GCN blocks connected in cascade mode.  
Each of these blocks contains a two-dimensional convolution operation with unit height to capture temporal features without spatial aggregation.  
The output of this convolution layer is then fed to two graph convolution layers, one using the adjacency matrix and the other using the second-order matrix.  
The role of each of these graph convolutions is to capture important correlations between different dimensions in the skeleton and is applied with the same transformation over each frame.  
The output of these two graph convolutions is concatenated and fed to a series of two-dimensional convolution layers with dropout operations.  
The output of the last GCN block goes through an aggregation using a Global Average Pooling (GAP) followed by a Fully Connected layer before being fed to the downstream task.  

Given the success of this architecture in multivariate time series forecasting such as traffic data, the authors in~\cite{stgcn-rehab} proposed to enhance it and use it for skeleton-based rehabilitation tasks.  
The enhancements over this architecture comprise:  
\begin{itemize}
    \item Replacing the feature aggregation GAP by multiple LSTM layers.  
    The replacement of the GAP by LSTM layers helps aggregate temporal information, while capturing recurrent features that are accumulated over the time axis;  
    \item Adding attention mechanism to the graph convolution.  
    Traditionally, the matrix used to perform the graph convolution operation is not trained; it is derived from the adjacency matrix.  
    In order to learn the importance of the spatial information, the authors in~\cite{stgcn-rehab} proposed to multiply the transformation matrix by a learnable attention matrix.  
    This adds an importance weight for each joint in the skeleton, instead of assuming all contribute equally.  
\end{itemize}  

\begin{figure}[!ht]
    \centering
    \includegraphics[width=1\linewidth]{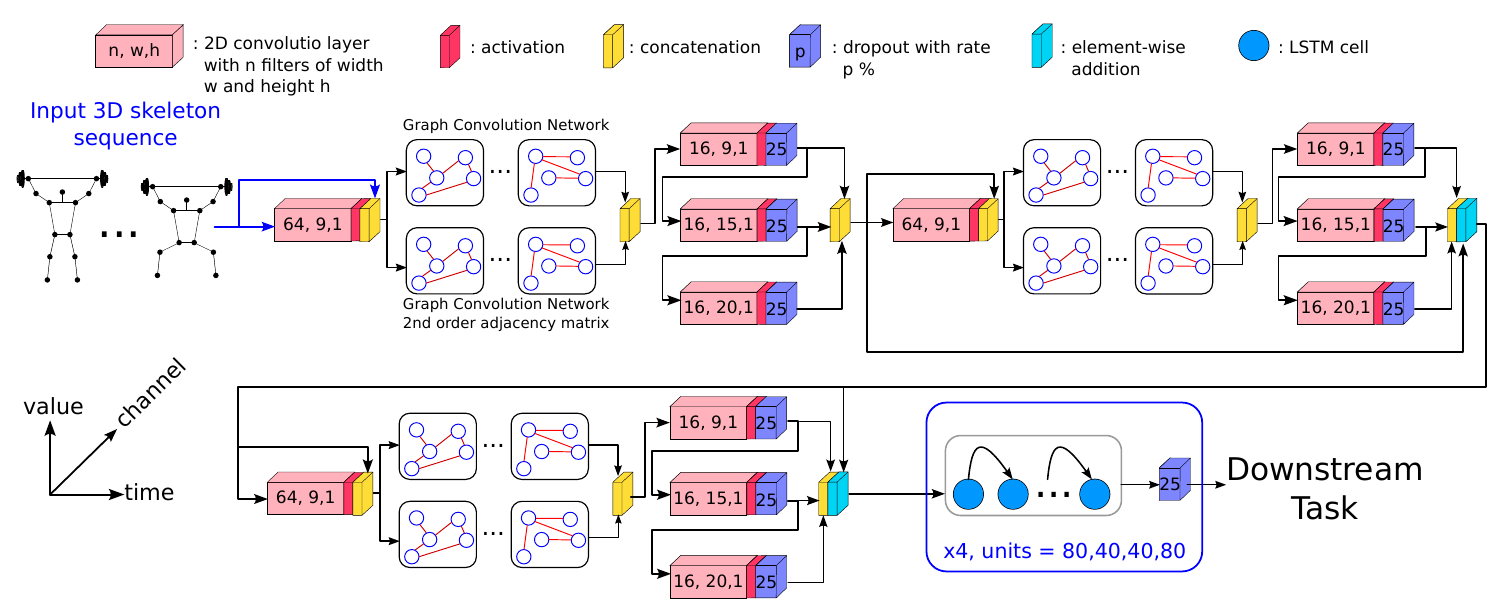}
    \caption{The Spatio-Temporal Graph Convolution Network (STGCN)~\cite{stgcn-rehab} deep learning architecture, originally proposed for skeleton based sequences in rehabilitation assessment.}
    \label{fig:st-gcn}
\end{figure}

A detailed view of the STGCN architecture and its enhanced version is presented in Figure~\ref{fig:st-gcn}.

\paragraph{Downstream Task Layer} All of the architectures mentioned above produce, for each input sample $\textbf{S}$, a vector representation $\textbf{h}$ of size $D_h$ that encodes the relevant features.
We use these representations for both classification and extrinsic regression tasks.

For the \textbf{classification task}, we add a Fully Connected (FC) layer that transforms the input vector into a new space using a learnable matrix transformation.
The output dimension of this FC layer is set to the number of classes $C$ in the dataset.
The resulting logits $\textbf{z}$ are produced by the following matrix transformation operation:
\begin{equation}
    \textbf{z} = W.\textbf{h} ,
\end{equation}

\noindent where $W$ is a learnable matrix of shape $(C,D_h)$.
The logits $\textbf{z}$ are then passed through a softmax activation function to produce a probability distribution over the classes. Each output node represents the probability of the input belonging to a specific class:
\begin{equation}
\text{softmax}(z_c) = \dfrac{e^{z_c}}{\sum_{j=1}^{C} e^{z_j}} ,
\end{equation}

\noindent
where $z_c$ is the logit corresponding to class $c$, and $C$ is the total number of classes in the given dataset.

Similarly, for the \textbf{extrinsic regression task}, we apply an FC layer to the output vector representation $\textbf{h}$.
In this case, the output dimension of the FC layer is set to $1$, as the goal is to predict a single continuous value.
No activation function is applied after the FC layer.

\section{Experimental Setup}
\label{experimental-setup}

\subsection{Data Preparation}

\subsubsection{Data Resampling}

Originally, the datasets provided in their original work contained motion sequences of varying lengths.
In our work, since all samples within each dataset were recorded at a consistent frame rate using the same technology, we resample all sequences in a given dataset to a common number of frames.
Specifically, we resample each sequence to the average sequence length within its dataset.
We use a Fourier-based resampling method provided by \textit{scipy}~\cite{scipy}\footnote{\url{https://docs.scipy.org/doc/scipy/reference/generated/scipy.signal.resample.html}}, which is appropriate given the uniform sampling rate of the data.

\subsubsection{Train-Test Folds Creation}

To mitigate potential bias in our results, we perform multiple train-test folds on each dataset.  
For each fold, we generate separate training and testing sets.  
The train-test split follows a cross-subject protocol, ensuring that no subject appears in both the training and testing sets.  
When applicable, we further ensure that the test set contains only samples from unhealthy subjects.  

We use the following procedure to determine the number of folds:  
\begin{itemize}
    \item If the dataset includes both healthy and unhealthy subjects:  
    \begin{itemize}
        \item If there are at least ten unhealthy subjects, we use 5-fold cross-subject split.  
        \item If there are fewer than ten unhealthy subjects, we adopt a leave-one-subject-out approach, resulting in $n_{unh}$ folds, where $n_{unh}$ is the number of unhealthy subjects in the dataset.  
    \end{itemize}
    \item If the dataset includes only healthy subjects:  
    \begin{itemize}
        \item If there are at least ten subjects, we use 5-fold cross-subject split.  
        \item If there are fewer than ten subjects, we use a leave-one-subject-out approach, resulting in $n$ folds, where $n$ is the total number of subjects in the dataset.  
    \end{itemize}
\end{itemize}

\subsubsection{Data Normalization}

To address potential issues related to feature scaling and to ensure stable model convergence, we apply $\min$-$\max$ normalization to the input data.  
This approach rescales the input features to a uniform range, mitigating the impact of large value magnitudes, while preserving temporal and spatial patterns.  
Normalization is performed independently for each feature dimension across all sequences, frames, and joints.  

Let $\mathcal{S} = \{\textbf{S}^{(i)}\}_{i=1}^{N}$ denote a dataset of $N$ skeleton-based motion sequences, where each sequence has length $L$, $J$ joints, and $D$-dimensional joint features.  
The normalized values are computed as:  
\begin{equation}
    \textbf{S}^{(i)}_{:,:,d} = \dfrac{\textbf{S}^{(i)}_{:,:,d} - \xi_d}{\chi_d - \xi_d} ,
\end{equation}

\noindent where $d \in \{1, \ldots, D\}$, and $\xi_d$, $\chi_d$ are the global minimum and maximum values for dimension $d$, over all samples frames and joints, defined as:  
\begin{equation}
    \xi_d = \min (\{\textbf{S}^{(i)}_{:,:,d}\}_{i=1}^N), \quad
    \chi_d = \max (\{\textbf{S}^{(i)}_{:,:,d}\}_{i=1}^N) .
\end{equation}

To prevent data leakage, normalization parameters ($\xi_d$, $\chi_d$) are computed exclusively on the training set.  
These statistics are then applied to normalize both the training and test sets.  
This procedure ensures that no information from the test data influences the training process.

\subsubsection{Multivariate Time Series View}

As mentioned in Section~\ref{sec:definitions}, the input skeleton-based sequence $\textbf{S}$ is represented as a three-dimensional tensor of shape $(L, J, D)$, where $L$ is the number of frames, $J$ is the number of skeleton joints, and $D$ is the dimensionality of each joint.

Among the models considered in our study, only the STGCN model (Section~\ref{sec:stgcn}) uses the input as a skeleton sequence. All other models treat the input as a Multivariate Time Series (MTS), as defined in Section~\ref{sec:definitions}.

To convert the skeleton-based input sequence into an MTS, we retain the temporal (frame) axis and flatten the joint and dimension axes. This transformation yields a two-dimensional matrix of shape $(L, J \cdot D)$.

\begin{figure}[!ht]
    \centering
    \includegraphics[width=\linewidth]{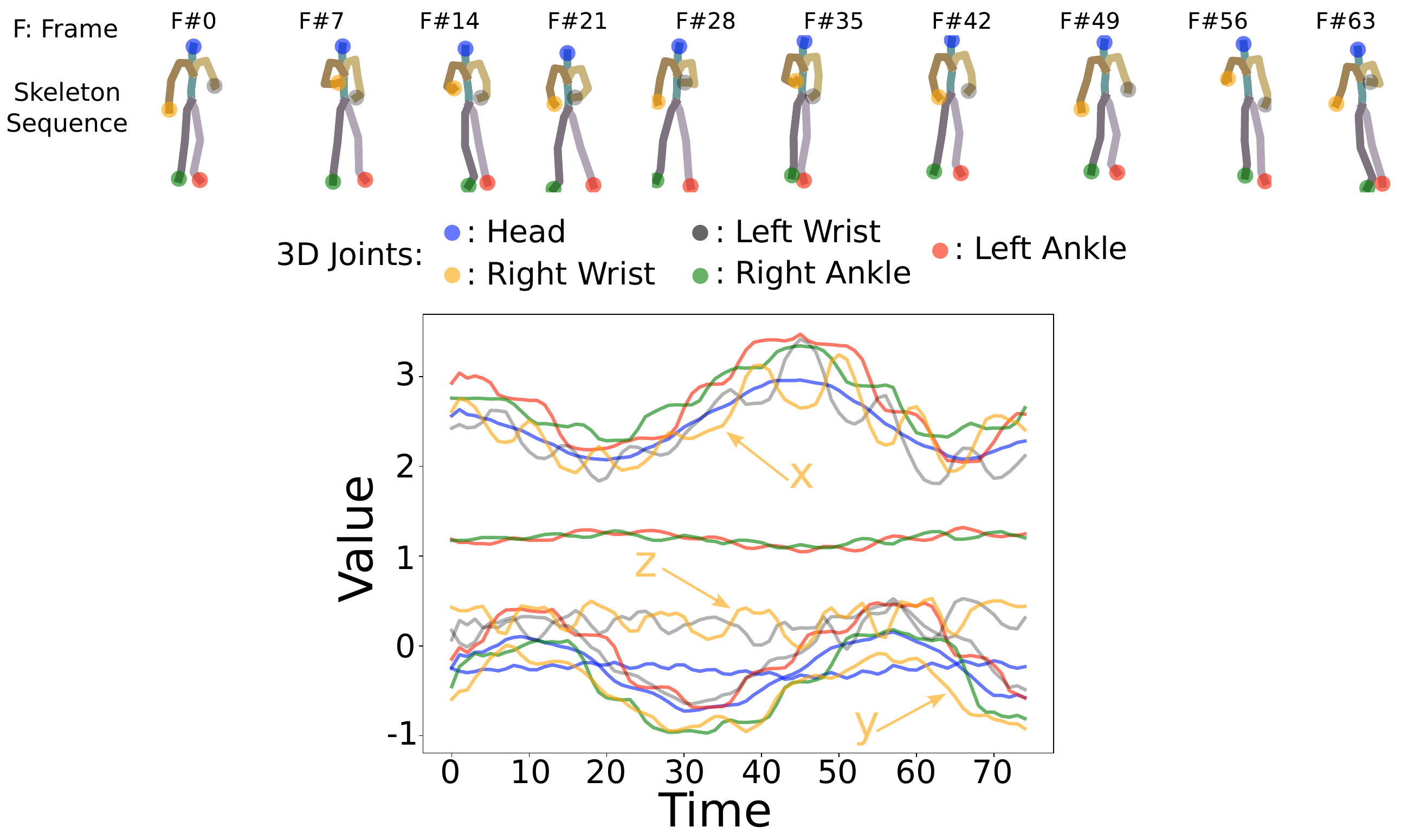}
    \caption{Transformation of a skeleton-based input sequence~\cite{gru-action-2motion} into a Multivariate Time Series (MTS) representation.}
    \label{fig:skeleton-to-mts}
\end{figure}

An illustration of this transformation is provided in Figure~\ref{fig:skeleton-to-mts}.

\subsection{Evaluation Criteria}
\subsubsection{Performance Evaluation}

In this work, we employ standard evaluation metrics for both classification and regression tasks.

\paragraph{Classification} We use both accuracy and balanced accuracy to account for potential class imbalance.  

Given a classifier that outputs a probability distribution $\hat{\textbf{y}}$ for each input sample, the predicted class label $\hat{y}$ is defined as the class with the highest predicted probability:  
\begin{equation}
    \hat{y} = \arg\max_{c \in \{1, \ldots, C\}} \hat{\textbf{y}}_c .
\end{equation}

The ground truth label is provided as a one-hot encoded vector $\textbf{y}$, where the correct class has a probability of $1.0$ and all others are $0.0$.  
Thus, the ground truth class label $y$ is also defined as the index with the highest probability.  

Accuracy is computed as the proportion of correctly predicted samples in the test set:  
\begin{equation}
    \text{accuracy} = \dfrac{1}{N_{\text{test}}} \sum_{i=1}^{N_{\text{test}}} \mathbbm{1}[y_i == \hat{y}_i] ,
\end{equation}

\noindent where $N_{\text{test}}$ is the number of samples in the test set, and $\mathbbm{1}[\cdot]$ is the indicator function defined as:  
\begin{equation}
    \mathbbm{1}[\text{condition}] = 
    \begin{cases}
        1, & \text{if condition is true} \\
        0, & \text{otherwise} .
    \end{cases}
\end{equation}

Balanced accuracy offers an alternative to standard accuracy by accounting for class imbalance in the evaluation.  
It prevents misleadingly high-performance scores that can occur when a model disproportionately predicts the majority class in an imbalanced test set.  

Balanced accuracy is defined as:
\begin{equation}
    \text{balanced accuracy} = \dfrac{1}{C} \sum_{c=1}^{C} \dfrac{
        \sum_{i=1}^{N_{\text{test}}} \mathbbm{1}[y_i = c] \cdot \mathbbm{1}[\hat{y}_i = c]
    }{
        \sum_{i=1}^{N_{\text{test}}} \mathbbm{1}[y_i = c] 
    }.
\end{equation}

\noindent
The numerator counts the number of samples from class \( c \) that were correctly predicted, while the denominator corresponds to the total number of samples whose true label is \( c \).  
This ratio represents the recall for class \( c \).  
Balanced accuracy is then obtained by averaging these recall values across all \( C \) classes.

\paragraph{Regression} We use both the Root Mean Squared Error (RMSE) and the Mean Absolute Error (MAE) as evaluation metrics for regression tasks.  
Given predicted labels $\hat{y}_i$ and ground truth labels $y_i$, these metrics are computed as follows:
\begin{equation}
    \text{RMSE} = \sqrt{\dfrac{1}{N_{test}} \sum_{i=1}^{N_{test}} (y_i - \hat{y}_i)^2} ,
\end{equation}
\begin{equation}
    \text{MAE} = \dfrac{1}{N_{test}} \sum_{i=1}^{N_{test}} |y_i - \hat{y}_i| .
\end{equation}

\noindent
MAE captures the average absolute error, treating all deviations equally, while RMSE emphasizes larger errors due to the squaring operation.  
As a result, RMSE is more sensitive to outliers and better reflects the variance in prediction errors.
Using both metrics provides a more comprehensive understanding of model performance.

It is important to note that prior to training, the regression labels of the training samples are normalized between $0.0$ and $1.0$ by dividing them by the maximum value provided in Table~\ref{tab:reg-datasets}.
During the prediction phase, given the model's output is not constrained, we bound it using the following operation:
\begin{equation}
    \hat{y} = \begin{cases}
        0.0 & \text{if } \mathcal{F}_{reg}(\textbf{S}) < 0.0\\
        1.0 & \text{if } \mathcal{F}_{reg}(\textbf{S}) > 1.0\\
        \mathcal{F}_{reg}(\textbf{S}) & \text{else} ,
    \end{cases}
\end{equation}

\noindent where $\mathcal{F}_{reg}$ is the regression model.

The regression metrics are calculated after de-normalizing the predicted labels $\hat{y}$ by multiplying them by the maximum value associated to their dataset in Table~\ref{tab:reg-datasets}.

\subsubsection{Efficiency Evaluation}

Evaluating deep learning models solely based on performance is often insufficient.  
In practice, it is equally important to consider the efficiency of a model, especially in resource-constrained environments.  

To assess model efficiency in our work, we report the following measures alongside the performance metrics described in the previous section:  
(1) training time,  
(2) inference time,  
(3) number of trainable parameters, and  
(4) FLoating-point OPerations per second (FLOPs).  

This allows us to perform a comprehensive comparison, aiming to identify models that are not only accurate but also computationally efficient.

\subsection{Implementation Details}~\label{sec:impl-det}

\subsubsection{Training Configuration and Computational Setup}

To ensure that model performance is driven solely by architectural differences, we adopt a unified training configuration across all evaluated models.  
All competing models are implemented within the same framework, using \textit{TensorFlow}-\textit{Keras}~\cite{tensorflow,keras}.  

Each model is trained for $1500$ epochs with a batch size of $64$.  
We employ a learning rate scheduling strategy based on \texttt{ReduceLROnPlateau}, with a reduction factor of $0.5$, a patience of $50$ epochs, and a minimum learning rate of $10^{-4}$.  
Model selection is based on training loss, with the best performing model checkpoint used for final evaluation on the test set.  

To ensure fairness in computational cost comparisons, all experiments are conducted on the same hardware setup.  
Specifically, we use a single machine equipped with an NVIDIA RTX 4090 GPU (24GB VRAM), an AMD Ryzen 9 7950X 16-core processor, and 64GB of RAM, running Ubuntu 24.04.  
This standardized environment is critical for assessing both performance and computational efficiency across models.
We use the \textit{aeon} open-source Python library~\cite{middlehurst2024aeon} for the implementation of the FCN and H-Inception classification and regression models.
The source code of our work is publicly available here: \url{https://github.com/MSD-IRIMAS/DeepRehabPile} and the detailed information on how to download the datasets of our archive here: \url{https://msd-irimas.github.io/pages/DeepRehabPile}

\subsubsection{Ensembling Strategy}

Ensembling is a widely used technique in machine learning and deep learning that leverages the principle of combining multiple models to improve robustness and generalization~\cite{abdullayev2025DiverseEnsemble,fawaz2019deepensembletsc}.  
For classification tasks, ensembling multiple models $\{\mathcal{M}^{(m)}_{clf}(.)\}_{m=1}^{M}$ on a single input sample $\textbf{S}$ involves averaging their predicted probability distributions to obtain the final prediction $\hat{\textbf{y}}$:
\begin{equation}
    \hat{y}_c = \dfrac{1}{M} \sum_{m=1}^M \hat{y}^{(m)}_c ,
\end{equation}

\noindent where $c \in [1, \ldots, C]$, and $C$ is the total number of classes.  
Here, $\hat{\textbf{y}} = \{\hat{y}_1, \ldots, \hat{y}_C\}$ denotes the final ensemble prediction, and $\hat{\textbf{y}}^{(m)} = \mathcal{M}^{(m)}_{clf}(\textbf{S}) = \{\hat{y}^{(m)}_1, \ldots, \hat{y}^{(m)}_C\}$ is the output of the $m$-th classification model $\mathcal{M}^{(m)}_{clf}$.  

In the case of regression, the ensemble prediction $\hat{y}$ is computed by averaging the predicted values from $M$ regression models $\{\mathcal{M}_{reg}^{(m)}\}_{m=1}^M$:
\begin{equation}
    \hat{y} = \dfrac{1}{M} \sum_{m=1}^M \hat{y}^{(m)} ,
\end{equation}

\noindent where $\hat{y}^{(m)} = \mathcal{M}_{reg}^{(m)}(\textbf{S})$ is the prediction produced by the $m$-th regression model $\mathcal{M}_{reg}^{(m)}$.

In this work, we apply an ensembling strategy to each dataset using the following procedure:
\begin{itemize}
    \item \textbf{Step 1:} Load the training and test sets from a single fold of the dataset.
    \item \textbf{Step 2:} Normalize the training data using min-max normalization, and normalize the testing data using the min and max values computed from the training data.
    \item \textbf{Step 3:} Train the same model five times on the training data, each time using a different random initialization.
    \item \textbf{Step 4:} Use the five trained models to generate predictions on the test set.
    \item \textbf{Step 5:} Combine the predictions by averaging them to produce the ensemble output.
    \item \textbf{Step 6:} Compute evaluation metrics based on the ensemble predictions. Repeat Steps 1–5 for all other folds, and report the average performance across all folds.
\end{itemize}

\begin{figure}[!ht]
    \centering
    \includegraphics[width=\linewidth]{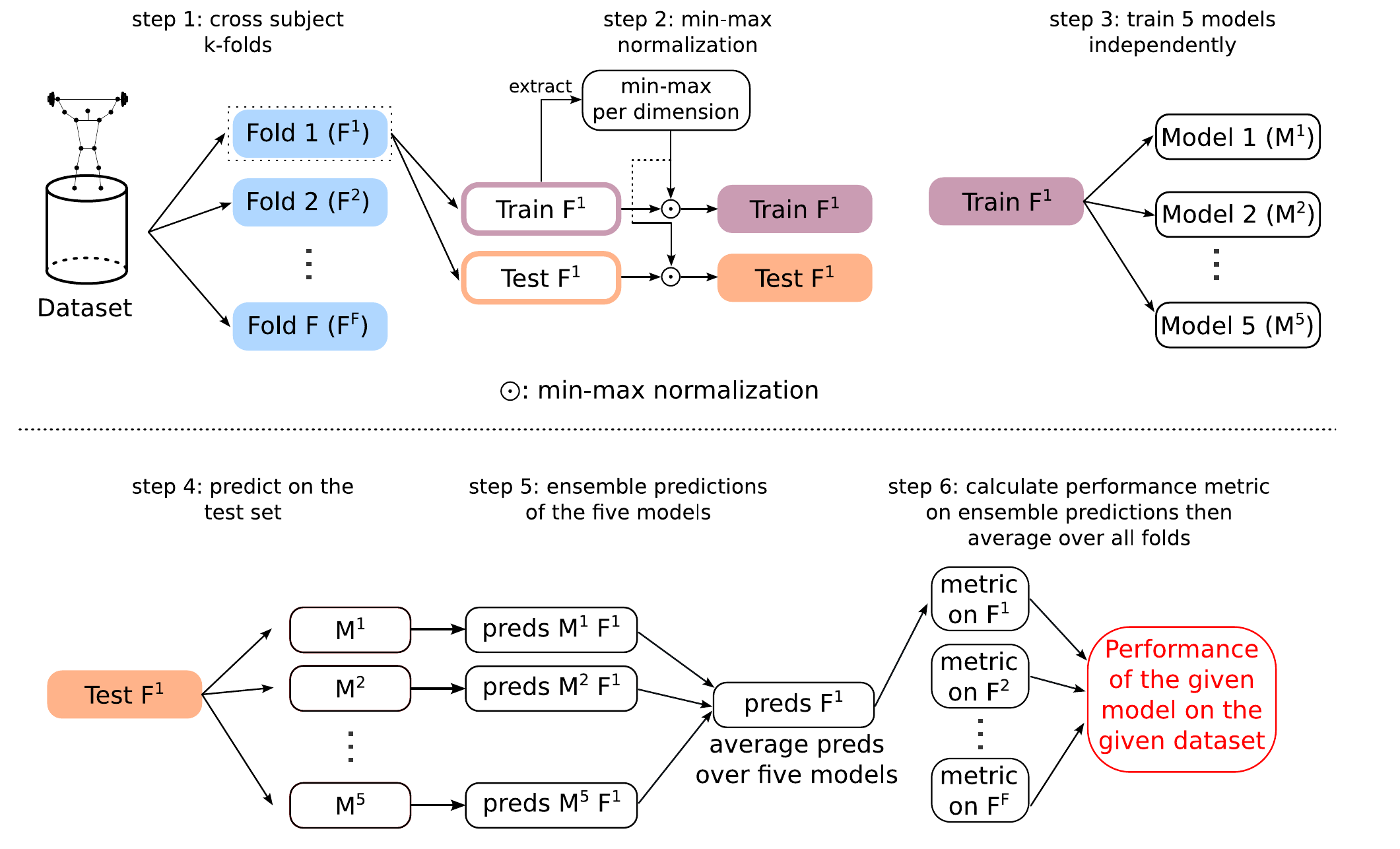}
    \caption{
    For each fold, five models with different random initializations are trained on the normalized data.
    Their predictions on the test set are averaged to produce ensemble outputs, which are then used to compute evaluation metrics.
    This process is repeated across all folds, and final performance is reported as the average of the metrics across folds.
    }
    \label{fig:exp-setup}
\end{figure}

This setup, illustrated in Figure~\ref{fig:exp-setup}, serves two main purposes.  
First, it reduces the effect of random initialization by ensembling multiple runs, which helps stabilize predictions.  
Second, it minimizes bias from a specific train-test split by averaging performance across all folds.

Note that the first averaging step (the ensemble) is done at the prediction level, without using the ground truth.  
The second averaging step is performed on the evaluation metrics, which are computed posterior to using the ground truth labels.

\section{Experimental Results}
\label{experimental-results}

\subsection{Performance}

In this section, we present a performance comparison of the nine models evaluated, considering classification and regression tasks separately.  
Given the use of multiple datasets, we employ a summarization method suitable for comparing multiple models across multiple datasets.  

We adopt the Multi-Comparison Matrix (MCM)~\cite{mcm}, which provides both a pairwise (1-vs-1) comparison between models and a ranking mechanism for evaluating overall performance.  
Models in the MCM are ranked from best to worst based on their average performance across all datasets.  

Each cell in the MCM provides three pieces of information when comparing a row model with a column model:  
(1) the difference in average performance, indicated by the cell color;  
(2) the win-tie-loss count; and  
(3) a $p$-value assessing the statistical significance of the performance difference between the two models.  

We use the Wilcoxon Signed-Rank Test~\cite{wilcoxon}, in a two-sided setup, to compute the $p$-values.  
If a $p$-value is less than $0.05$, the difference in performance is considered statistically significant.
If the $p$-value exceeds $0.05$, we cannot draw a conclusion about statistical significance.

\subsubsection{Regression}

\begin{figure}[!ht]
    \centering
    \includegraphics[width=\linewidth]{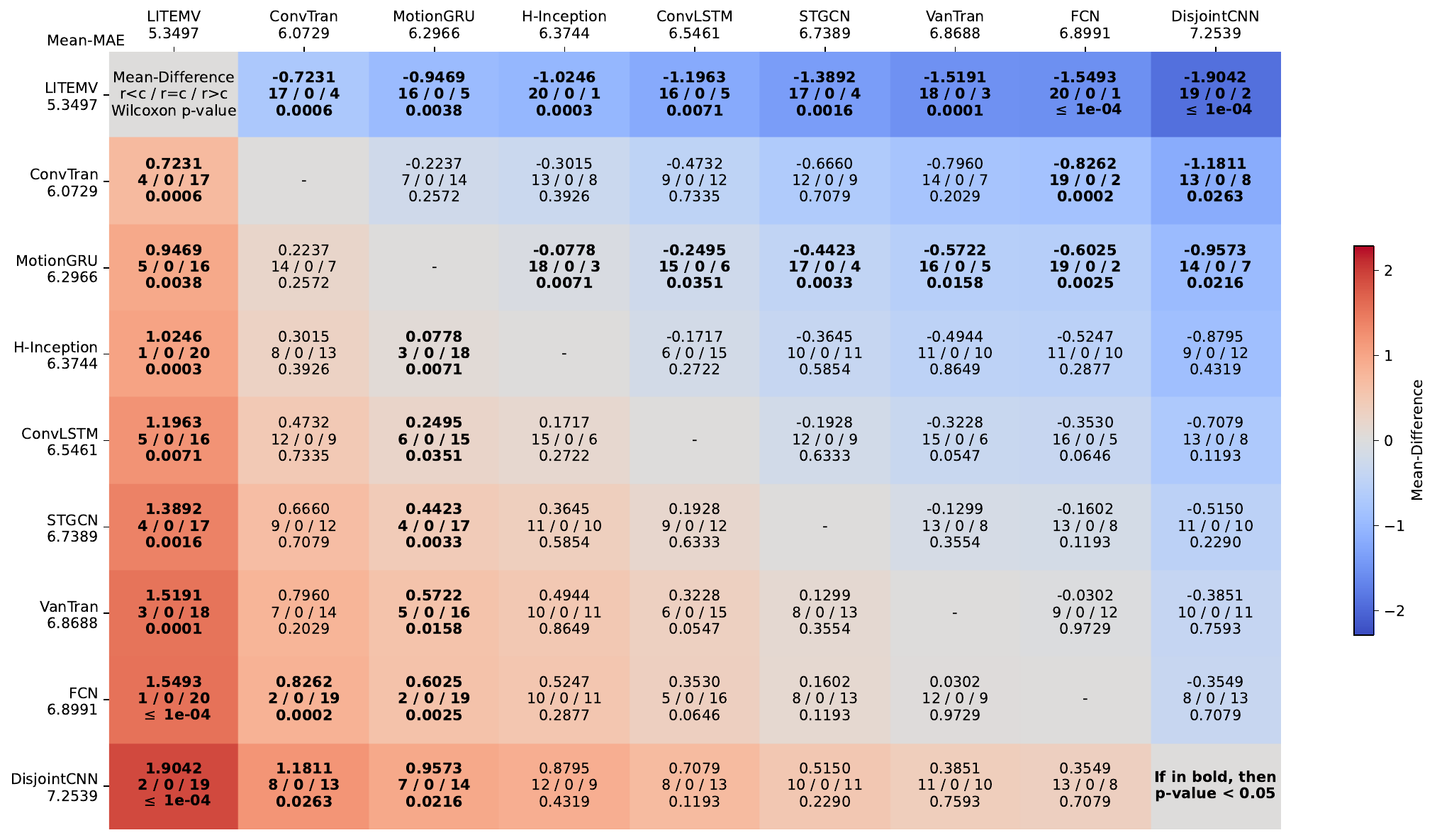}
    \caption{A Multi-Comparison Matrix (MCM) comparing nine models, in their ensemble version, over $21$ extrinsic regression datasets on the Mean Absolute Error (MAE) metric.}
    \label{fig:mcm-reg-mae}
\end{figure}

\begin{figure}[!ht]
    \centering
    \includegraphics[width=\linewidth]{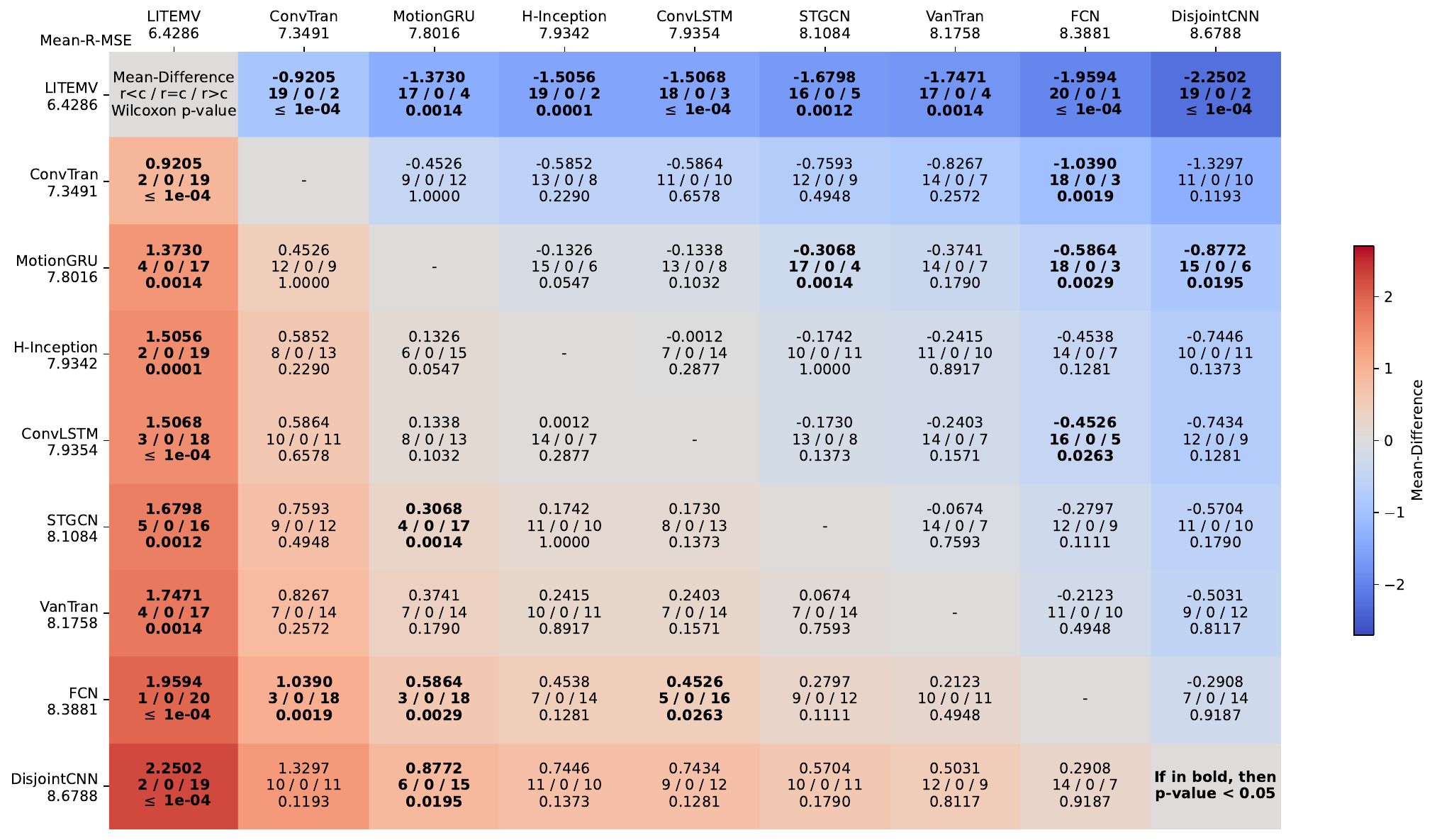}
    \caption{A Multi-Comparison Matrix (MCM) comparing nine models, in their ensemble version, over $21$ extrinsic regression datasets on the Root Mean Squared Error (RMSE) metric.}
    \label{fig:mcm-reg-rmse}
\end{figure}

In Figures~\ref{fig:mcm-reg-mae} and~\ref{fig:mcm-reg-rmse}, we present the MCMs for the MAE and RMSE metrics, respectively, for the extrinsic regression task.  
A noticeable observation from these two MCMs is that the best-performing model, LITEMV, across all datasets statistically outperforms all competitors, with small $p$-values, with an average performance of $5.3497$ and $6.4286$ on the MAE and RMSE measures, respectively.  

Furthermore, it is important to highlight the variability in model rankings.  
For instance, although LITEMV performs very well and is CNN-based, other CNN-based models do not perform as strongly, as ConvTran and MotionGRU outperform them.  

ConvTran, the second-best model, is a self-attention-based model with a convolutional embedding.  
This suggests the presence of long-term dependencies in the datasets, with high correlation across all temporal dimensions.  
This is consistent with the nature of the task, where information from the beginning, middle, and end of an exercise can all contribute to assessing how well the exercise was performed by a subject.
However, VanTran, which is also a self-attention-based model, does not perform as well as ConvTran.  
This decline in performance may be attributed to differences in their embedding strategies: VanTran uses a non-linear matrix transformation over all frames, while ConvTran captures temporal patterns through convolutional embeddings.  
This suggests that, for this regression task, not only are long-term dependencies important, but the preservation of informative features is also crucial for achieving strong downstream performance.

Meanwhile, H-Inception does not perform as well as LITEMV, even though both models utilize similar components such as hand-crafted convolutional filters and multiplexed convolutions.  
However, LITEMV employs Depthwise Separable Convolutions (DWSCs), which capture not only temporal correlations but also spatial ones, an essential aspect that H-Inception lacks.

Moreover, while FCN is also a CNN-based model, its small receptive field and vanilla architecture limit its performance.  
Although DisjointCNN has its own method for capturing spatial correlations similar to LITEMV, it lacks both hand-crafted filters and multiplexed convolutions.  
As a result, DisjointCNN's performance is significantly lower compared to both LITEMV and H-Inception.

In terms of recurrent models, MotionGRU outperforms ConvLSTM despite not using convolutional embeddings.  
Unlike Self-Attention mechanisms, recurrent networks can capture long-term temporal dependencies through their internal memory.  
They are also capable of retaining feature-level information, which is important for regression tasks.

The only GNN-based model in our comparison, STGCN, ranks sixth in terms of both average MAE and RMSE performance.  
Although this model is widely used in the human rehabilitation assessment community, it has not previously been evaluated under the same experimental setup as used in this study.  
To our knowledge, it has also not been directly compared to the other models included in our benchmark.  
Notably, in the literature, STGCN is almost always evaluated on either the KIMORE or UIPRMD extrinsic regression datasets.  
While our experimental protocol differs, particularly in enforcing cross-subject splits, STGCN achieves 4 wins over LITEMV on the MAE metric and 5 wins on the RMSE metric, and all of these wins come from those two datasets.  
This suggests that STGCN performs well under a fair evaluation setup on KIMORE and UIPRMD, likely due to the necessity of capturing structural correlations between joints in order to perform the downstream task.

\subsubsection{Classification}

\begin{figure}[!ht]
    \centering
    \includegraphics[width=\linewidth]{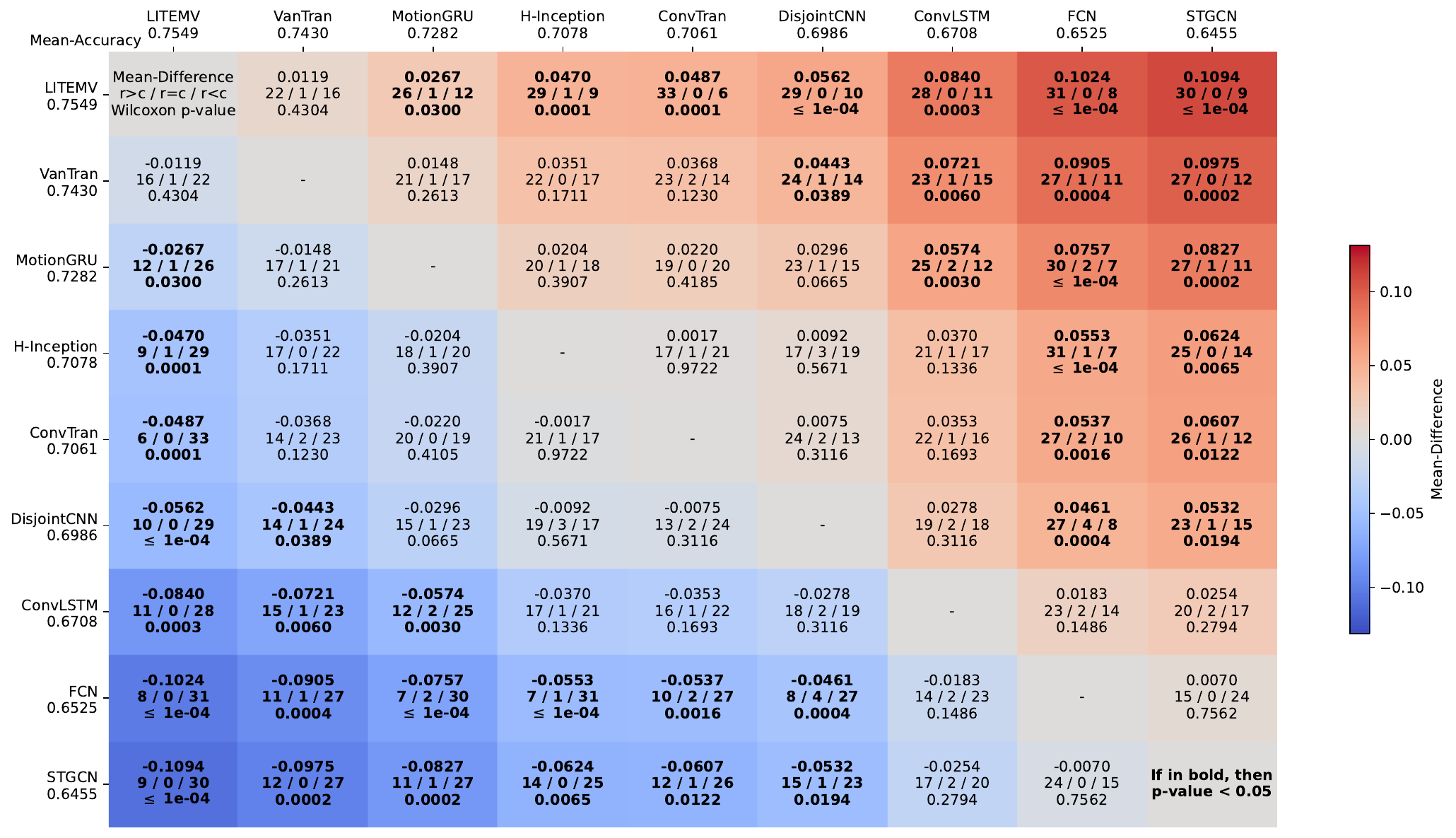}
    \caption{A Multi-Comparison Matrix (MCM) comparing nine models, in their ensemble version, over $39$ classification datasets on the Accuracy metric.}
    \label{fig:mcm-cls-acc}
\end{figure}

\begin{figure}[!ht]
    \centering
    \includegraphics[width=\linewidth]{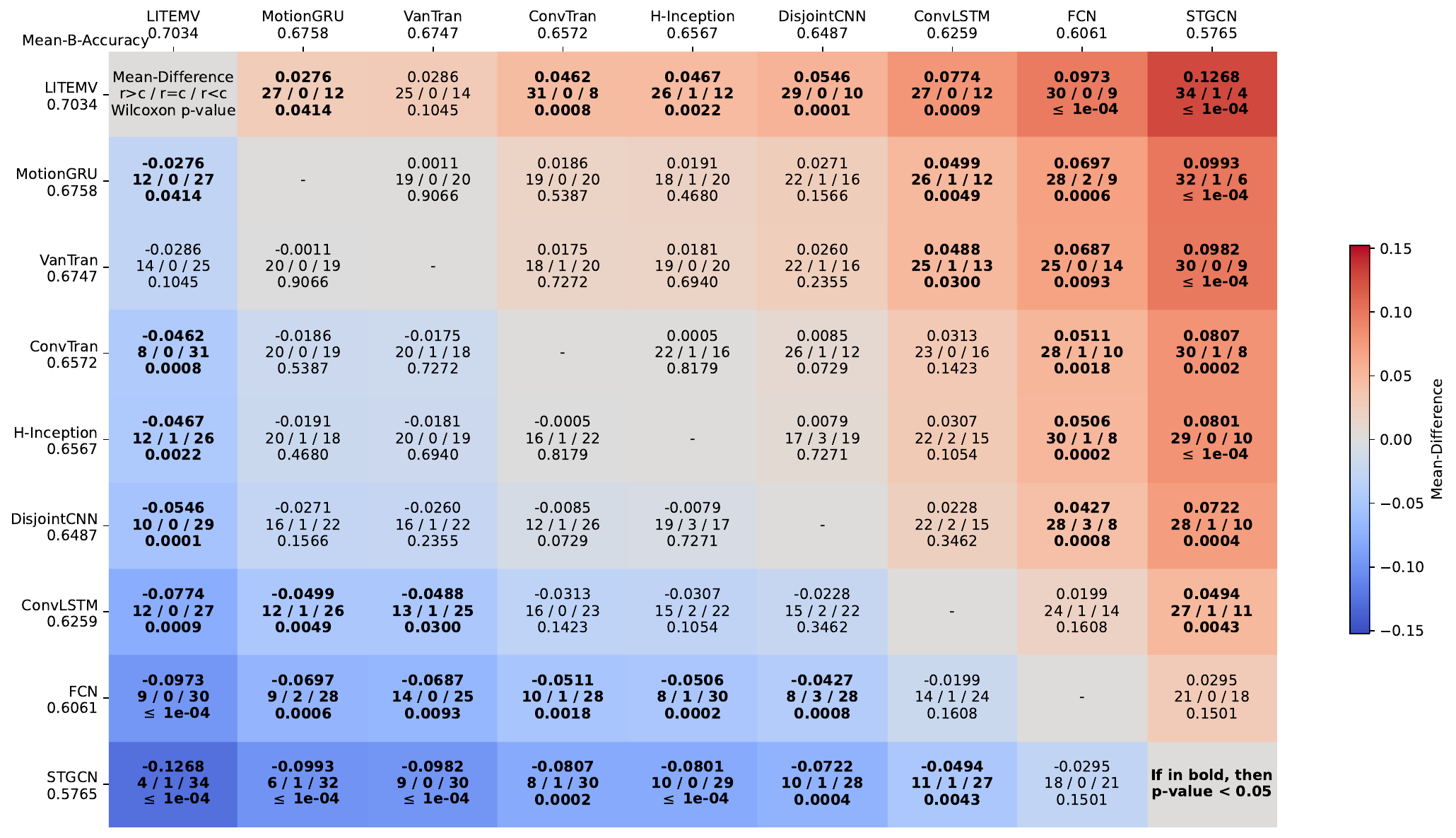}
    \caption{A Multi-Comparison Matrix (MCM) comparing nine models, in their ensemble version, over $39$ classification datasets on the Balanced Accuracy metric.}
    \label{fig:mcm-cls-bal-acc}
\end{figure}

In Figures~\ref{fig:mcm-cls-acc} and~\ref{fig:mcm-cls-bal-acc}, we present the MCMs for classification results using accuracy and balanced accuracy metrics, respectively.
Unlike the regression performance presented in the previous section, classification results do not reveal a clear winner.
While LITEMV still demonstrates the best overall performance on average, VanTran closely follows, with only a marginal decrease in average performance, just $1.19\%$ in accuracy and $2.86\%$ in balanced accuracy.
Furthermore, the $p$-value between LITEMV and VanTran is higher than $0.05$, indicating that, across the $39$ classification datasets, no statistically significant difference in performance can be concluded between these two models.
Moreover, in terms of balanced accuracy, MotionGRU surpasses VanTran and ranks second overall, owing to its slight performance improvement on unbalanced datasets.

Interestingly, a clear swap in ranking occurs between VanTran and ConvTran when moving from regression to classification tasks.
Although both models employ the Self-Attention mechanism, this swap is likely due to their differing input encoding strategies: ConvTran applies temporal convolution before Self-Attention, while VanTran uses a sliding linear embedding across the temporal axis.
In this application, the convolutional embedding used by ConvTran appears to be more effective for regression tasks, potentially due to its ability to process local patterns before applying Self-Attention.
In contrast, for the classification tasks considered here, where VanTran outperforms ConvTran, this added local encoding step seems to offer less advantage.

The three models, ConvTran, H-Inception, and DisjointCNN, exhibit relatively small differences in performance among each other.
However, it is worth noting that MotionGRU is better positioned relative to LITEMV in terms of balanced accuracy, with a $p$-value of $0.04$.
This suggests that MotionGRU handles unbalanced datasets slightly better in some cases compared to ConvTran, H-Inception and DisjointCNN. Nevertheless, its performance is still insufficient to outperform LITEMV on average.

Overall, it is important to note that the best-performing model, LITEMV, achieves an average accuracy of only $75.49\%$, indicating that this classification task remains challenging and far from straightforward. This highlights the need for continued exploration and the development of more effective model architectures.
Additionally, there is a noticeable gap between accuracy and balanced accuracy across all models. For example, in the case of LITEMV, this gap is nearly $5\%$, and similar discrepancies are observed for the other models as well.
This suggests that all models are experiencing some difficulty in handling the class imbalance present in the classification datasets.

In contrast, FCN and ConvLSTM perform poorly across both metrics. These results indicate that simple CNN-based architectures are inadequate for this task, and that GRU-based models tend to perform better than LSTM-based ones in this application.

Finally, STGCN performs the worst among all models in the classification setting, unlike its relatively stronger performance in regression, achieving results only slightly better than chance.
To the best of our knowledge, STGCN has not previously been evaluated on classification tasks and has primarily been applied to regression. These results suggest that, within the context of rehabilitation assessment, GCN-based architectures may not be well-suited for classification tasks.

\subsection{Efficiency}

\begin{figure}[!ht]
    \centering
    \includegraphics[width=0.7\linewidth]{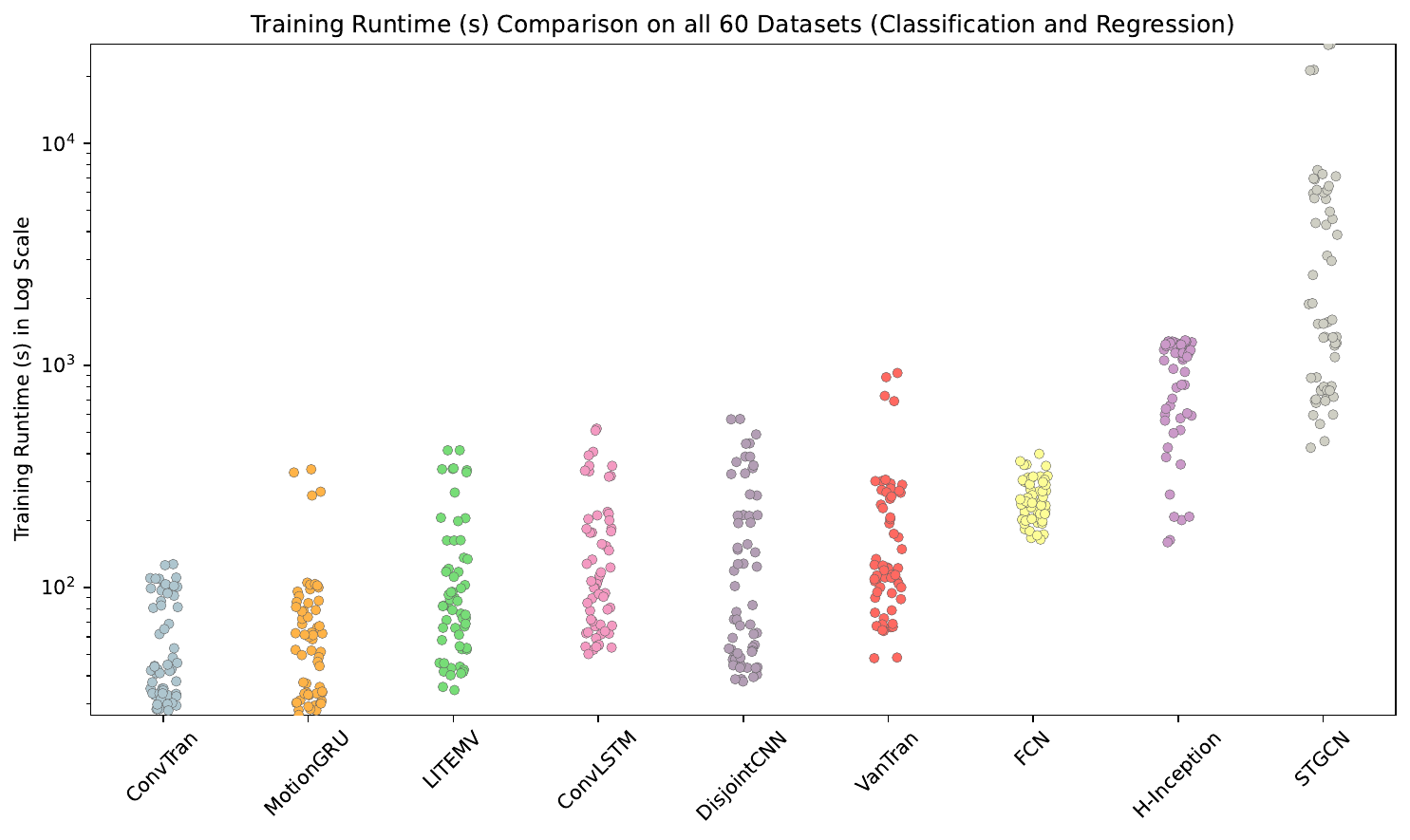}
    \caption{Strip plot comparing training runtime (in seconds) of nine models across 60 datasets, ordered by average value from lowest to highest.}
    \label{fig:training-time}
\end{figure}

\begin{figure}[!ht]
    \centering
    \includegraphics[width=0.7\linewidth]{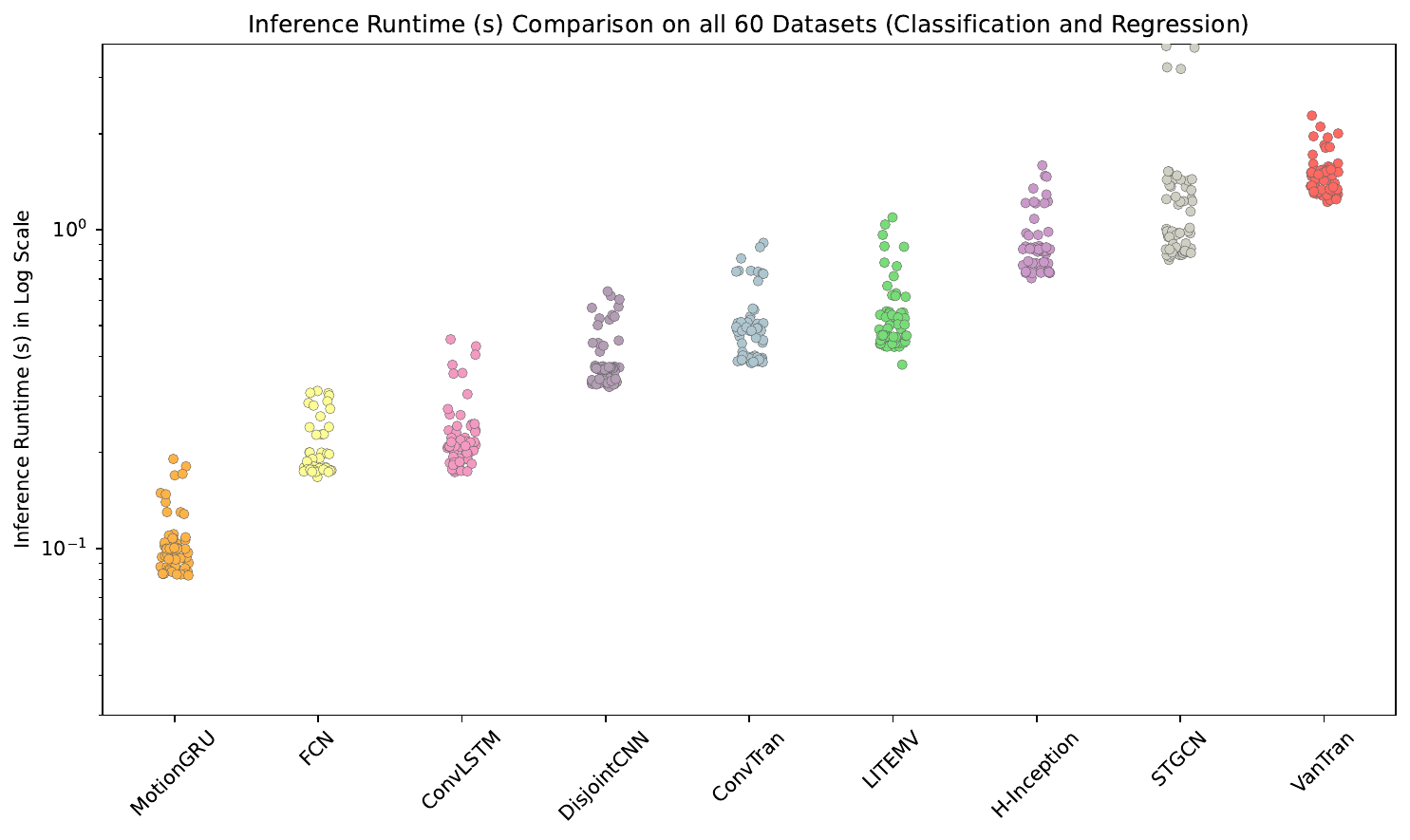}
    \caption{Strip plot comparing inference runtime (in seconds) of nine models across 60 datasets, ordered by average value from lowest to highest.}
    \label{fig:inference-time}
\end{figure}

\begin{figure}[!ht]
    \centering
    \includegraphics[width=0.7\linewidth]{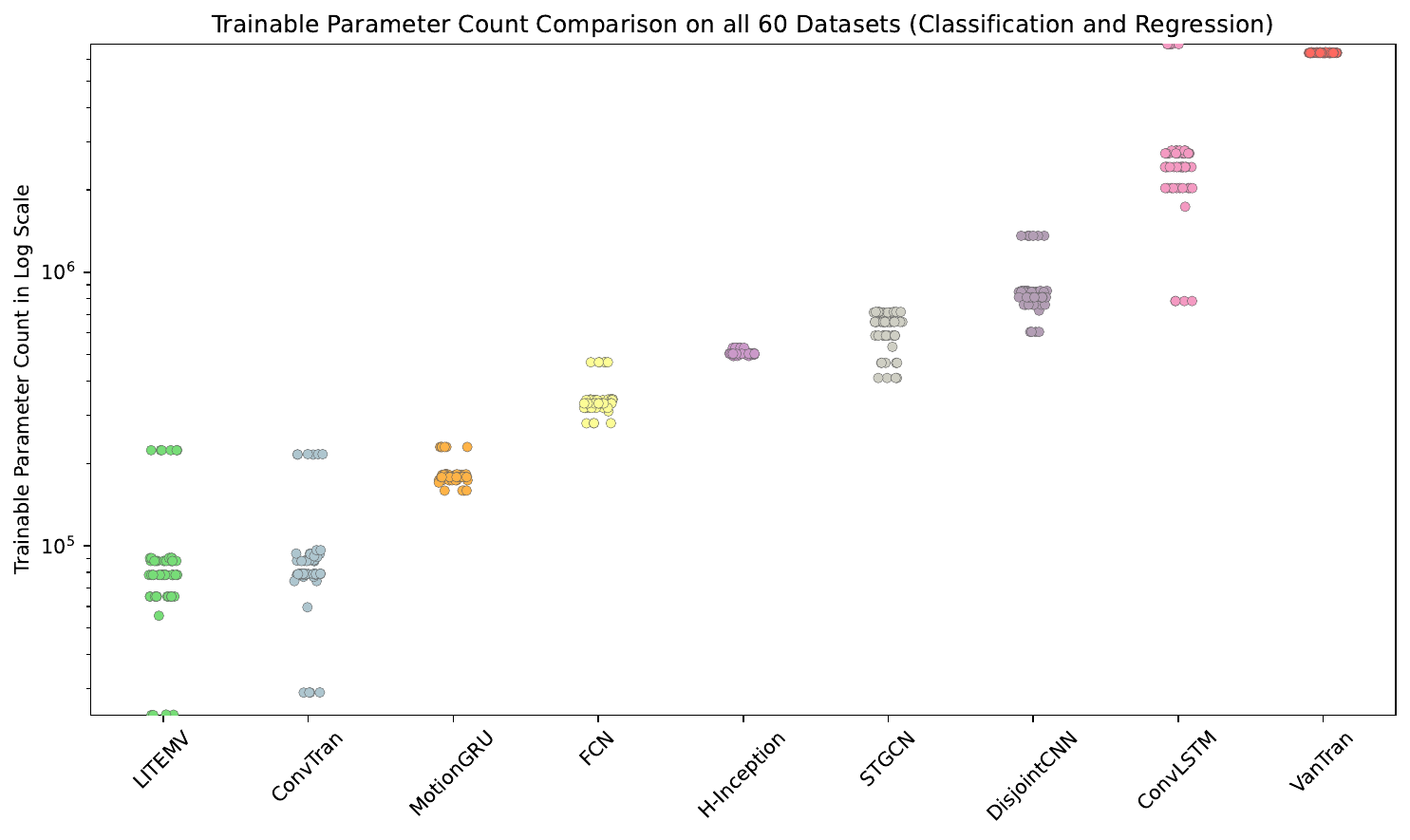}
    \caption{Strip plot comparing number of trainable parameters of nine models across 60 datasets, ordered by average value from lowest to highest.}
    \label{fig:params}
\end{figure}

\begin{figure}[!ht]
    \centering
    \includegraphics[width=0.7\linewidth]{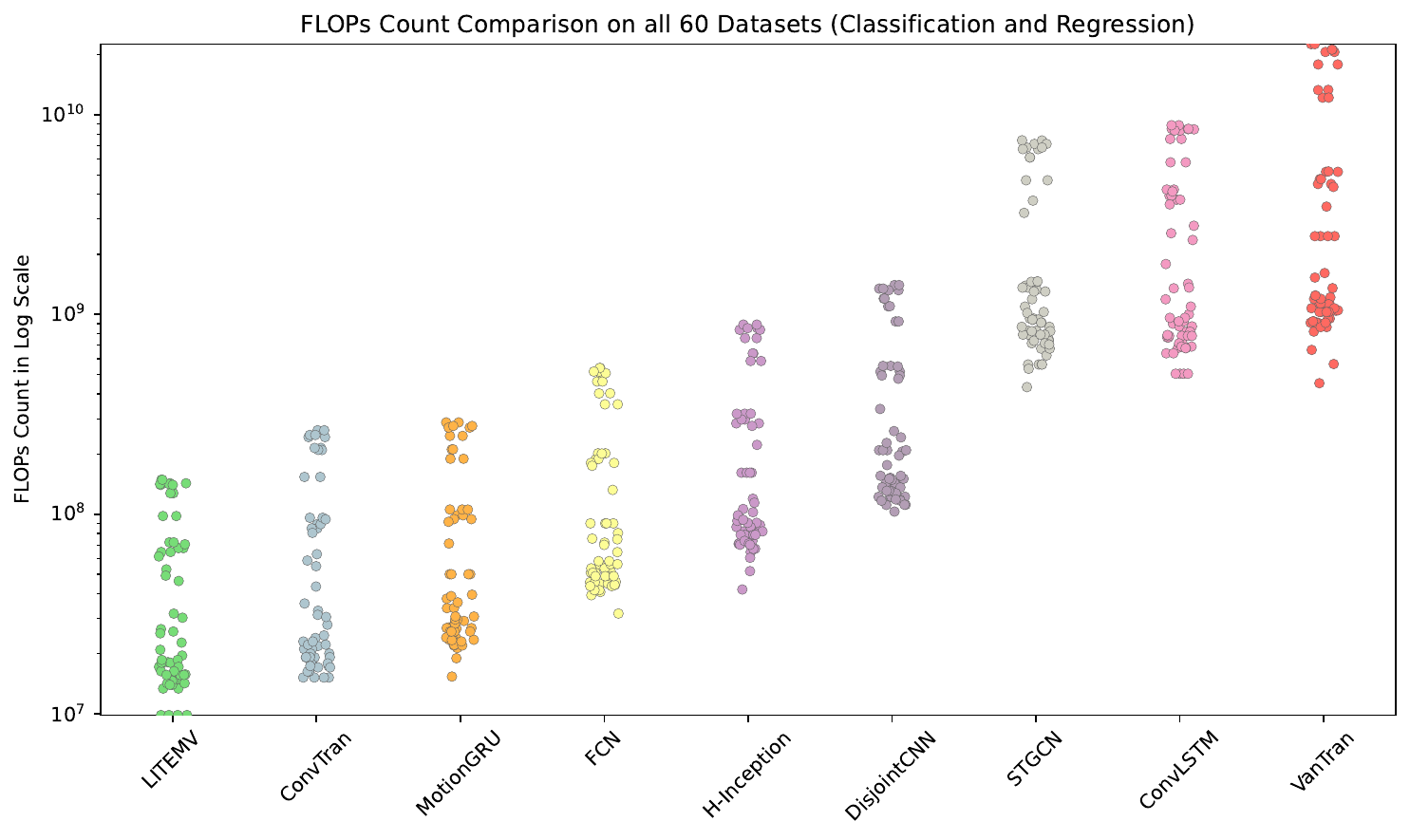}
    \caption{Strip plot comparing number of FLOPs of nine models across 60 datasets, ordered by average value from lowest to highest.}
    \label{fig:flops}
\end{figure}

In this section, we compare the nine models in our study in terms of 4 different aspects: training runtime, inference runtime, number of FLoating-point OPerations (FLOPs) per forward pass and the number of trainable parameters.
Given that these pieces of information are independent of the downstream task being either classification or extrinsic regression, we aggregate the comparison over all 60 datasets regardless of their task.
We present the results of these four listed aspects in the form of a strip plot in Figures~\ref{fig:training-time},~\ref{fig:inference-time}, ~\ref{fig:params} and~\ref{fig:flops}.

First, Figure~\ref{fig:training-time} shows a high standard deviation in training time for nearly all models, with the exception of FCN.
This low deviation for FCN can be attributed to the model's slight independence from dataset-specific information and its use of very small kernel sizes, which speeds up gradient computation.
ConvTran is the fastest model to train, despite its high dependency on the series length.
This minimal impact from series length is likely due to ConvTran having only a single Self-Attention layer.
In contrast, VanTran includes multiple Self-Attention layers, placing it sixth in training speed and contributing to its high variance across datasets.

Second, Figure~\ref{fig:inference-time} highlights a key insight: the standard deviation of inference time is nearly consistent across all nine models, with STGCN having only one notable outlier.
MotionGRU is clearly the fastest model for inference, which is controversial intuitively given its RNN-based architecture.
However, this intuition is somewhat diminished in our case, as the average series length in the dataset is $267$, relatively short, limiting the impact of recurrent computation efficiency.
Moreover, there is a clear discrepancy between the ranking of models in terms of training time and inference time.
This can be attributed to the fact that inference does not involve gradient computation, which reveals that some models have a much higher computational cost during the backward pass compared to their feed-forward pass.

Third, Figure~\ref{fig:params} illustrates which models are more dependent, in terms of trainable parameters count, on dataset characteristics such as sequence length, number of joints, and data dimensionality. For example, H-Inception and VanTran appear to be the least dependent given the small gap between the smallest and highest value.
This is due to the fact that H-Inception applies a bottleneck at the input to reduce this dependency, and VanTran’s embedding layer is a simple matrix transformation that scales linearly with the number of joints and dimensions.
LITEMV and ConvTran exhibit similar parameter counts, as both utilize convolutional embeddings that separate temporal and spatial feature extraction. While traditional Self-Attention-based models typically have a small number of parameters, ConvTran compensates for this by incorporating a learnable relative positional encoder, increasing the count of trainable parameters.

Finally, Figure~\ref{fig:flops} shows that most models exhibit similar difference between the smallest and highest value in terms of the number of FLOPs.
In the case of VanTran, the larger deviation and its ranking as the most computationally expensive model are due to its FLOPs scaling quadratically with the series length of the dataset.
LITEMV demonstrates the lowest FLOPs, significantly smaller than the CNN-based models.
Additionally, ConvTran ranks second in computational efficiency, with the lowest FLOPs after LITEMV, and is closely followed by MotionGRU.
The difference in ranking between VanTran and ConvTran stems from the fact that ConvTran uses only a single self-attention layer, whereas VanTran employs four.
Although MotionGRU is a recurrent model, the number of operations in a single GRU layer remains consistent across the sequence and primarily scales with dataset characteristics.
As a result, its computational cost is closer to that of CNN-based models in terms of FLOPs.
 
\subsection{Finding The One Model To Rule Them All}

Overall, in order to find the trade off between performance and efficiency between the nine models, regardless of the downstream task, we utilize the average rank over all $60$ datasets of classification and extrinsic regression, as illustrated in Figure~\ref{fig:trade-off-perf-eff}.
The figure also includes two additional efficiency related information, the number of FLOPs ($x$-axis) and the number of trainable parameters (size of the circles).

Overall, LITEMV is the best-performing model as well as the most efficient, both in terms of the number of parameters and FLOPs.
MotionGRU is ranked on average closest to LITEMV, with slight less efficiency.
Both VanTran and ConvTran exhibit very similar average ranks, however, VanTran is considerably less efficient.

\begin{figure}
    \centering
    \begin{subfigure}{0.45\textwidth}
    \includegraphics[width=\textwidth]{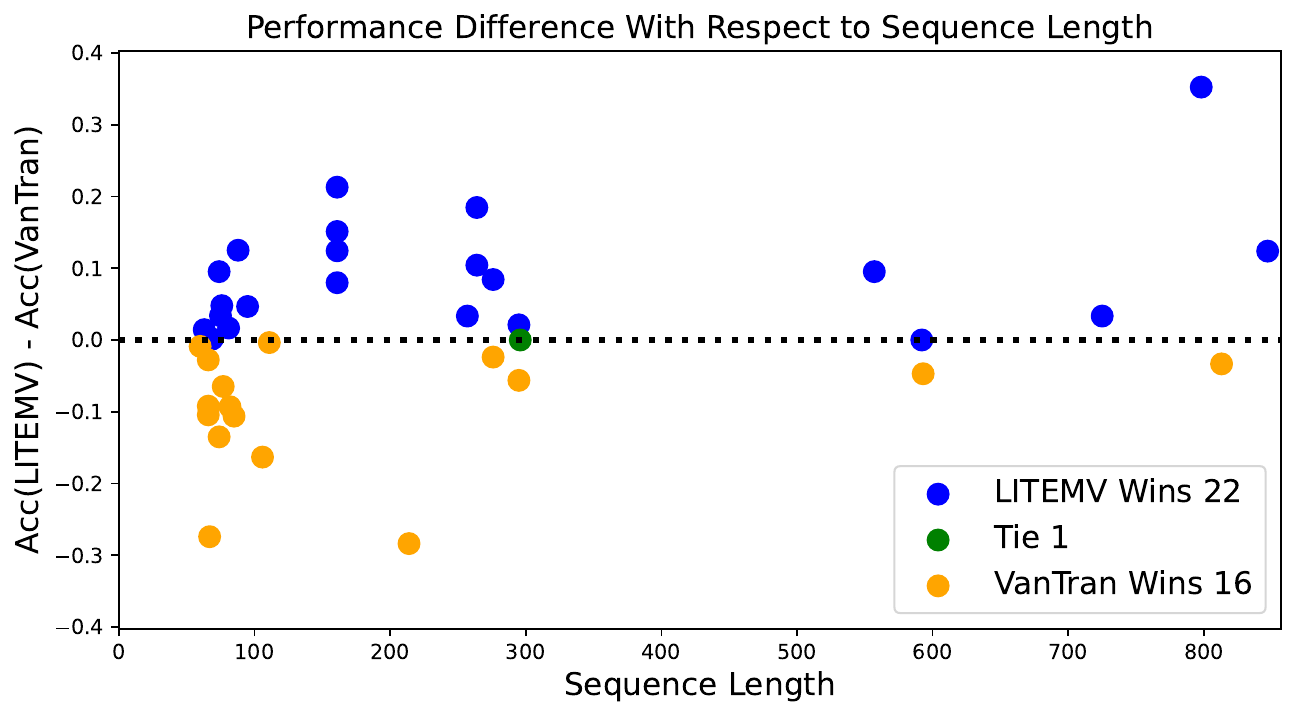}
    \caption{Comparing with respect to the sequence length (number of frames) of the datasets.}
\end{subfigure}
\hfill
\begin{subfigure}{0.45\textwidth}
    \includegraphics[width=\textwidth]{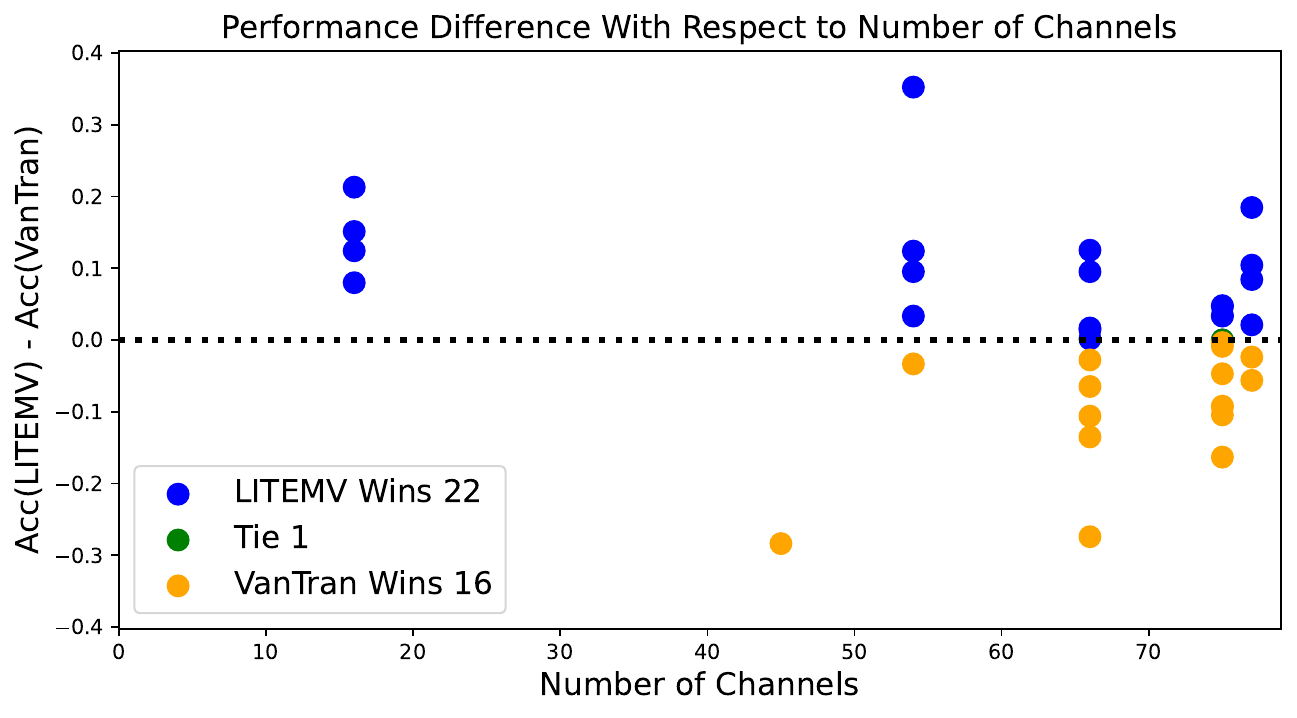}
    \caption{Comparing with respect to the number of channels of the dataset.}
\end{subfigure}
    \caption{Comparing in terms of accuracy metric both LITEMV and VanTran in their ensemble form over the $39$ classification datasets of our archive with respect to the sequence length and number of channels.}
    \label{fig:cls-vantran-vs-litemv}
\end{figure}

By combining the analysis from Figure~\ref{fig:trade-off-perf-eff} with the previous MCMs for classification and extrinsic regression (Figures~\ref{fig:mcm-cls-acc},\ref{fig:mcm-cls-bal-acc},\ref{fig:mcm-reg-rmse}, and~\ref{fig:mcm-reg-mae}), we can confidently state that LITEMV offers the best trade-off between performance and efficiency.
However, when considering performance alone, it is worth noting that LITEMV and VanTran show no statistically significant difference.
To further investigate the performance differences between these two models, Figure~\ref{fig:cls-vantran-vs-litemv} presents a one-vs-one scatter plot comparing VanTran and LITEMV across 39 classification datasets in terms of accuracy with respect to sequence length and number of channels.
From Figure~\ref{fig:cls-vantran-vs-litemv}, we observe that no clear pattern emerges when comparing the models with respect to sequence length. However, when analyzing performance in relation to the number of channels (number of joints multiplied by the number of dimensions per joint), LITEMV consistently outperforms VanTran on datasets with a small number of channels.
This observation aligns with the original findings in the LITEMV article~\cite{litemv}, which showed similar trends when comparing LITEMV to ConvTran on the UEA Time Series Classification Archive~\cite{uea-archive}.

\section{Conclusions}
\label{conclusion}

In this work, we address the task of skeleton-based human motion rehabilitation assessment, an important challenge in the medical and healthcare domain.
We focus specifically on the application of deep learning models to video-based skeleton sequence data, a promising approach that has seen significant progress in recent years.

One of the major gaps in this field is the lack of standardized benchmarking.
Existing works often use different datasets, inconsistent evaluation protocols, and rarely provide publicly available source code.
This lack of reproducibility and comparability hinders progress and makes it difficult for new researchers to build upon prior work.

To address these limitations, we propose a unified framework that evaluates nine deep learning models specifically designed for sequential skeleton data.
We benchmark these models across $60$ rehabilitation assessment datasets, $39$ for classification and $21$ for extrinsic regression.

Our experimental results show that the LITEMV model, particularly in its ensemble form, consistently outperforms others in both accuracy and efficiency.

We believe this work provides a valuable foundation for future research by offering a comprehensive benchmark, complemented with publicly available source code and datasets, to support reproducibility and drive further innovation in automated rehabilitation assessment.

\section*{Acknowledgments}
This work was supported by the ANR DELEGATION project (grant ANR-21-CE23-0014) of the French Agence Nationale de la Recherche. The authors would like to acknowledge the High Performance Computing Center of the University of Strasbourg for supporting this work by providing scientific support and access to computing resources. Part of the computing resources were funded by the Equipex Equip@Meso project (Programme Investissements d'Avenir) and the CPER Alsacalcul/Big Data. The authors would also like to thank the creators and providers of the original datasets in our archive.

\section*{Declarations}

\begin{itemize}
\item \textbf{Funding:} This work was supported by the ANR DELEGATION project (grant ANR-21-CE23-0014) of the French Agence Nationale de la Recherche (ANR)
\item \textbf{Conflict of interest/Competing interests:} The authors certify that they have no conflict of interest in the subject matter or materials discussed in this manuscript. 
\item \textbf{Ethics approval:} Not Applicable
\item \textbf{Consent to participate:} Not Applicable
\item \textbf{Consent for publication:} Not Applicable
\item \textbf{Availability of data and materials:} All of the datasets used in this work are publicly available.
\item \textbf{Code availability:} The source code is available on this github repository: \url{https://github.com/MSD-IRIMAS/DeepRehabPile}
\item \textbf{Authors' contributions:} Conceptualization: AIF, MD; Experiments: AIF, MD; Methodology: AIF, MD; Validation: AIF, MD, SB, JW, GF; Writing—original draft: AIF, MD; Writing—review and editing: AIF, MD, SB, JW, GF; Funding acquisition: MD, SB, JW, GF; all authors have read and agreed to the published version of the manuscript.
\end{itemize}

\bibliography{rehab_pile}

\newpage
\appendix

\section{Regression and Classification Datasets Label Distribution}\label{apx:label-distribution}

\begin{figure}[h]
  \centering
  \begin{subfigure}{0.32\textwidth}
    \includegraphics[width=\textwidth]{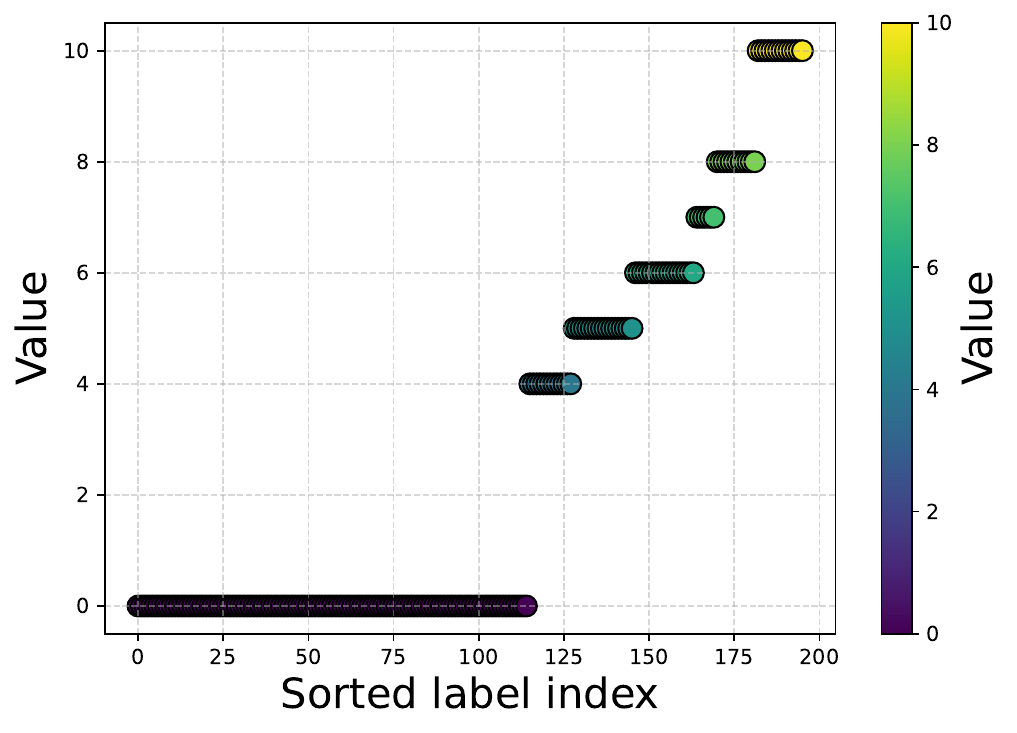}
    \caption{Bend waist to left (BWL) exercise.}
    \label{fig:sub1}
  \end{subfigure}
  \hfill
  \begin{subfigure}{0.32\textwidth}
    \includegraphics[width=\textwidth]{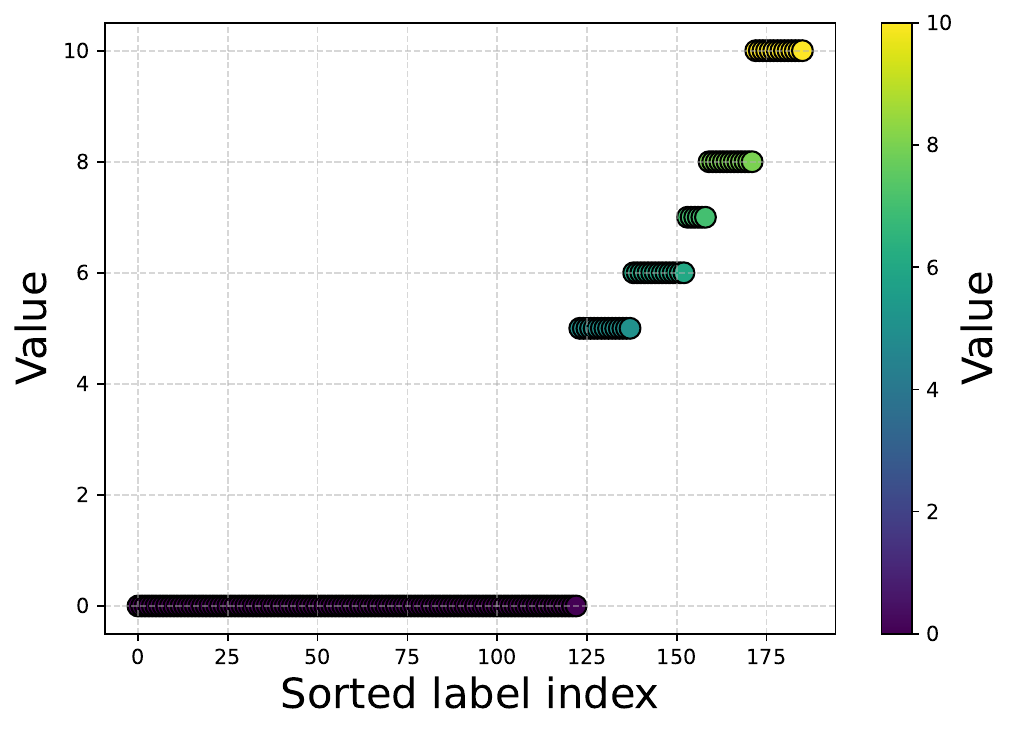}
    \caption{Bend waist to right (BWR) exercise.}
    \label{fig:sub2}
  \end{subfigure}
  \hfill
  \begin{subfigure}{0.32\textwidth}
    \includegraphics[width=\textwidth]{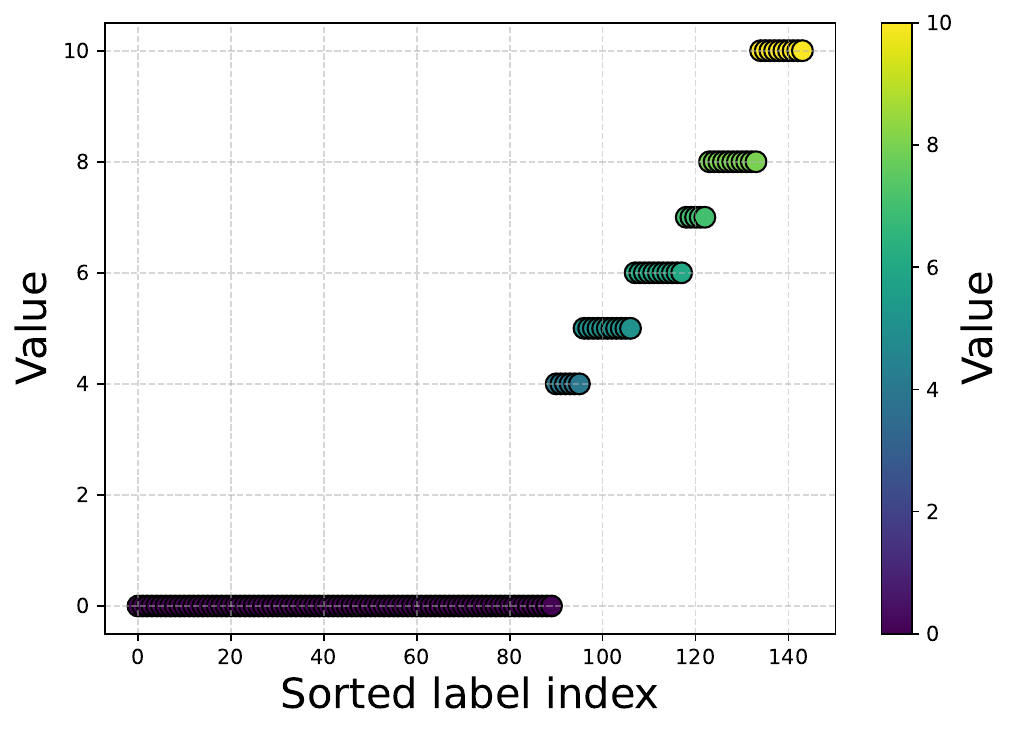}
    \caption{Hands up and down (HUD) exercise.}
    \label{fig:sub2}
  \end{subfigure}\\
  \begin{subfigure}{0.32\textwidth}
    \includegraphics[width=\textwidth]{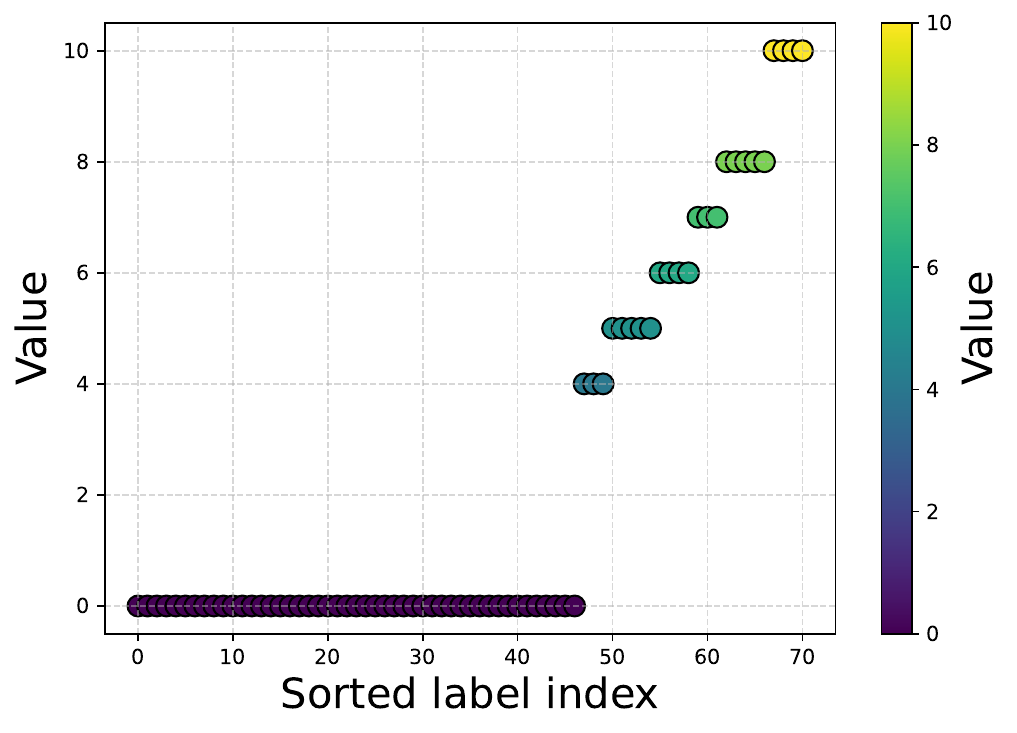}
    \caption{Walk backward (WB) exercise.}
    \label{fig:sub1}
  \end{subfigure}
  \hfill
  \begin{subfigure}{0.32\textwidth}
    \includegraphics[width=\textwidth]{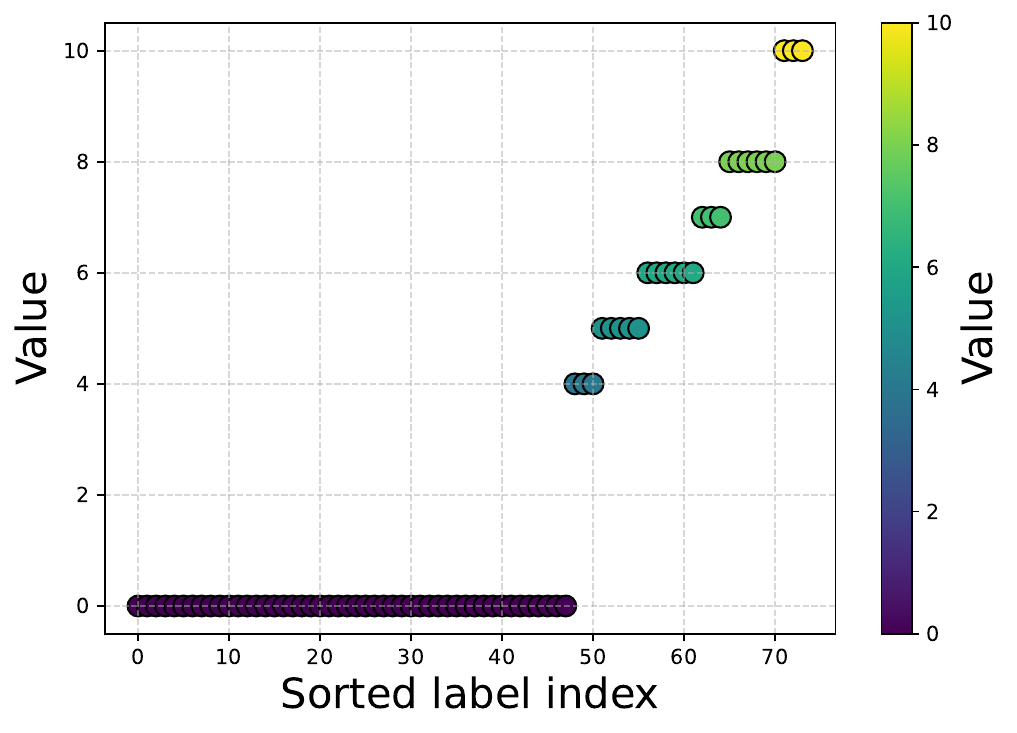}
    \caption{Walk forward (WF) exercise.}
    \label{fig:sub1}
  \end{subfigure}
  \hfill
  \begin{subfigure}{0.32\textwidth}
    \includegraphics[width=\textwidth]{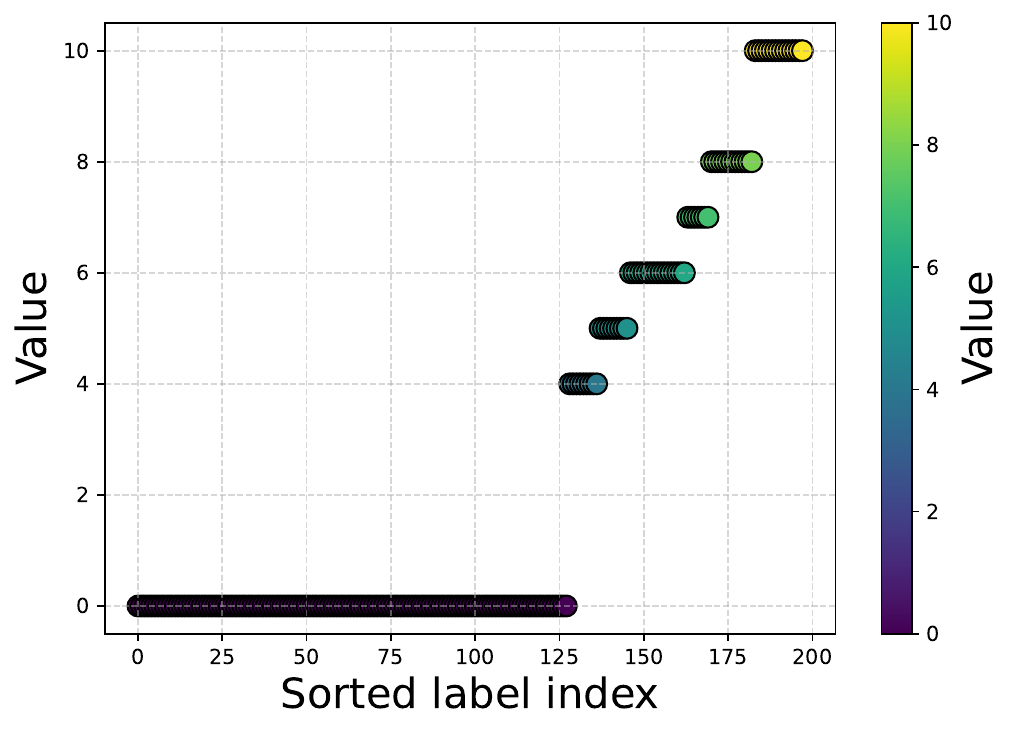}
    \caption{Wave hands (WH) exercise.}
    \label{fig:sub1}
  \end{subfigure}
  
  \caption{Regression labels distribution for the $6$ exercises of the Elderly Home Exercice (EHE) dataset~\cite{bruce2021skeleton}.}
  \label{fig:ehe-labels}
\end{figure}

\begin{figure}[h]
  \centering
  \begin{subfigure}{0.32\textwidth}
    \includegraphics[width=\textwidth]{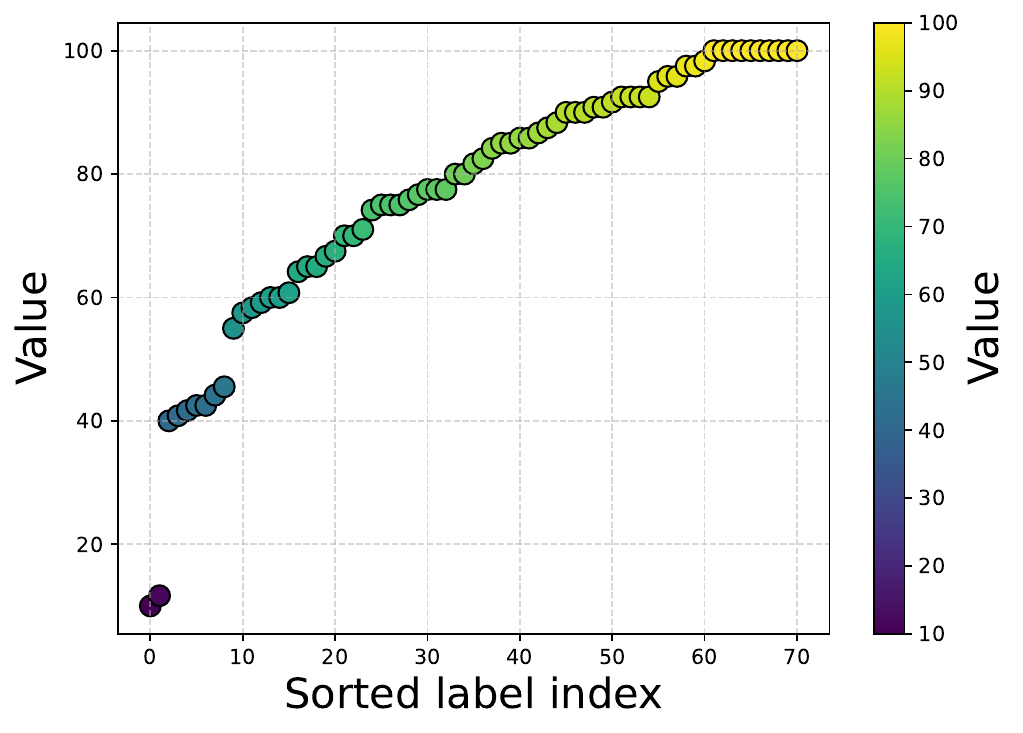}
    \caption{Lifting of the arms (LA) exercise.}
    \label{fig:sub1}
  \end{subfigure}
  \hfill
  \begin{subfigure}{0.32\textwidth}
    \includegraphics[width=\textwidth]{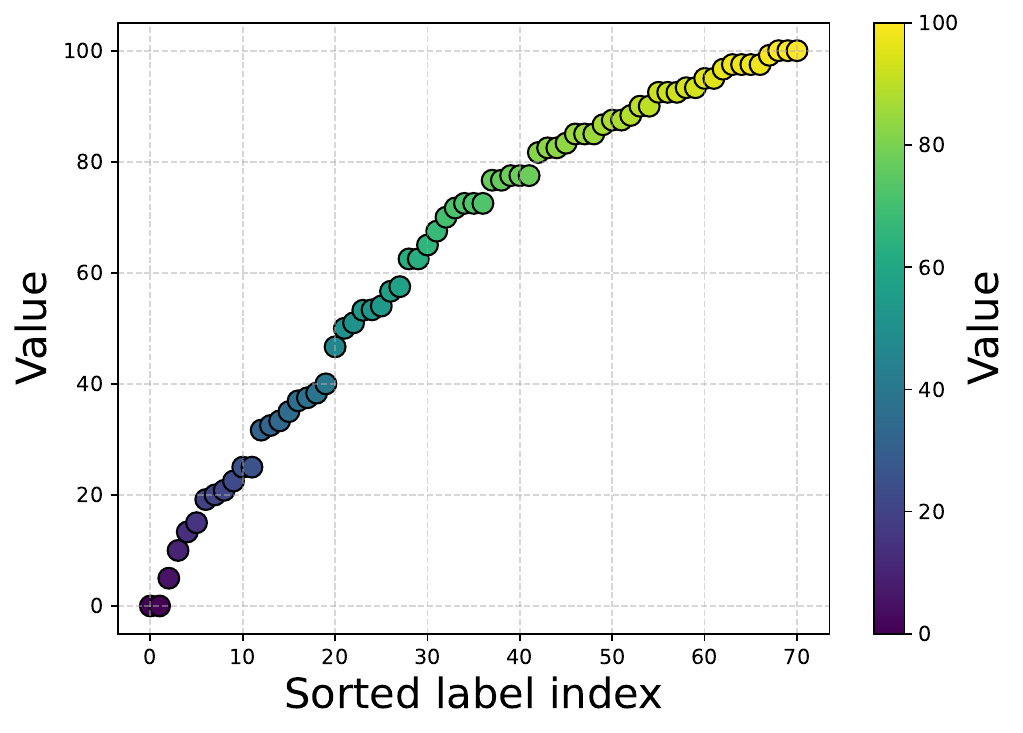}
    \caption{Lateral tilt of the trunk with the arms in extension (LT) exercise.}
    \label{fig:sub2}
  \end{subfigure}
  \hfill
  \begin{subfigure}{0.32\textwidth}
    \includegraphics[width=\textwidth]{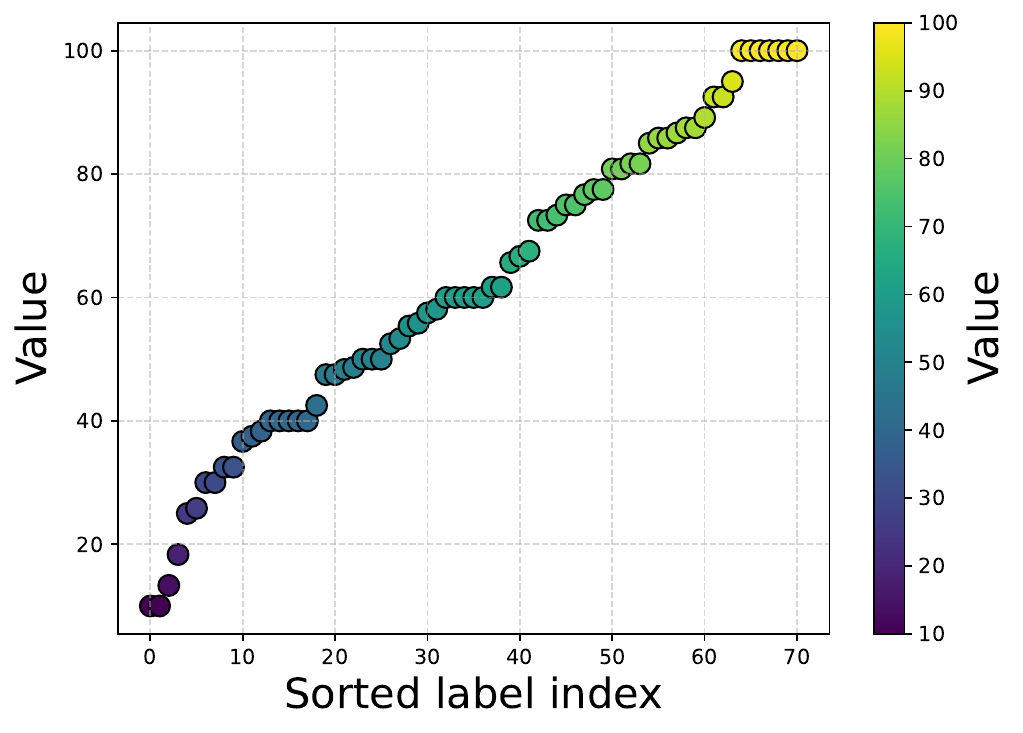}
    \caption{Pelvis rotations on the transverse plane (PR) exercise.}
    \label{fig:sub2}
  \end{subfigure}\\
  \begin{subfigure}{0.32\textwidth}
    \includegraphics[width=\textwidth]{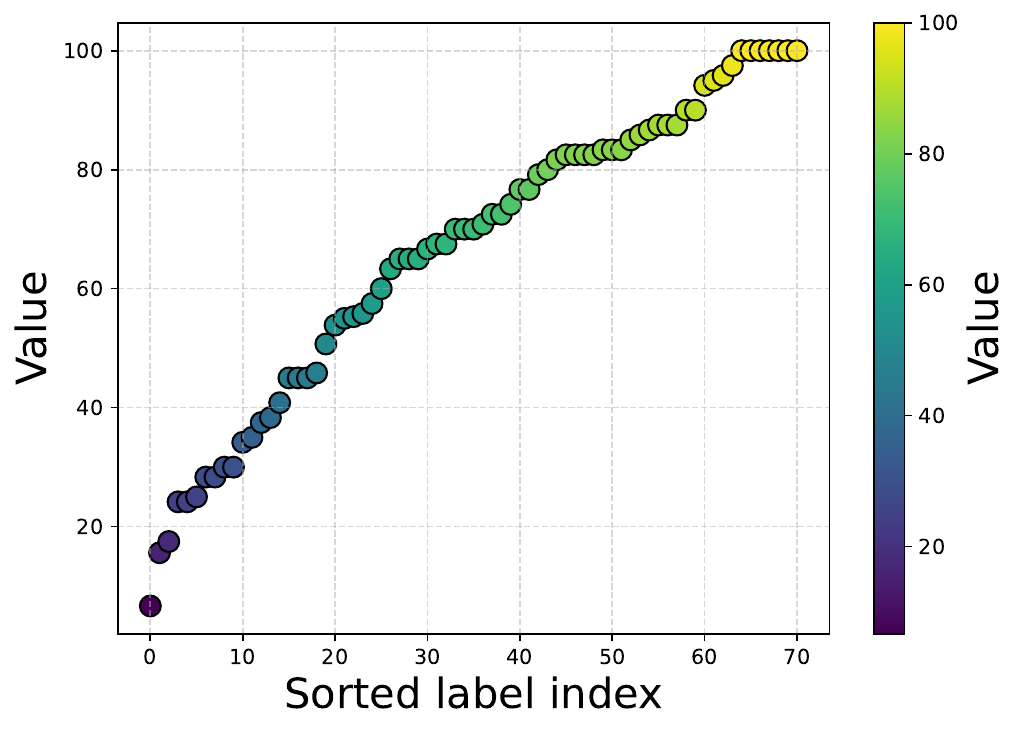}
    \caption{Squatting (Sq) exercise.}
    \label{fig:sub1}
  \end{subfigure}
  \begin{subfigure}{0.32\textwidth}
    \includegraphics[width=\textwidth]{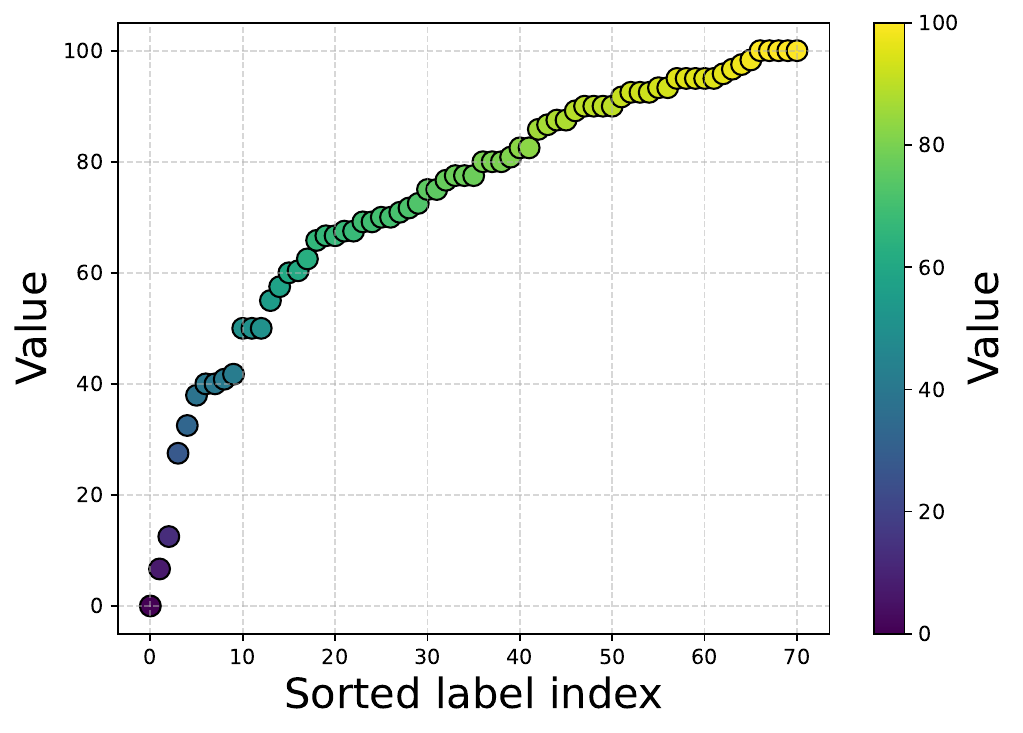}
    \caption{Trunk rotation (TR) exercise.}
    \label{fig:sub1}
  \end{subfigure}
  
  \caption{Regression labels distribution for the $5$ exercises of the KInematic assessment of MOvement for remote monitoring of physical REhabilitation (KIMORE) dataset~\cite{capecci2019kimore}.}
  \label{fig:kimore-reg-labels}
\end{figure}

\begin{figure}[h]
  \centering
  \begin{subfigure}{0.32\textwidth}
    \includegraphics[width=\textwidth]{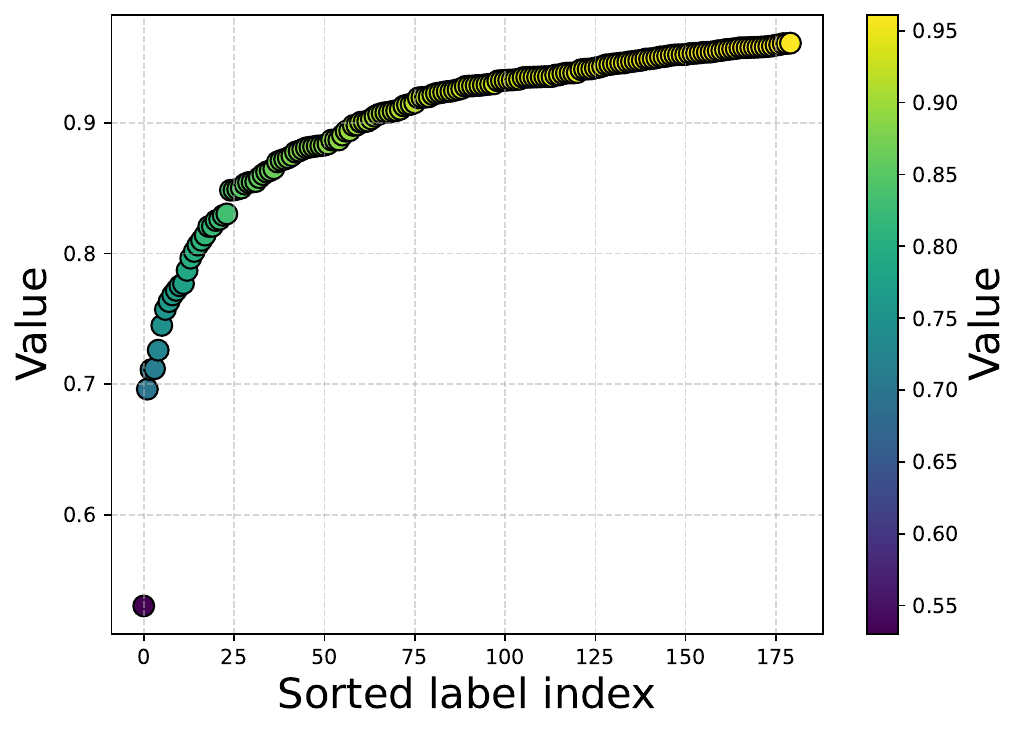}
    \caption{Deep Squat (DS) exercise.}
    \label{fig:sub1}
  \end{subfigure}
  \hfill
  \begin{subfigure}{0.32\textwidth}
    \includegraphics[width=\textwidth]{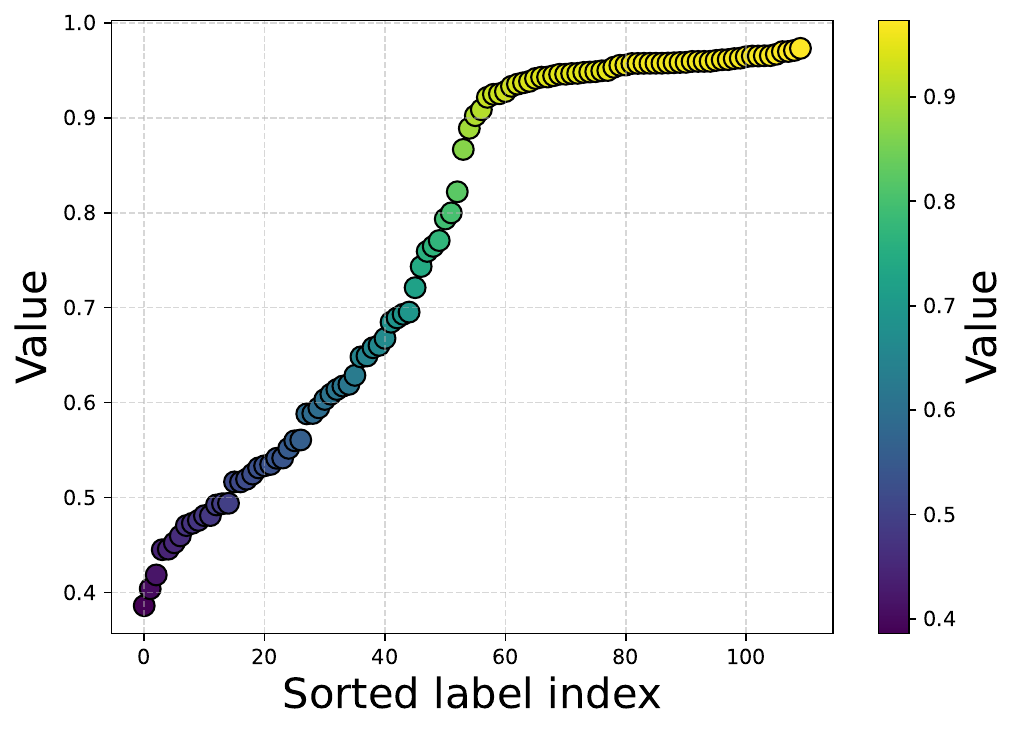}
    \caption{Hurdle Step (HS) exercise.}
    \label{fig:sub2}
  \end{subfigure}
  \hfill
  \begin{subfigure}{0.32\textwidth}
    \includegraphics[width=\textwidth]{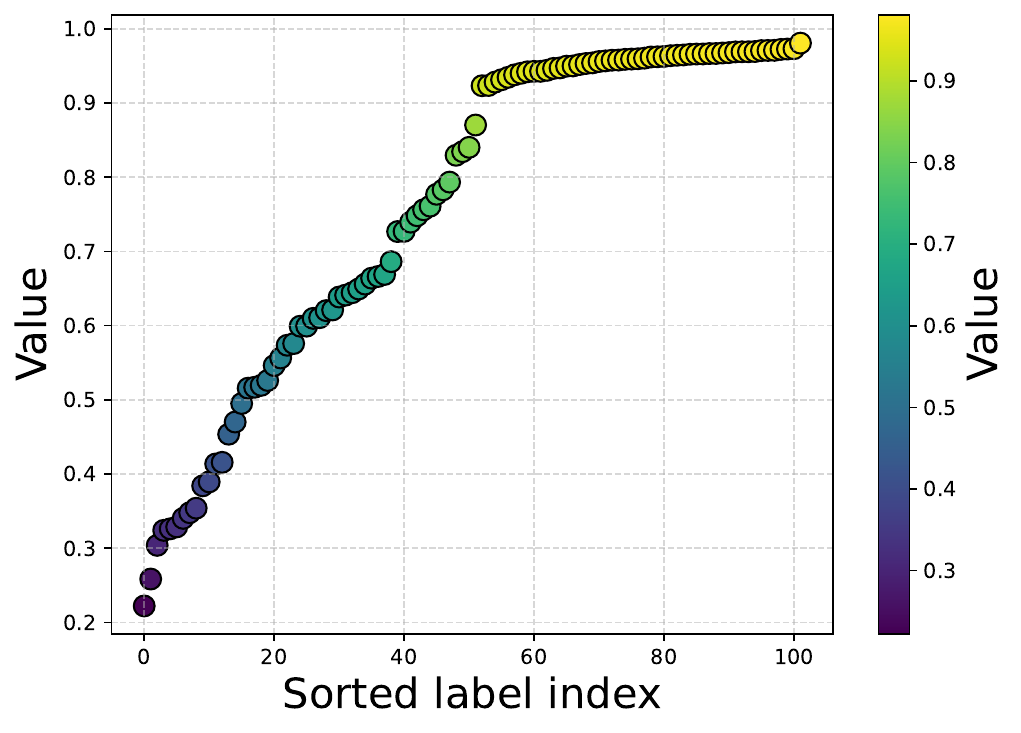}
    \caption{Inline Lunge (IL) exercise.}
    \label{fig:sub2}
  \end{subfigure}\\
  \begin{subfigure}{0.32\textwidth}
    \includegraphics[width=\textwidth]{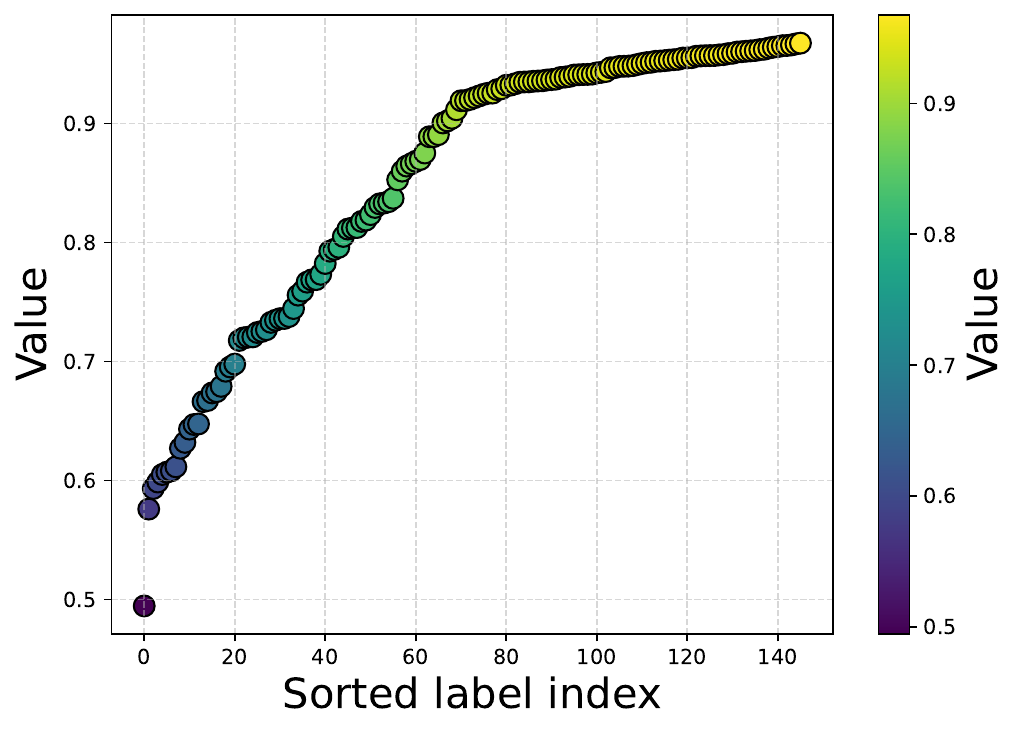}
    \caption{Standing Active Straight Leg Raise (SASLR) exercise.}
    \label{fig:sub1}
  \end{subfigure}
  \hfill
  \begin{subfigure}{0.32\textwidth}
    \includegraphics[width=\textwidth]{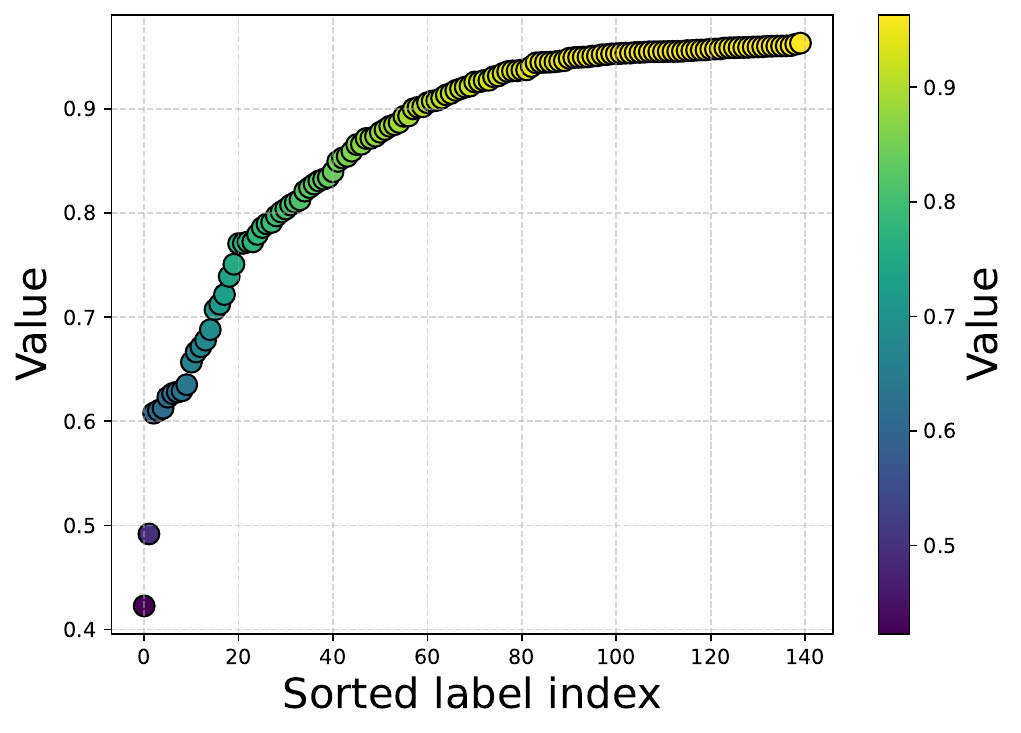}
    \caption{Side Lunge (SL) exercise.}
    \label{fig:sub1}
  \end{subfigure}
  \hfill
  \begin{subfigure}{0.32\textwidth}
    \includegraphics[width=\textwidth]{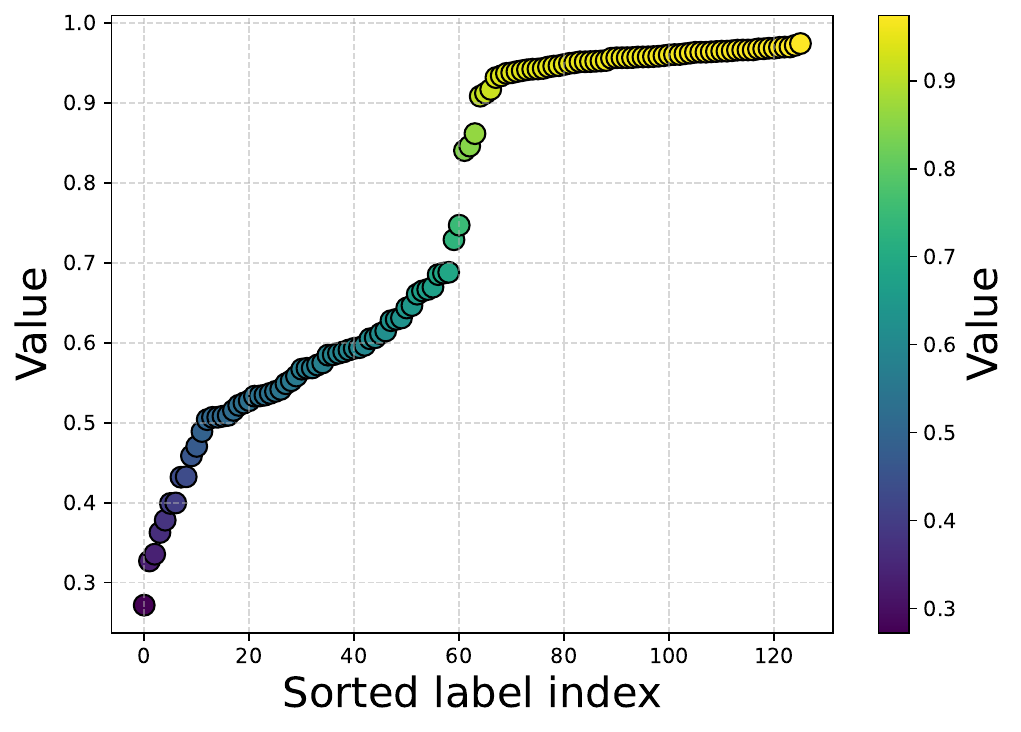}
    \caption{Standing Shoulder Abduction (SSA) exercise.}
    \label{fig:sub1}
  \end{subfigure}\\
  \begin{subfigure}{0.32\textwidth}
    \includegraphics[width=\textwidth]{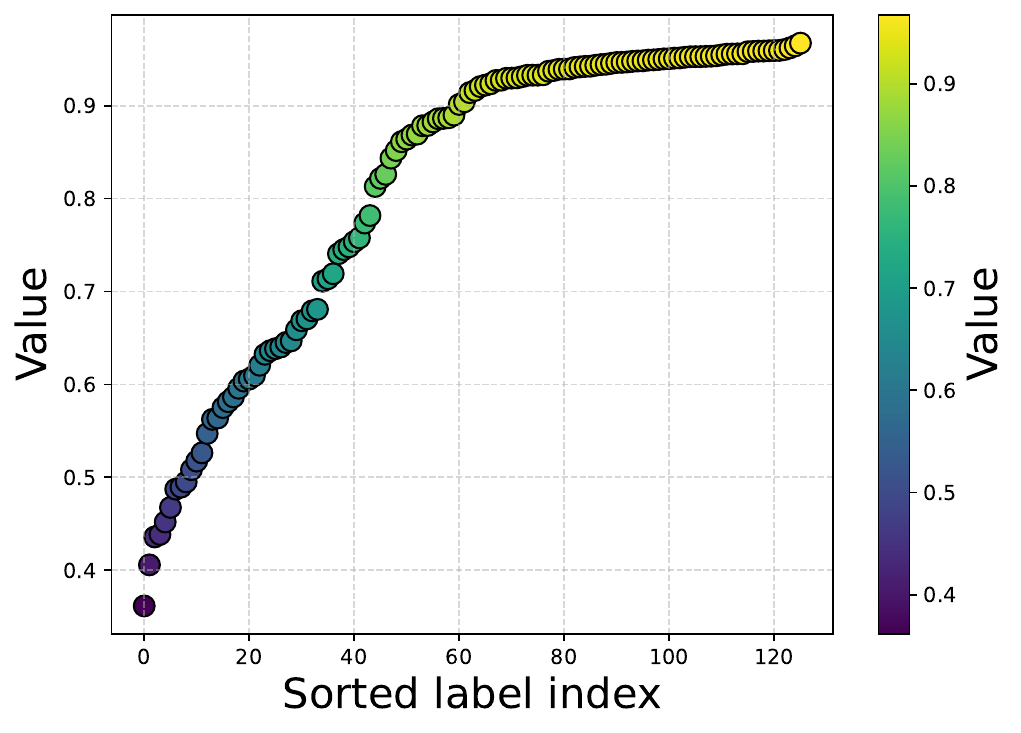}
    \caption{Standing Shoulder Extension (SSE) exercise.}
    \label{fig:sub1}
  \end{subfigure}
  \hfill
  \begin{subfigure}{0.32\textwidth}
    \includegraphics[width=\textwidth]{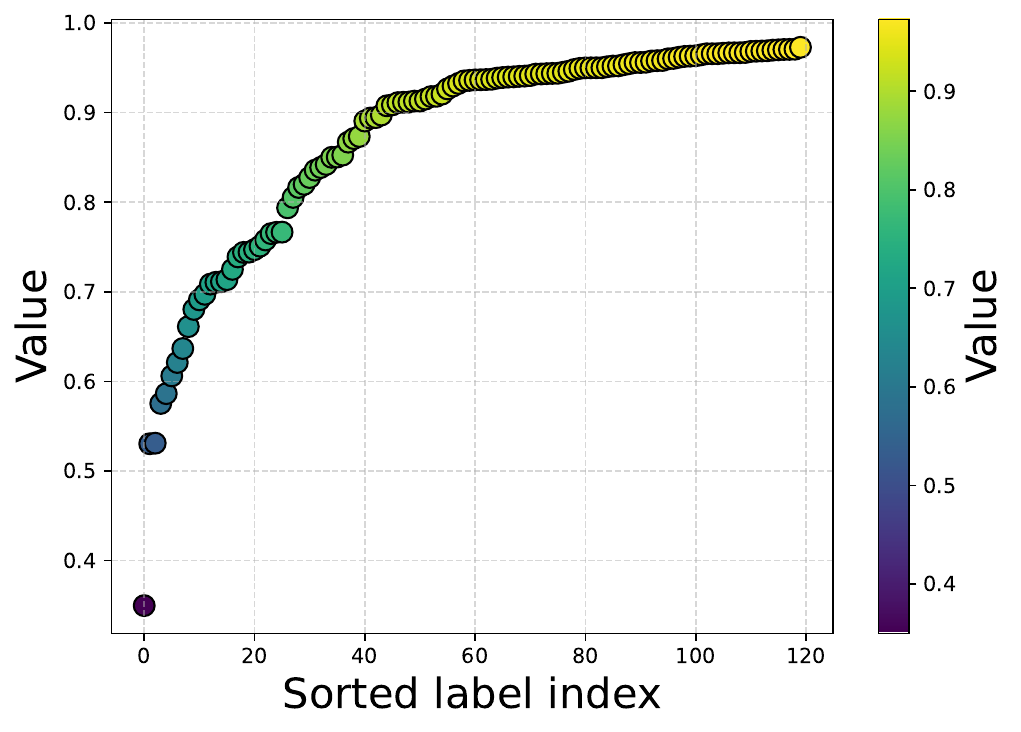}
    \caption{Standing Shoulder Internal-External Rotation (SSIER) exercise.}
    \label{fig:sub1}
  \end{subfigure}
  \hfill
  \begin{subfigure}{0.32\textwidth}
    \includegraphics[width=\textwidth]{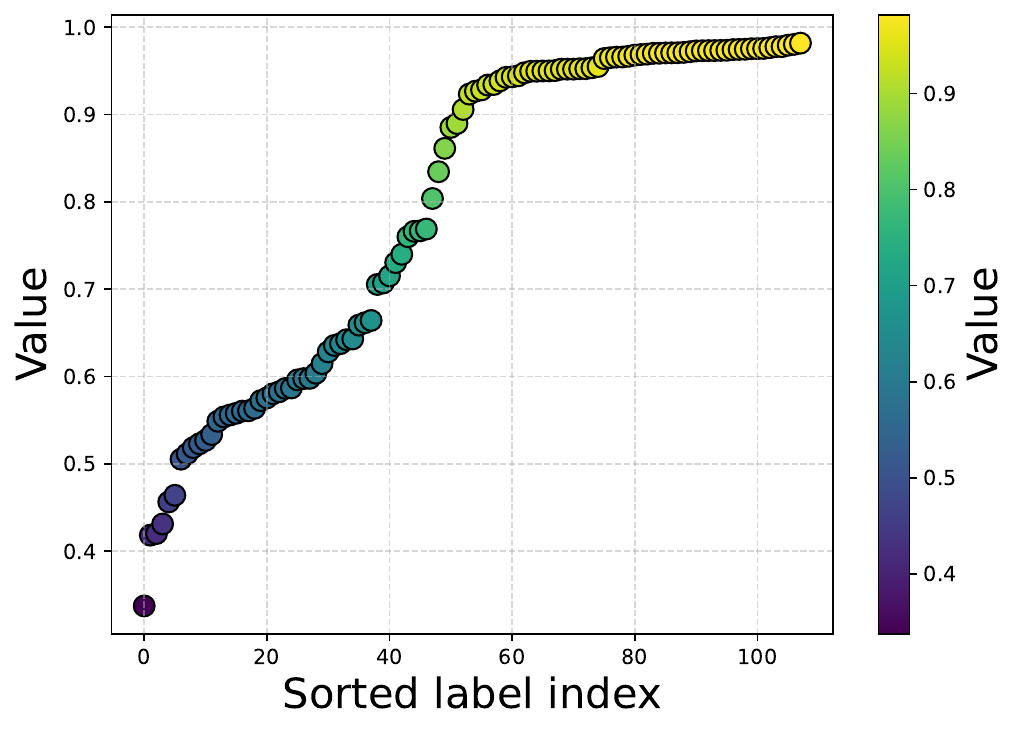}
    \caption{Standing Shoulder Scaption (SSS) exercise.}
    \label{fig:sub1}
  \end{subfigure}\\
  \begin{subfigure}{0.32\textwidth}
    \includegraphics[width=\textwidth]{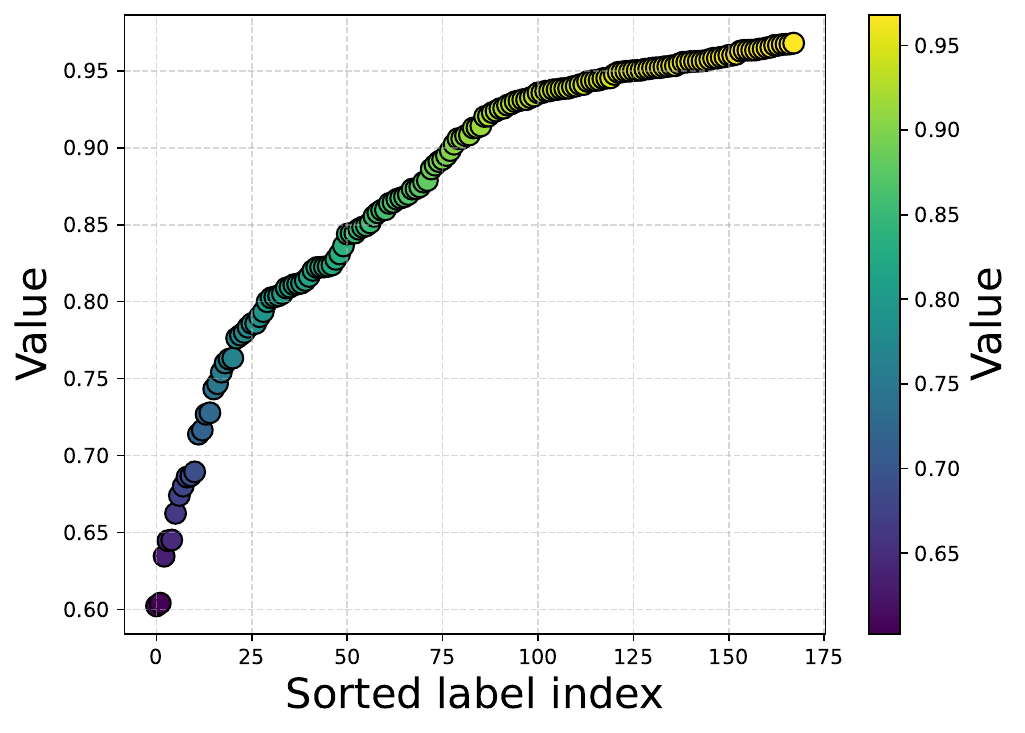}
    \caption{Sit to Stand (STS) exercise.}
    \label{fig:sub1}
  \end{subfigure}
  
  \caption{Regression labels distribution of the $10$ exercises of the s University of Idaho-Physical Rehabilitation Movement Data (UI-PRMD) dataset~\cite{vakanski2018data}.}
  \label{fig:uiprmd-reg-labels}
\end{figure}

\begin{figure}[h]
  \centering
  \begin{subfigure}{0.32\textwidth}
    \includegraphics[width=\textwidth]{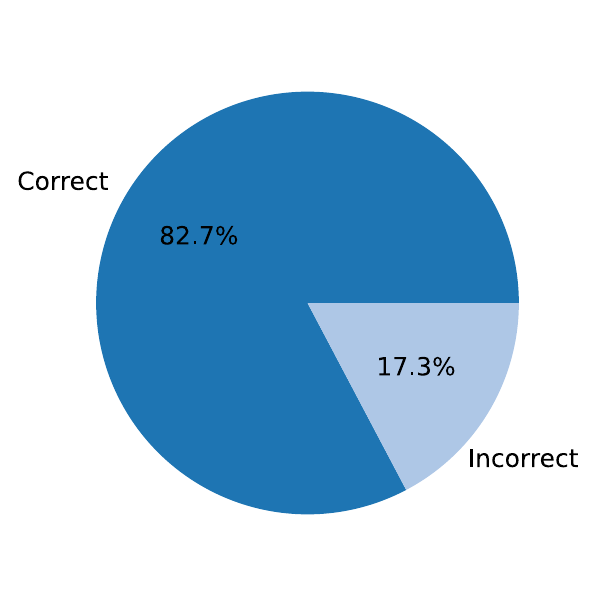}
    \caption{Elbow Flexion Left (EFL) exercise.}
    \label{fig:sub1}
  \end{subfigure}
  \hfill
  \begin{subfigure}{0.32\textwidth}
    \includegraphics[width=\textwidth]{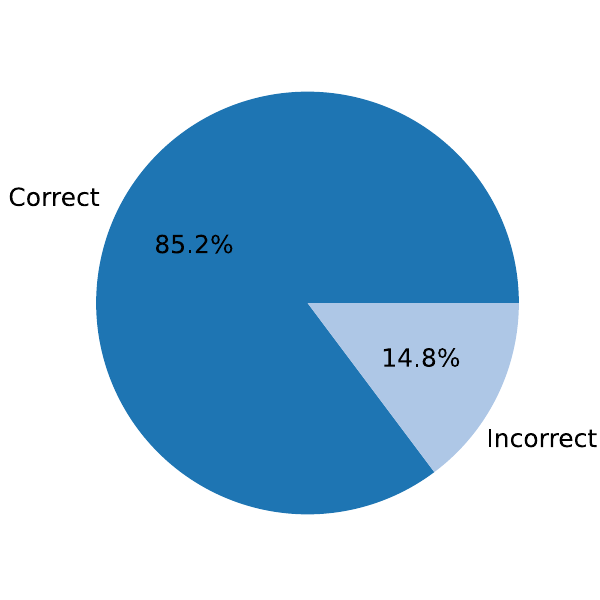}
    \caption{Elbow Flexion Right (EFR) exercise.}
    \label{fig:sub2}
  \end{subfigure}
  \hfill
  \begin{subfigure}{0.32\textwidth}
    \includegraphics[width=\textwidth]{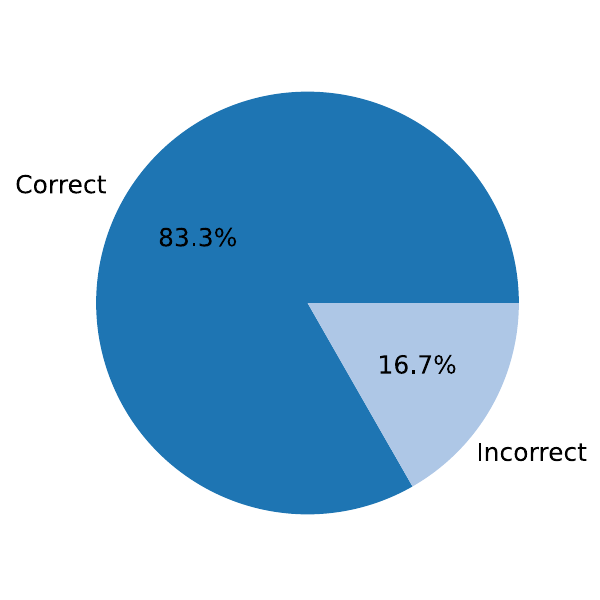}
    \caption{Shoulder Abduction Left (SAL) exercise.}
    \label{fig:sub2}
  \end{subfigure}\\
  \begin{subfigure}{0.32\textwidth}
    \includegraphics[width=\textwidth]{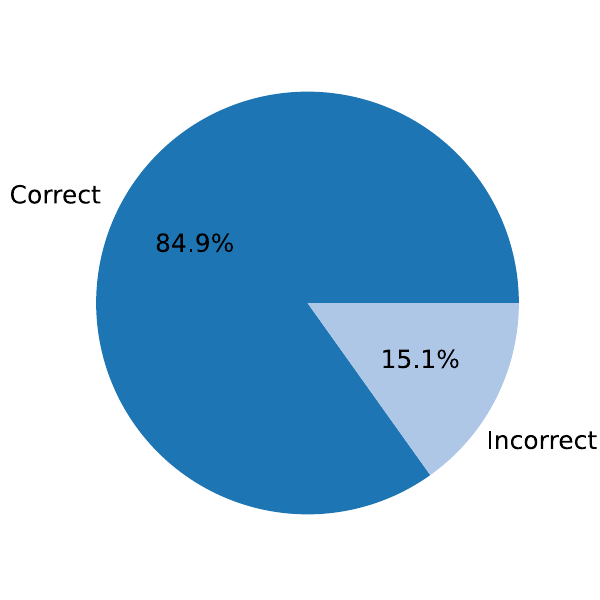}
    \caption{Shoulder Abduction Right (SAR) exercise.}
    \label{fig:sub1}
  \end{subfigure}
  \hfill
  \begin{subfigure}{0.32\textwidth}
    \includegraphics[width=\textwidth]{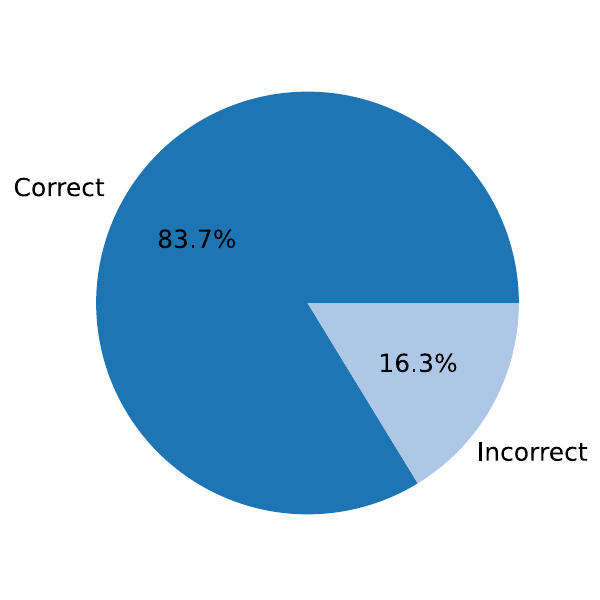}
    \caption{Shoulder Forward Elevation (SFE) exercise.}
    \label{fig:sub1}
  \end{subfigure}
  \hfill
  \begin{subfigure}{0.32\textwidth}
    \includegraphics[width=\textwidth]{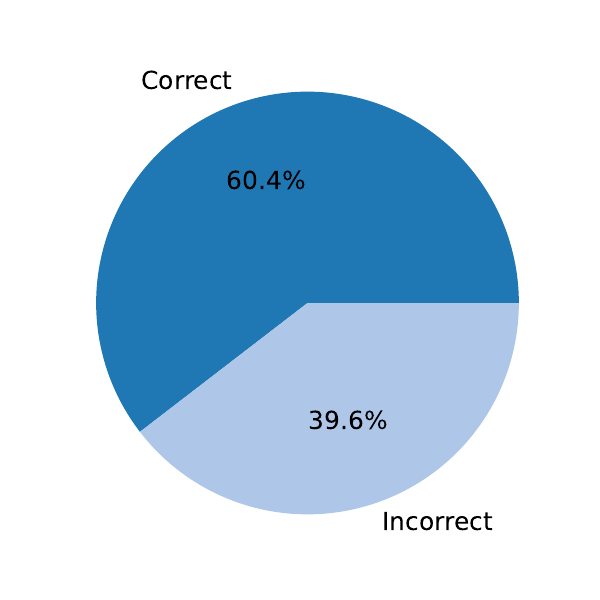}
    \caption{Shoulder Flexion Left (SFL) exercise.}
    \label{fig:sub1}
  \end{subfigure}\\
  \begin{subfigure}{0.32\textwidth}
    \includegraphics[width=\textwidth]{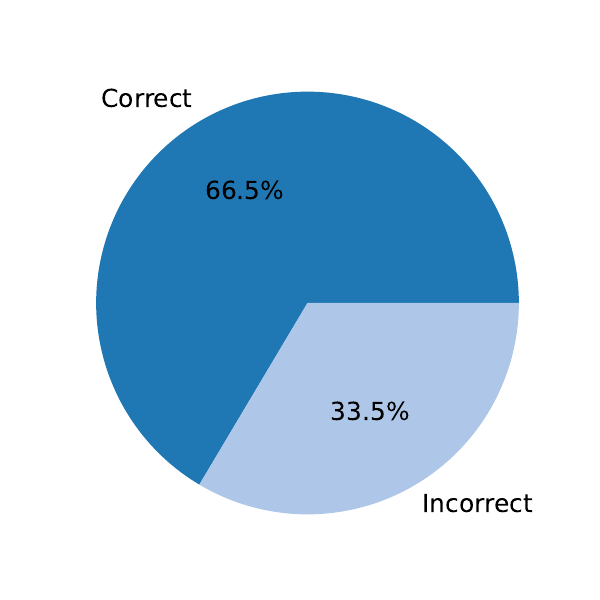}
    \caption{Shoulder Flexion Right (SFR) exercise.}
    \label{fig:sub1}
  \end{subfigure}
  \hfill
  \begin{subfigure}{0.32\textwidth}
    \includegraphics[width=\textwidth]{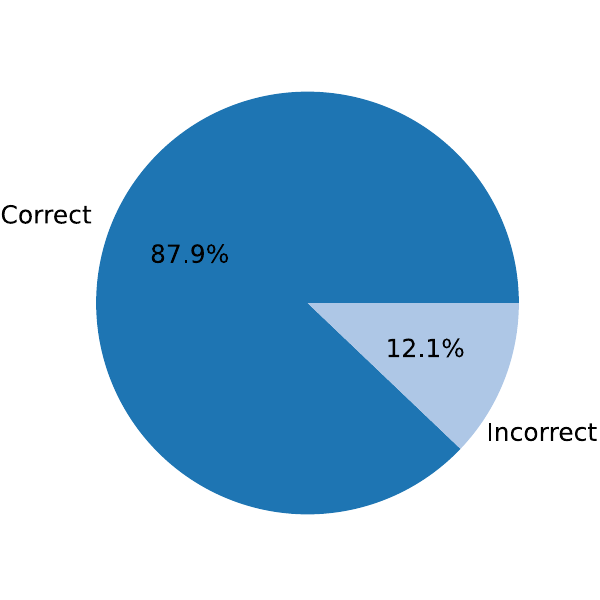}
    \caption{Side tap Left (STL) exercise.}
    \label{fig:sub1}
  \end{subfigure}
  \hfill
  \begin{subfigure}{0.32\textwidth}
    \includegraphics[width=\textwidth]{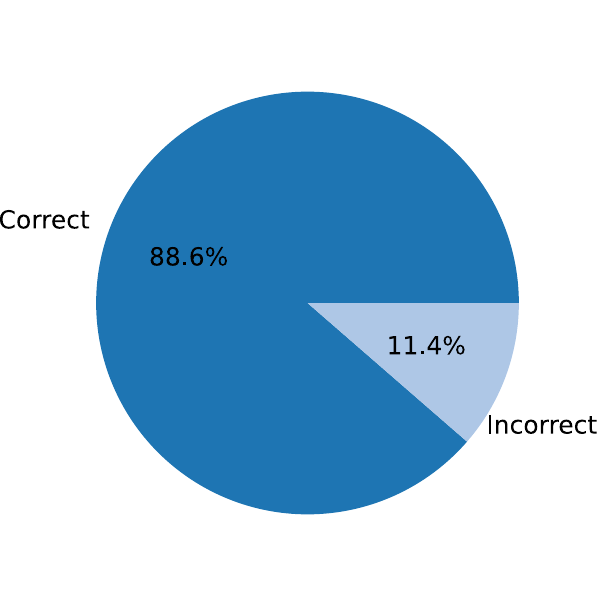}
    \caption{Side tap Right (STR) exercise.}
    \label{fig:sub1}
  \end{subfigure}
  
  \caption{Classification labels distribution of the $9$ exercises of the IntelliRehabDS (IRDS) dataset~\cite{miron2021intellirehabds}.}
  \label{fig:irds-labels}
\end{figure}

\begin{figure}[h]
  \centering
  \begin{subfigure}{0.32\textwidth}
    \includegraphics[width=\textwidth]{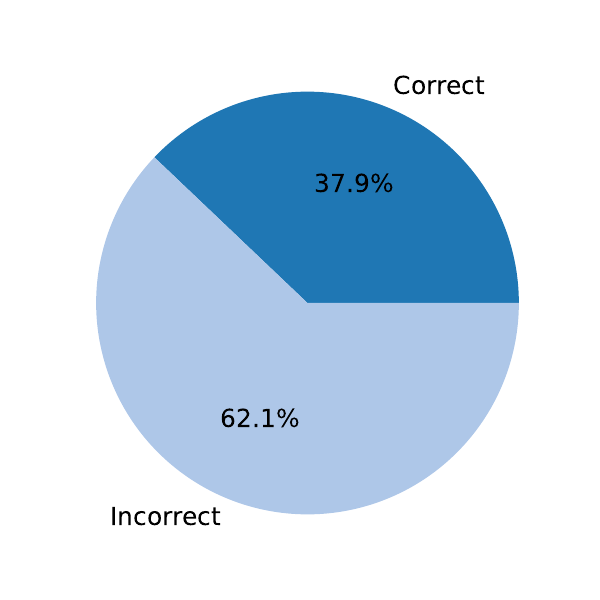}
    \caption{Binary - Upper limbs flexed at 90 degrees (CTK) exercise.}
    \label{fig:sub1}
  \end{subfigure}
  \hfill
  \begin{subfigure}{0.32\textwidth}
    \includegraphics[width=\textwidth]{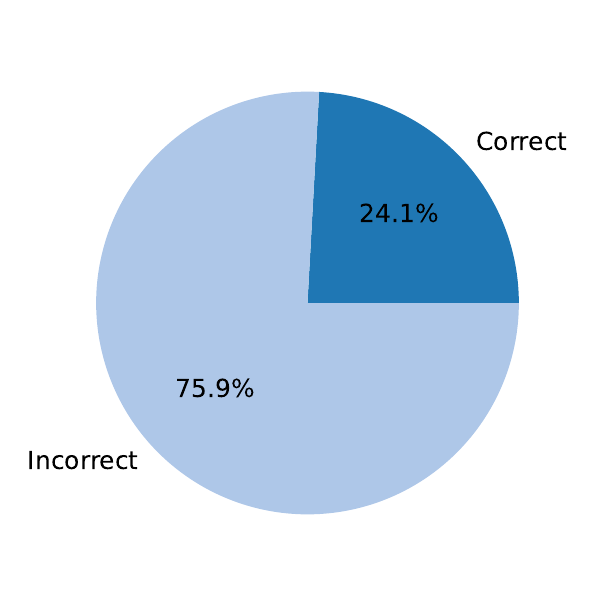}
    \caption{Binary - Lateral Bending of the Trunk (ELK) exercise.}
    \label{fig:sub2}
  \end{subfigure}
  \hfill
  \begin{subfigure}{0.32\textwidth}
    \includegraphics[width=\textwidth]{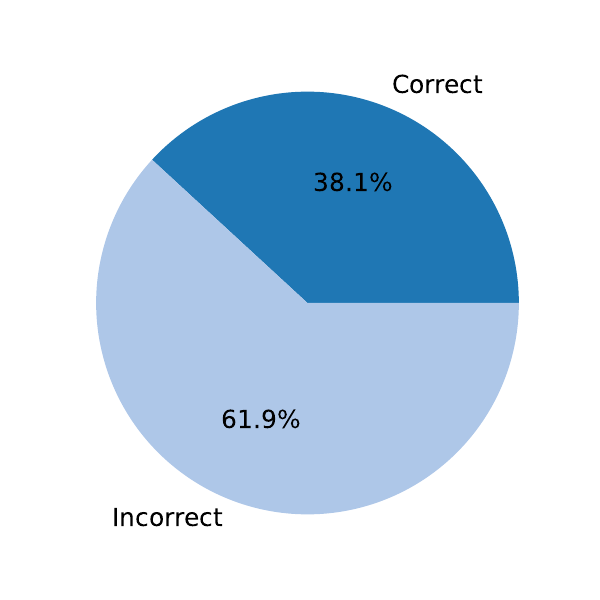}
    \caption{Binary - Rotation of the Trunk (RTK) exercise.}
    \label{fig:sub2}
  \end{subfigure}\\
  \begin{subfigure}{0.32\textwidth}
    \includegraphics[width=\textwidth]{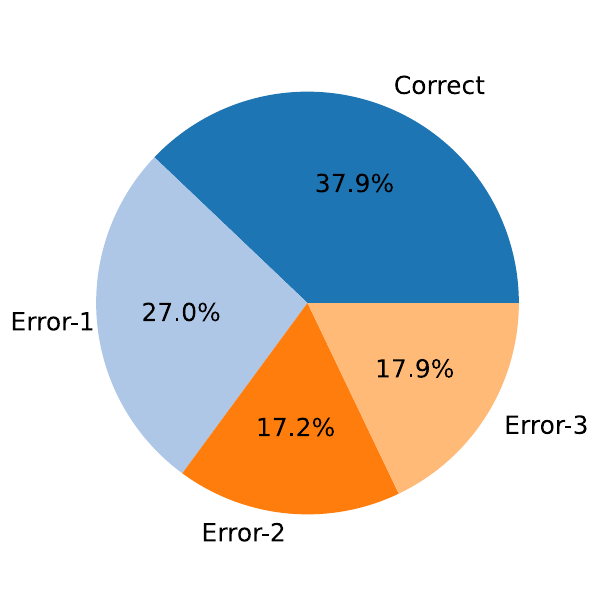}
    \caption{Multi-class - Upper limbs flexed at 90 degrees (CTK) exercise.}
    \label{fig:sub1}
  \end{subfigure}
  \hfill
  \begin{subfigure}{0.32\textwidth}
    \includegraphics[width=\textwidth]{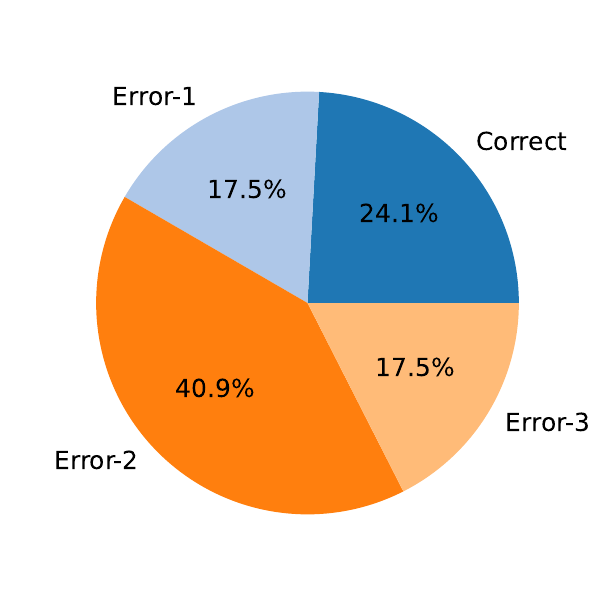}
    \caption{Multi-class - Lateral Bending of the Trunk (ELK) exercise.}
    \label{fig:sub1}
  \end{subfigure}
  \hfill
  \begin{subfigure}{0.32\textwidth}
    \includegraphics[width=\textwidth]{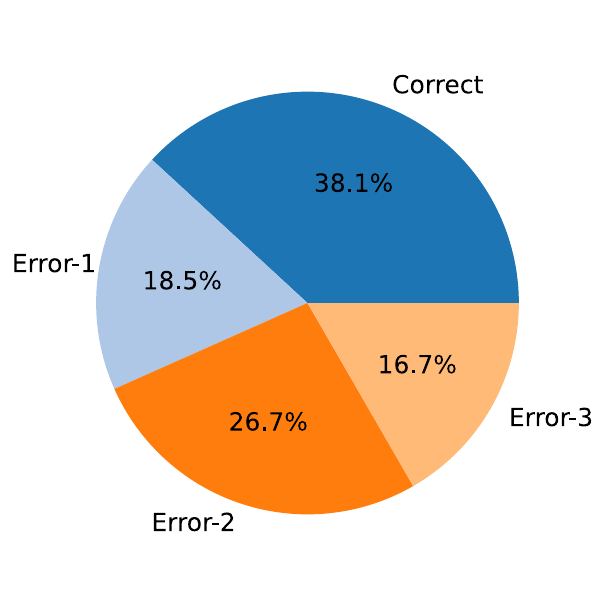}
    \caption{Multi-class - Rotation of the Trunk (RTK) exercise.}
    \label{fig:sub1}
  \end{subfigure}
  
  \caption{Classification labels distribution of the 3 exercises of the KERAAL dataset~\cite{Nguyen2024IJCNN} in both binary and multi-class cases.}
  \label{fig:keraal-labels}
\end{figure}

\begin{figure}[h]
  \centering
  \begin{subfigure}{0.32\textwidth}
    \includegraphics[width=\textwidth]{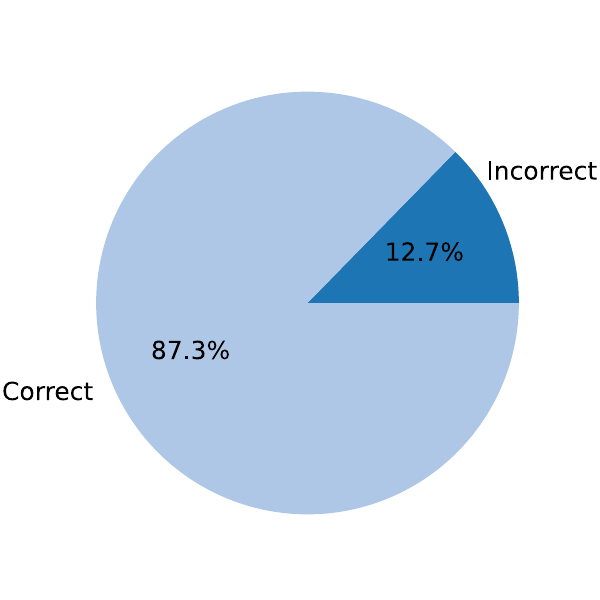}
    \caption{Lifting of the arms (LA) exercise.}
    \label{fig:sub1}
  \end{subfigure}
  \hfill
  \begin{subfigure}{0.32\textwidth}
    \includegraphics[width=\textwidth]{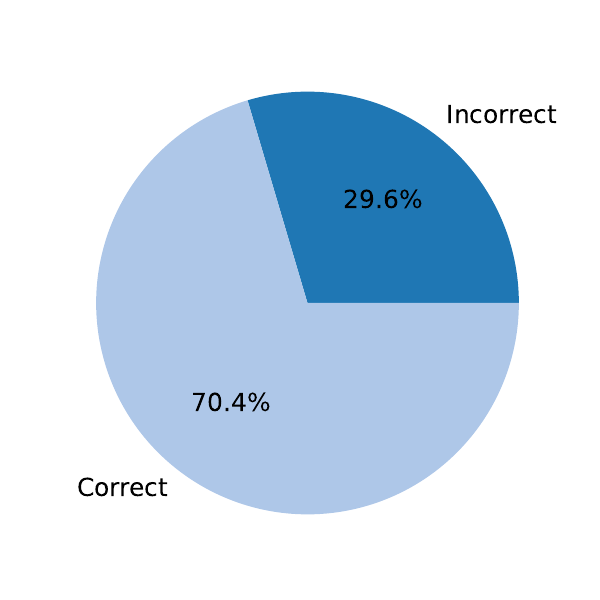}
    \caption{Lateral tilt of the trunk with the arms in extension (LT) exercise.}
    \label{fig:sub2}
  \end{subfigure}
  \hfill
  \begin{subfigure}{0.32\textwidth}
    \includegraphics[width=\textwidth]{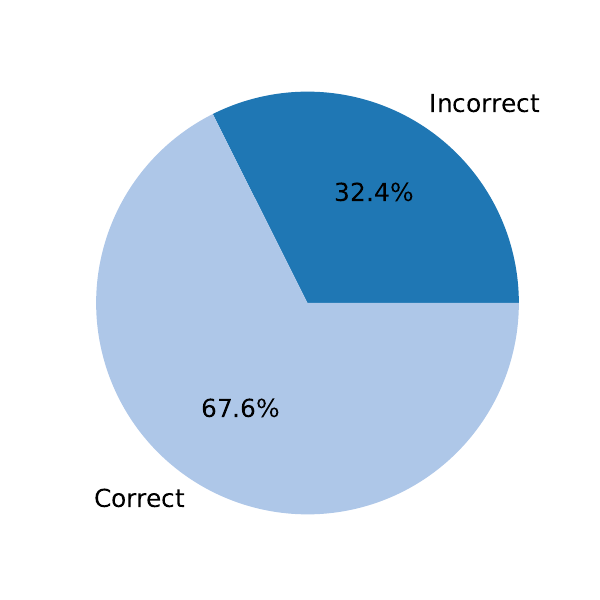}
    \caption{Pelvis rotations on the transverse plane (PR) exercise.}
    \label{fig:sub2}
  \end{subfigure}\\
  \begin{subfigure}{0.32\textwidth}
    \includegraphics[width=\textwidth]{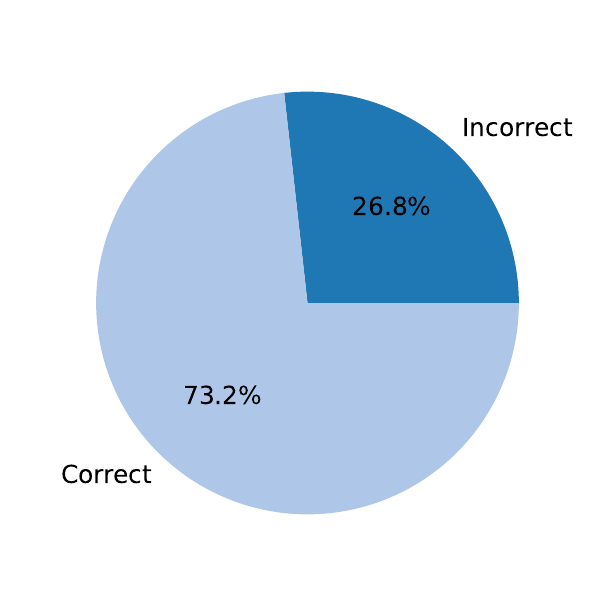}
    \caption{Squatting (Sq) exercise.}
    \label{fig:sub1}
  \end{subfigure}
  \begin{subfigure}{0.32\textwidth}
    \includegraphics[width=\textwidth]{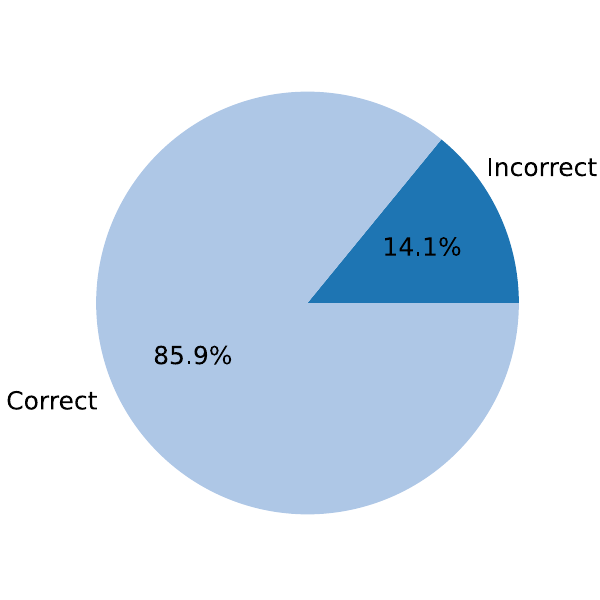}
    \caption{Trunk rotation (TR) exercise.}
    \label{fig:sub1}
  \end{subfigure}
  
  \caption{Classification labels distribution for the $5$ exercises of the KInematic assessment of MOvement for remote monitoring of physical REhabilitation (KIMORE) dataset~\cite{capecci2019kimore}.}
  \label{fig:kimore-clf-labels}
\end{figure}

\begin{figure}[h]
  \centering
  \begin{subfigure}{0.45\textwidth}
    \includegraphics[width=\textwidth]{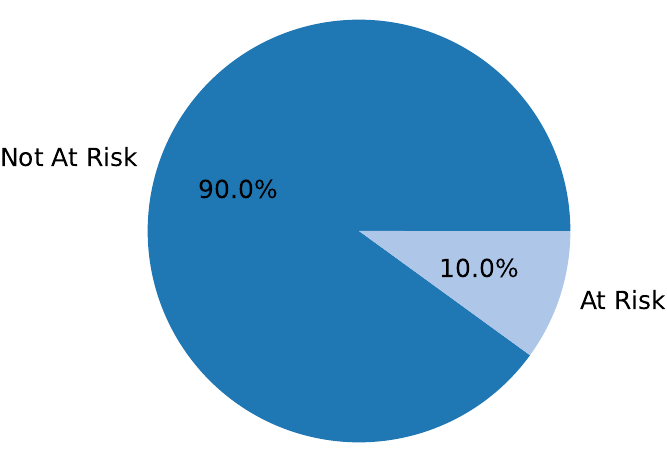}
    \caption{3m walk Front View (3WFV) exercise.}
    \label{fig:sub1}
  \end{subfigure}
  \hfill
  \begin{subfigure}{0.45\textwidth}
    \includegraphics[width=\textwidth]{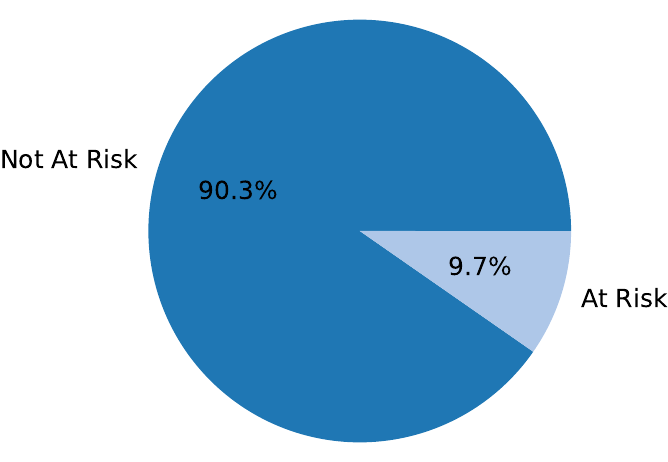}
    \caption{Get Up And Go Front View (GGFV) exercise.}
    \label{fig:sub2}
  \end{subfigure}\\
  \begin{subfigure}{0.45\textwidth}
    \includegraphics[width=\textwidth]{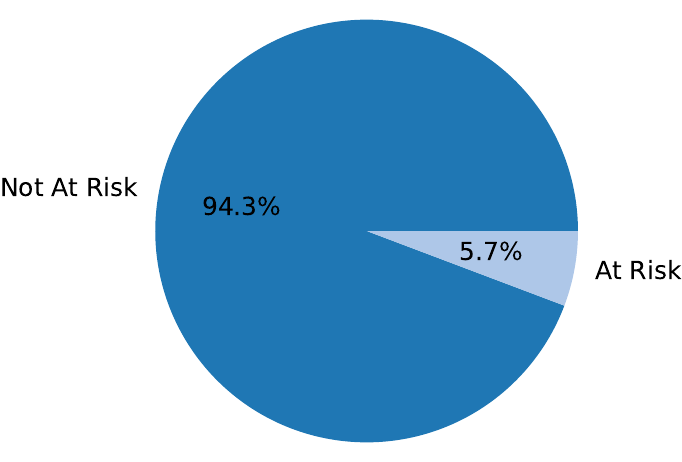}
    \caption{Quiet Standing, Eyes Closed (QSEC) exercise.}
    \label{fig:sub2}
  \end{subfigure}
  \hfill
  \begin{subfigure}{0.45\textwidth}
    \includegraphics[width=\textwidth]{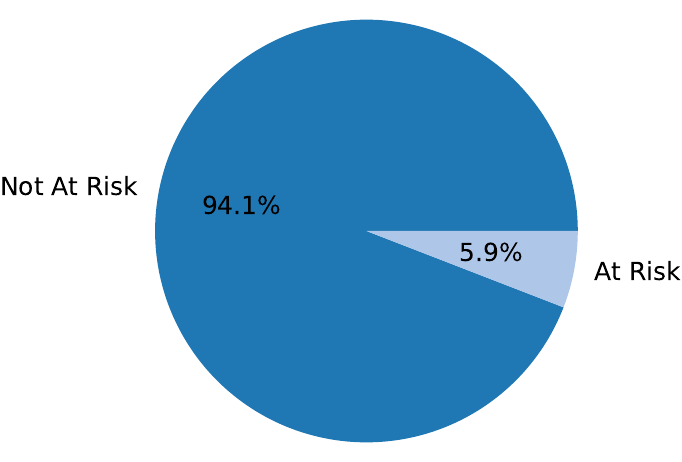}
    \caption{Quiet Standing, Eyes Open (QSEO) exercise.}
    \label{fig:sub1}
  \end{subfigure}
  
  \caption{Classification labels distribution of the $4$ exercises of the KINECAL dataset~\cite{maudsley2023kinecal}.}
  \label{fig:kinecal-labels}
\end{figure}

\begin{figure}[h]
    \centering
\includegraphics[width=0.5\linewidth]{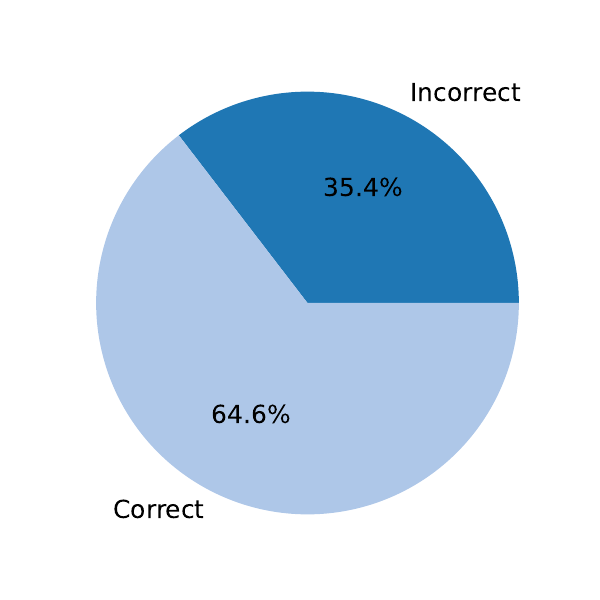}
    \caption{Classification labels distribution of the Walking Up Stairs (WUS) exercise of the Sensor Platform for Healthcare in a Residential Environment (SPHERE) dataset~\cite{paiement2014online}.}
    \label{fig:sphere-labels}
\end{figure}

\begin{figure}[h]
  \centering
  \begin{subfigure}{0.45\textwidth}
    \includegraphics[width=\textwidth]{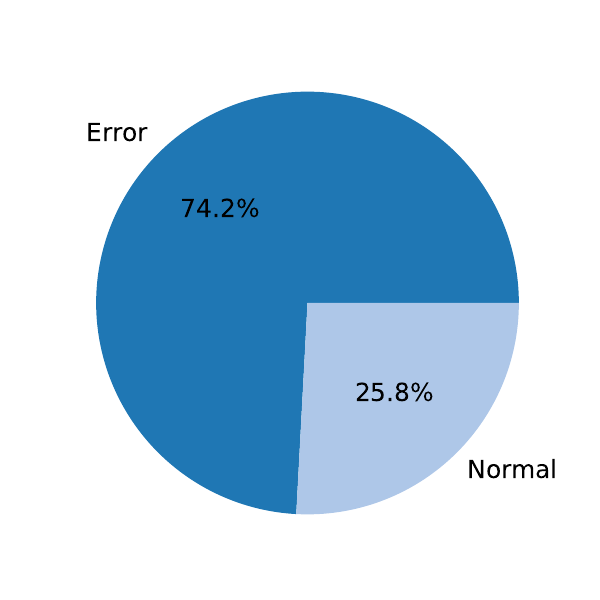}
    \caption{Binary - Military Press (MP) exercise.}
  \end{subfigure}
  \hfill
  \begin{subfigure}{0.45\textwidth}
    \includegraphics[width=\textwidth]{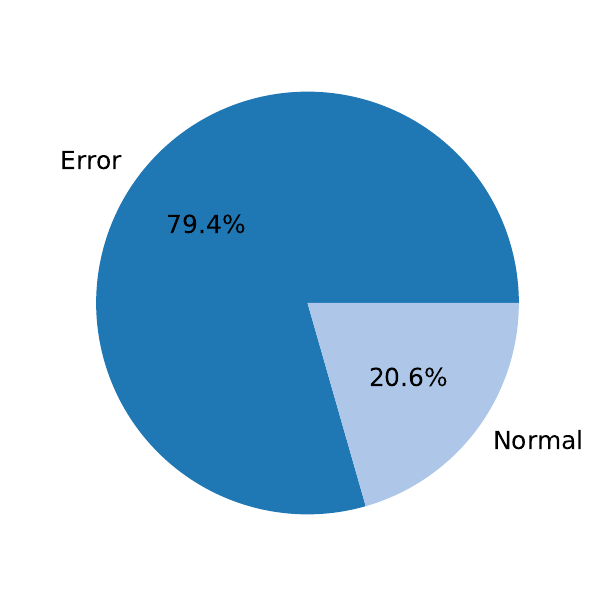}
    \caption{Binary - Rowing exercise.}
  \end{subfigure}\\
  \begin{subfigure}{0.45\textwidth}
    \includegraphics[width=\textwidth]{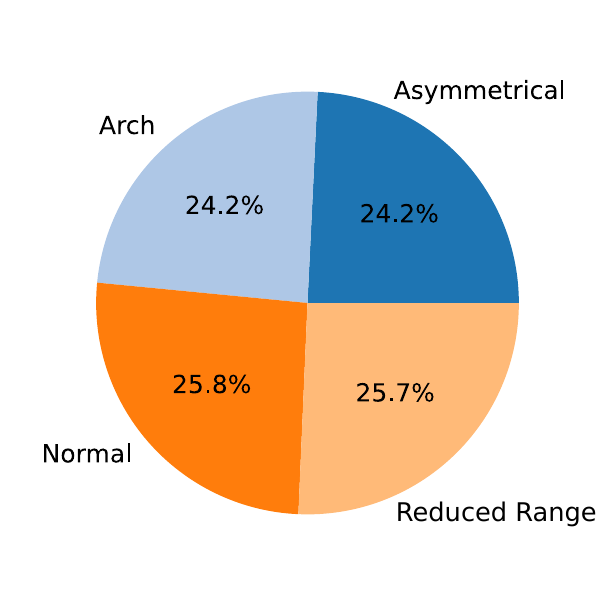}
    \caption{Multi-class - Military Press (MP) exercise.}
  \end{subfigure}
  \hfill
  \begin{subfigure}{0.45\textwidth}
    \includegraphics[width=\textwidth]{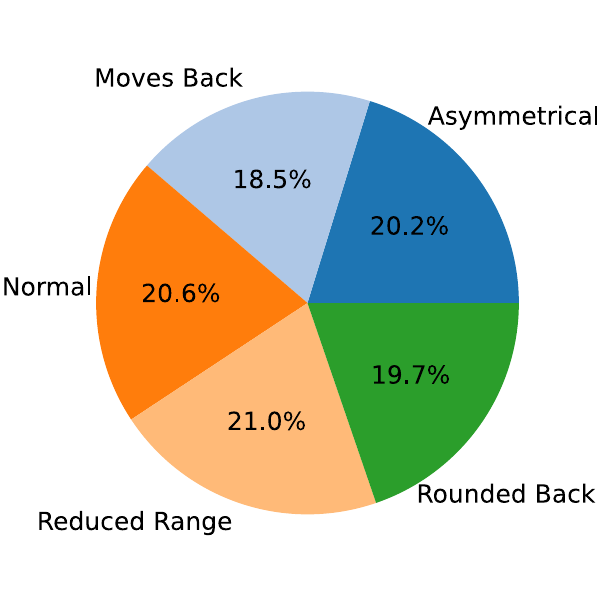}
    \caption{Multi-class - Rowing exercise.}
  \end{subfigure}
  
  \caption{Classification labels distribution of the two exercises of the University College Dublin Human Exercises (UCDHE) dataset~\cite{singh2023examination} in both binary and multi-class cases.}
  \label{fig:ucdhe-labels}
\end{figure}

\begin{figure}[h]
  \centering
  \begin{subfigure}{0.3\textwidth}
    \includegraphics[width=\textwidth]{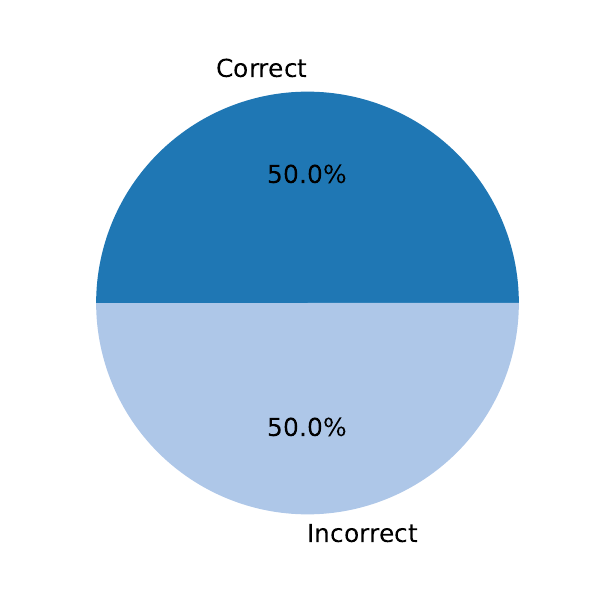}
    \caption{Deep Squat (DS) exercise.}
    \label{fig:sub1}
  \end{subfigure}
  \hfill
  \begin{subfigure}{0.3\textwidth}
    \includegraphics[width=\textwidth]{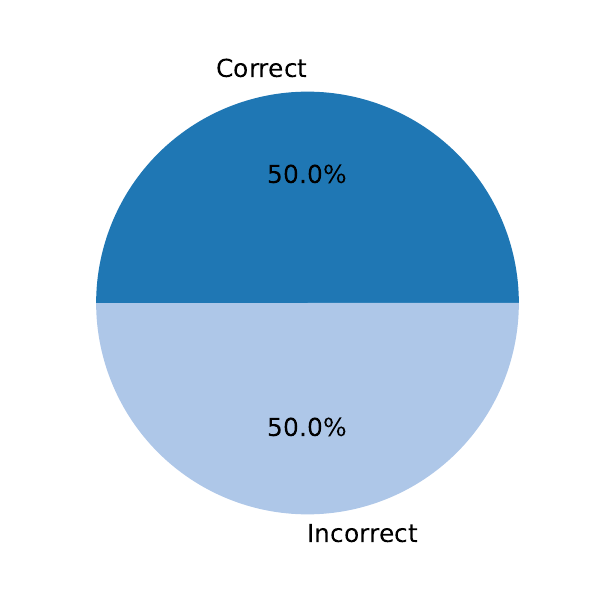}
    \caption{Hurdle Step (HS) exercise.}
    \label{fig:sub2}
  \end{subfigure}
  \hfill
  \begin{subfigure}{0.3\textwidth}
    \includegraphics[width=\textwidth]{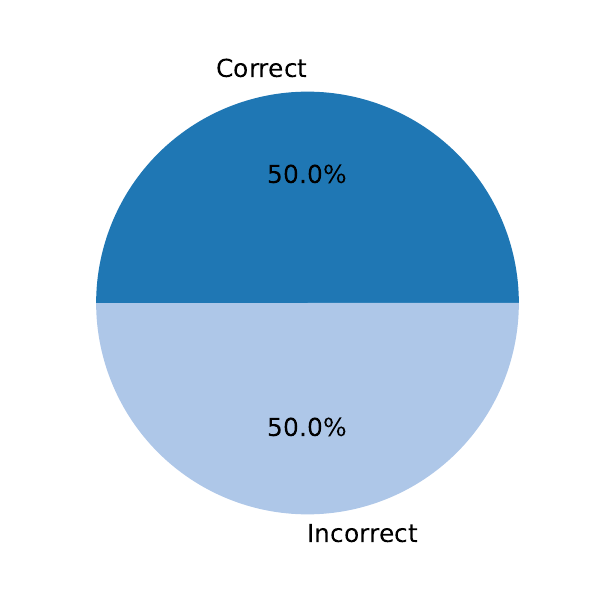}
    \caption{Inline Lunge (IL) exercise.}
    \label{fig:sub2}
  \end{subfigure}\\
  \begin{subfigure}{0.3\textwidth}
    \includegraphics[width=\textwidth]{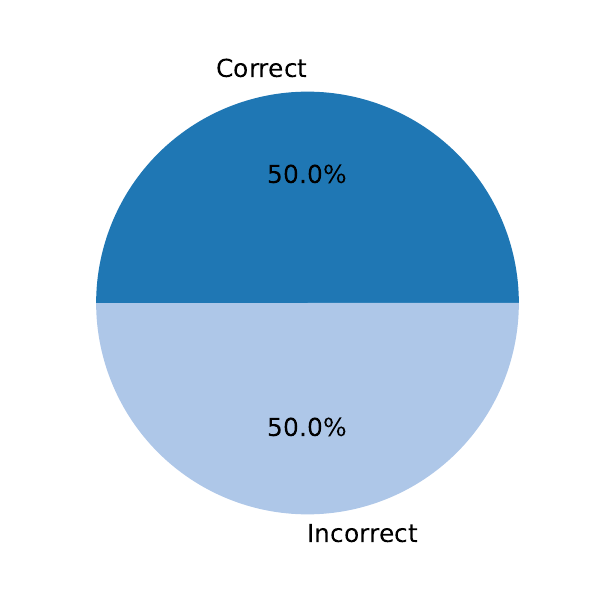}
    \caption{Standing Active Straight Leg Raise (SASLR) exercise.}
    \label{fig:sub1}
  \end{subfigure}
  \hfill
  \begin{subfigure}{0.3\textwidth}
    \includegraphics[width=\textwidth]{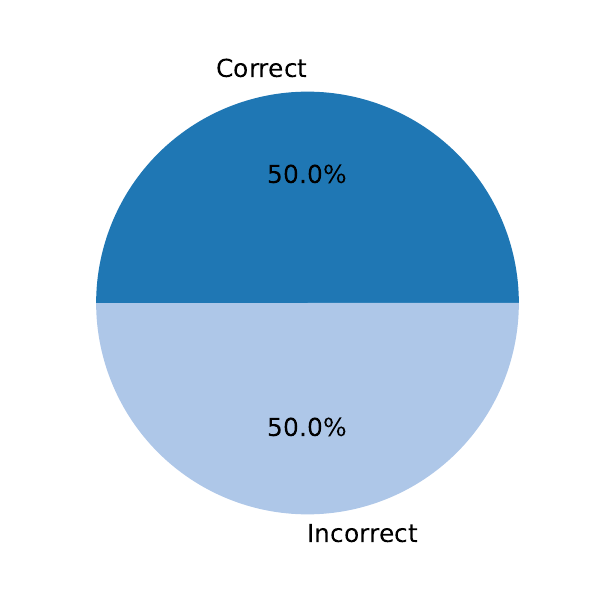}
    \caption{Side Lunge (SL) exercise.}
    \label{fig:sub1}
  \end{subfigure}
  \hfill
  \begin{subfigure}{0.3\textwidth}
    \includegraphics[width=\textwidth]{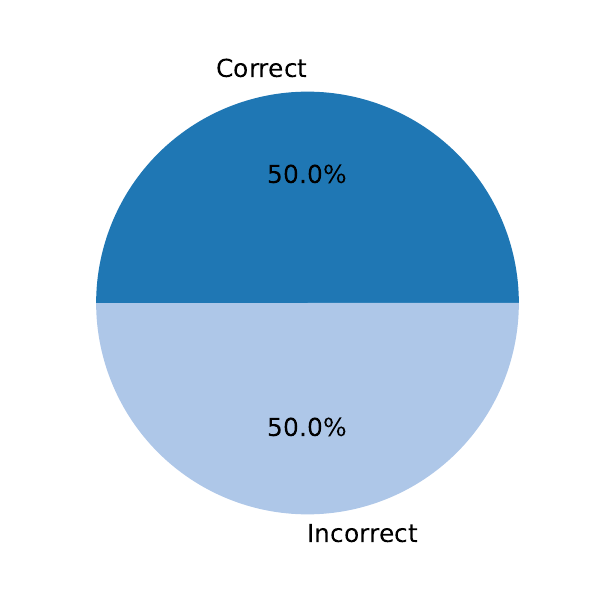}
    \caption{Standing Shoulder Abduction (SSA) exercise.}
    \label{fig:sub1}
  \end{subfigure}\\
  \begin{subfigure}{0.3\textwidth}
    \includegraphics[width=\textwidth]{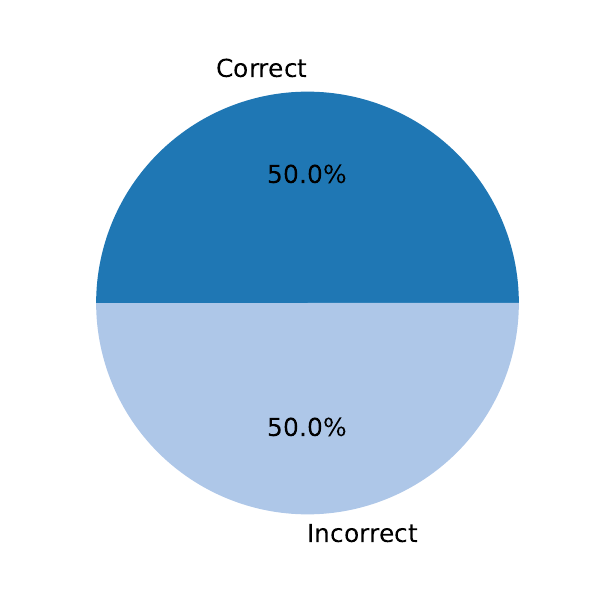}
    \caption{Standing Shoulder Extension (SSE) exercise.}
    \label{fig:sub1}
  \end{subfigure}
  \hfill
  \begin{subfigure}{0.3\textwidth}
    \includegraphics[width=\textwidth]{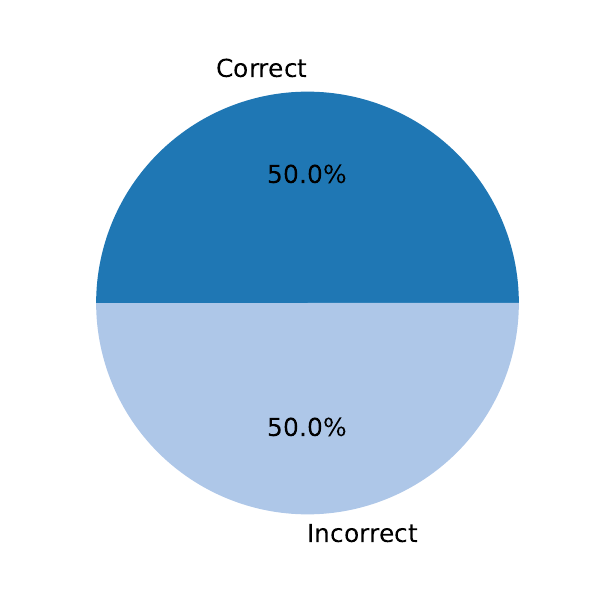}
    \caption{Standing Shoulder Internal-External Rotation (SSIER) exercise.}
    \label{fig:sub1}
  \end{subfigure}
  \hfill
  \begin{subfigure}{0.3\textwidth}
    \includegraphics[width=\textwidth]{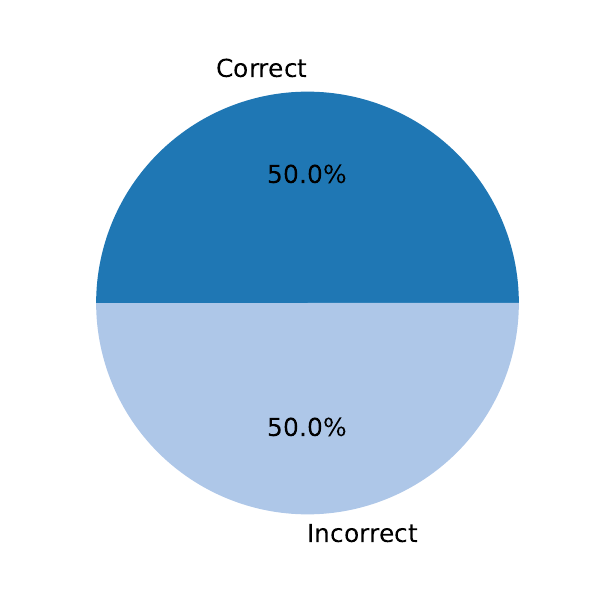}
    \caption{Standing Shoulder caption (SSS) exercise.}
    \label{fig:sub1}
  \end{subfigure}\\
  \begin{subfigure}{0.3\textwidth}
    \includegraphics[width=\textwidth]{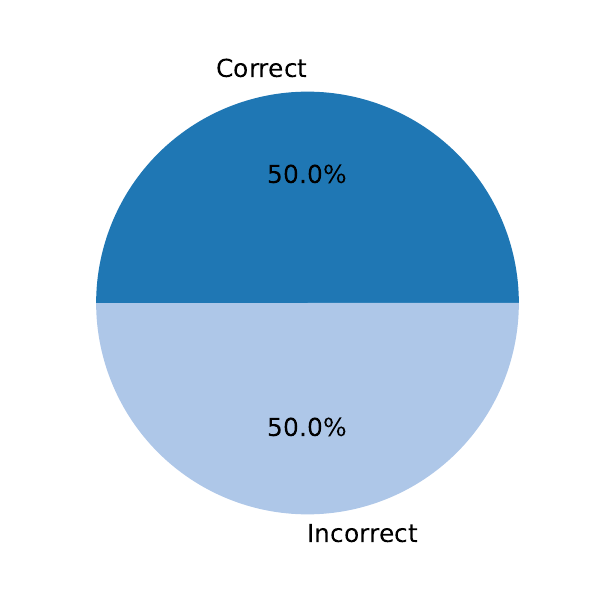}
    \caption{Sit to Stand (STS) exercise.}
    \label{fig:sub1}
  \end{subfigure}
  
  \caption{Classification labels distribution of the $10$ exercises of the s University of Idaho-Physical Rehabilitation Movement Data (UI-PRMD) dataset~\cite{vakanski2018data}.}
  \label{fig:uiprmd-clf-labels}
\end{figure}

\end{document}